\documentclass[letterpaper]{article} % DO NOT CHANGE THIS
\usepackage{aaai2026}   % DO NOT CHANGE THIS

\usepackage{times}
\usepackage{helvet}
\usepackage{courier}
\usepackage[hyphens]{url}           % 自动换行的URL
\usepackage{microtype}              % 微调字间距
\usepackage{graphicx}
\usepackage{booktabs}
\usepackage{multirow}
\usepackage{makecell}
\usepackage[table]{xcolor}
\usepackage{threeparttable}
\usepackage[caption=false]{subfig}  % 避免caption冲突
\usepackage{array}
\usepackage{enumitem}

\usepackage{algorithm}
\usepackage[endLComment=,italicComments=false]{algpseudocodex}

\usepackage{amsmath}
\usepackage{amssymb}
\usepackage{mathtools}
\usepackage{amsthm}

\usepackage{natbib}

\usepackage{comment}

\theoremstyle{plain}

\theoremstyle{definition}

\theoremstyle{remark}

\usepackage{pifont}  % ✓ ✗
\newcommand{\cmark}{\ding{51}} 
\newcommand{\xmark}{\ding{55}}

\definecolor{myred}{HTML}{F54254}
\definecolor{myorange}{HTML}{FFB135}
\definecolor{mygreen}{HTML}{10BD35}
\definecolor{myblue}{HTML}{598BE7}
\definecolor{mypurple}{HTML}{9A1C6B}
\definecolor{plgray}{HTML}{999999}
\definecolor{hiccup}{HTML}{00149B}
\definecolor{cyl}{HTML}{801dae}
\definecolor{mask}{HTML}{E7F0F9}

\usepackage{tikz}

\renewcommand{\texttt}[1]{\ifmmode\text{\ttfamily #1}\else{\ttfamily #1}\fi}

\begin{document}
\title{One-Step Generative Policies with Q-Learning: A Reformulation of MeanFlow}
% \author{Hiccup\\
% Association for the Advancement of Artificial Intelligence\\
% 2275 East Bayshore Road, Suite 160\\
% Palo Alto, California 94303\\
% }
\author {
    % Authors
    Zeyuan Wang \textsuperscript{\rm 1},
    Da Li \textsuperscript{\rm 2,3},
    Yulin Chen\textsuperscript{\rm 1},
    Ye Shi \textsuperscript{\rm 4},
    Liang Bai \textsuperscript{\rm 1},
    Tianyuan Yu \textsuperscript{\rm 1} \thanks{Corresponding author},
    Yanwei Fu \textsuperscript{\rm 5,6},
}
\affiliations {
    % Affiliations
    \textsuperscript{\rm 1}Laboratory for Big Data and Decision, National University of Defense Technology, China \\
    \textsuperscript{\rm 2} Samsung AI Center Cambridge
    \textsuperscript{\rm 3} Queen Mary University of London
    \textsuperscript{\rm 4} ShanghaiTech University \\
    \textsuperscript{\rm 5} Fudan University 
    \textsuperscript{\rm 6} Shanghai Innovation Institute 
    
    \ \{wzeyuan,ty.yu\}@nudt.edu.cn, dali.academic@gmail.com, shiye@shanghaitech.edu.cn, yanweifu@fudan.edu.cn
}
\maketitle

\begin{abstract}

We introduce a one-step generative policy for offline reinforcement learning that maps \emph{noise} directly to \emph{actions} via a \emph{residual reformulation} of MeanFlow, making it compatible with Q-learning.
While one-step Gaussian policies enable fast inference, they struggle to capture complex, multimodal action distributions. Existing flow-based methods improve expressivity but typically rely on distillation and two-stage training when trained with Q-learning.
To overcome these limitations, we propose to reformulate MeanFlow to enable \emph{direct noise-to-action generation} by integrating the velocity field and noise-to-action transformation into a single policy network—eliminating the need for separate velocity estimation. We explore several reformulation variants and identify an effective \emph{residual formulation} that supports expressive and stable policy learning. Our method offers three key advantages: 1) efficient one-step noise-to-action generation, 2) expressive modelling of multimodal action distributions, and 3) efficient and stable policy learning via Q-learning in a single-stage training setup.
Extensive experiments on 73 tasks across the OGBench and D4RL benchmarks demonstrate that our method achieves strong performance in both offline and offline-to-online reinforcement learning settings. Code is available at https://github.com/HiccupRL/MeanFlowQL.

\end{abstract}

\noindent

\section{Introduction}

Offline reinforcement learning (RL) enables the training of decision-making agents from fixed datasets, eliminating the need for online environment interaction~\citep{offlinerlHafner}.
This is especially valuable in safety-critical domains such as robotics, autonomous driving, and healthcare, where exploration can be costly or risky. A central challenge in offline RL is learning \emph{expressive yet efficient policies} that can capture complex, multi-modal action distributions while remaining amenable to stable value-based optimisation. 
% ~\citep{offlieRLtutorial}
% \begin{figure}[ht]
%   \centering
%   \includegraphics[trim=0 85 340 0, clip,width=.48\textwidth, page=6]{result.pdf} 
%   \caption{Illustration of our motivation. Unlike most flow-based methods that first predict transition velocities from noise and subsequently generate actions, our approach directly incorporates noise modelling within the network to produce actions without intermediate velocity estimation. This design effectively avoids inaccuracies in velocity prediction, particularly during early training stages, which commonly result in out-of-bound actions requiring clipping. By eliminating the mismatch between predicted and executed actions, our method significantly reduces distortions in value targets, thereby preventing training instability in the critic network. }
%   \label{fig:introduction}
%   \vspace{-0.2cm}
% \end{figure}

\begin{figure}[t]
  \centering
  \includegraphics[trim=0 100 330 0, clip,width=.48\textwidth, page=1]{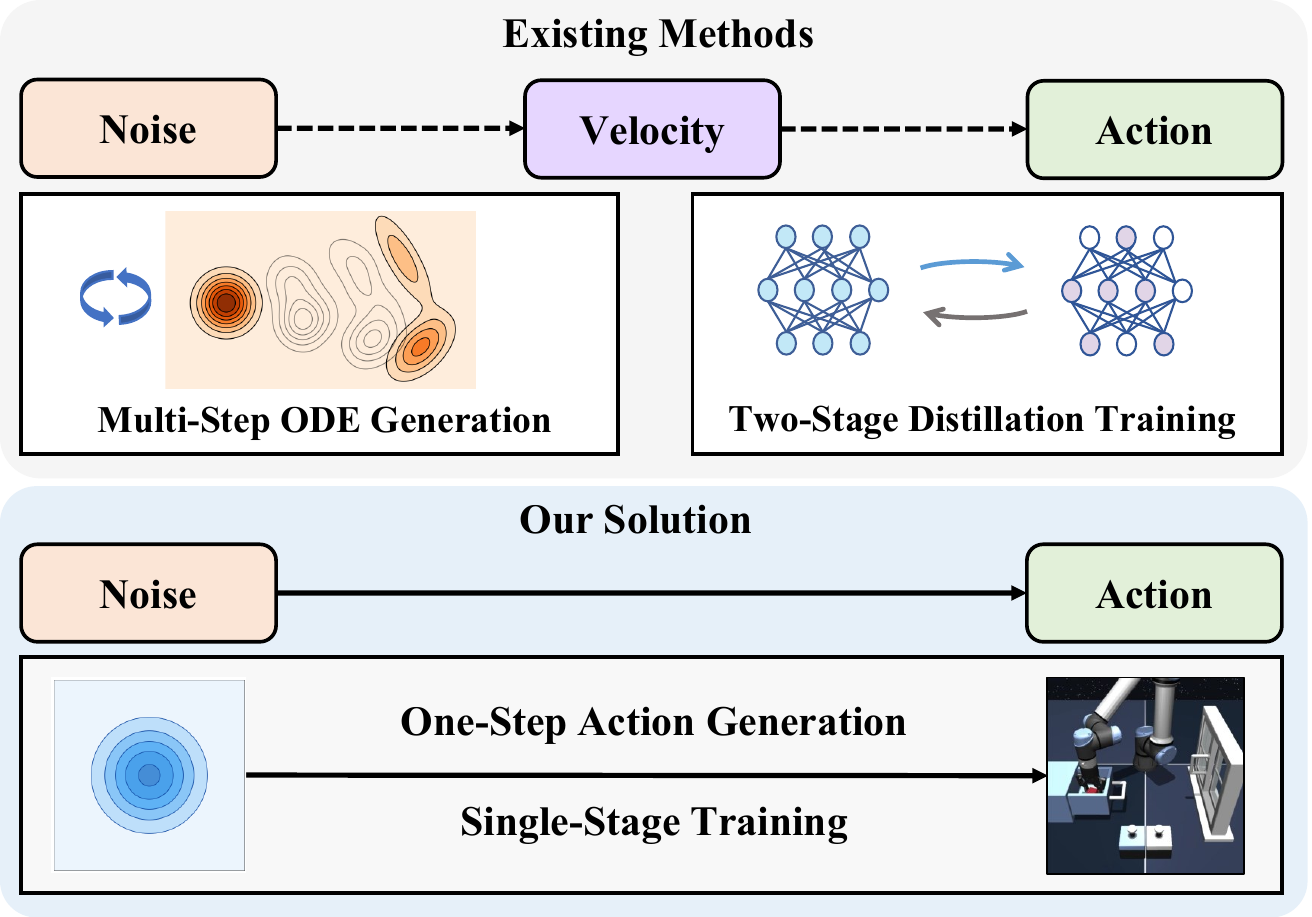} 
  \caption{Existing generative policy approaches v.s. our proposed solution. Existing methods rely on velocity-based modelling, requiring either multi-step ODE integration or two-stage distillation training. In contrast, our method enables one-step action generation directly from noise, trained via a single-stage, end-to-end process—improving both inference efficiency and training stability.}
  \label{fig:introduction}
  \vspace{-0.2cm}
\end{figure}

One-step Gaussian policies are widely adopted due to their fast and simple sampling procedures. However, their unimodal nature limits expressivity, especially in tasks that involve diverse or multi-modal  behaviour patterns~\citep{BC,ReBRAC}. In contrast, flow- and diffusion-based generative models~\citep{esser2024scaling,jin2024pyramidal,liu2023generative} have shown strong potential for modelling rich, multimodal distributions and have been increasingly applied in offline RL~\citep{fql,team2024octo,o2024open}.
Most of these approaches, however, focus on \emph{ behaviour cloning}~\citep{BC,chi2023diffusion,consistency_policy,zhang2025flowpolicy,sheng2025mp1} and do not support \emph{value-based learning}, limiting their ability to improve policies with Q-functions.

Integrating expressively generative models into Q-learning is challenging. Flow- and diffusion-based policies typically require \emph{multi-step generation}, relying on ODE or SDE solvers, which introduce \emph{backpropagation through time (BPTT)}~\citep{ding2023consistency}. This leads to computational inefficiencies and instability when optimizing jointly with a critic. To address this, recent works~\citep{chen2024diffusion,fql} adopt a \emph{two-stage distillation framework}, where a multi-step generative policy is first trained via behaviour cloning, then distilled into a one-step policy together optimised with Q-values. While this avoids BPTT, it introduces additional complexity and degrades the final policy's expressivity due to the distillation bottleneck.

In this work, we propose an \emph{expressive one-step generative policy} that can be directly optimized with Q-functions in a \emph{single-stage} training process---without requiring distillation or iterative sampling. Our approach is inspired by the MeanFlow framework~\citep{meanflow}, which enables efficient one-step sampling by modelling the average velocity of noise-to-data transitions. Although MeanFlow was originally developed for visual generation, we repurpose its structure for learning a generative policy in offline RL.

A direct adaptation of MeanFlow involves two separate processes: first, predicting a velocity vector, and then integrating it to transform noise into actions. However, we find that this approach leads to a critical issue in Q-learning: during early training, the generated actions frequently lie outside valid bounds and require \emph{post-hoc clipping}. This mismatch between policy outputs and the actions used for Bellman target computation destabilizes training and degrades performance.

One might attempt to simplify MeanFlow by reformulating its two-step process into a single network with direct, residual mapping—i.e., using \( a = \epsilon - u(\epsilon, b, t) \), where $a$ is predicted action, \( \epsilon \sim \mathcal{N}(0, I) \) is sampled noise, \( u \) is a learnable average velocity field, and \( b, t \) denote the start and end times. While this formulation eliminates the need for explicit velocity integration, we find that such a naive reformulation suffers from poor training dynamics. In toy experiments, this model underfits and fails to capture multimodal data distributions, reintroducing the risk of generating mismatched actions—mirroring the challenges seen in the original two-step MeanFlow adaptation.

To address these issues, we propose a revised \emph{residual MeanFlow formulation} with a carefully constructed mapping:
\begin{equation}
g(a_t, b, t) = a_t - u(a_t, b, t),
\end{equation}
where \( a_t \) is a linear interpolation between an offline-dataset action and noise. Unlike the naive form, this formulation retains theoretical equivalence to the original MeanFlow under mild assumptions, with guarantees provided by the \emph{Universal Approximation Theorem}~\citep{tabuada2020universal} (see Methodology and Appendix B.3 for further discussion).

Our method offers three key advantages: i) Efficient one-step noise-to-action generation without iterative inference; ii) Expressive modelling of complex, multimodal action distributions via reformulated MeanFlow-based policies; iii) Stable joint optimisation with Q-functions through direct, single-stage training---without distillation. In addition, we introduce two practical enhancements to improve the efficacy of our method, including  
a value-guided rejection sampling and an adaptive behaviour cloning regularisation (See Methodology section for details). Collectively, they contribute to the robustness and performance of our one-step policy, leading to strong results across 73 tasks on the OGBench~\citep{ogbench} and D4RL~\citep{d4rl} benchmarks in both the offline and offline-to-online settings (see Experiments).

\begin{figure}[t]
  \centering
  \includegraphics[trim=0 305 420 0, clip,width=.48\textwidth, page=2]{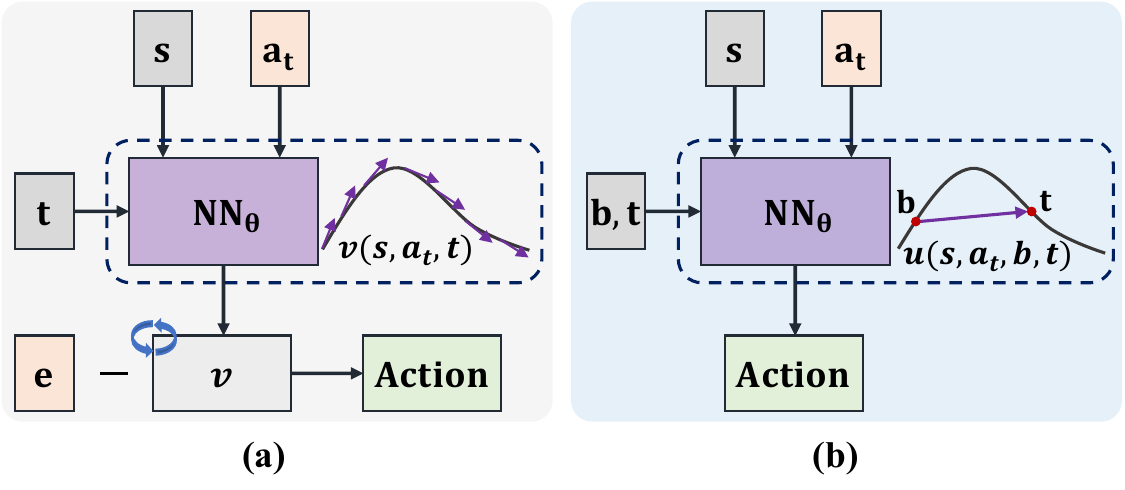}
  \caption{
Comparison between traditional and proposed flow-based policy network schemes.  
(a) Given state $s$ and noisy input $a_t$, traditional flow-based methods estimate the \emph{instantaneous velocity} $v(s, a_t, t)$ at time $t$, as a one-step approximation of multi-step ODE integration.
(b) Our method reformulate MeanFlow to estimate the \emph{average velocity} over the interval $[b, t]$, enabling direct one-step action generation.
}
  \label{fig:intro_network}
  \vspace{-0.2cm}
\end{figure}

\section{Preliminary}
\subsection{Offline Reinforcement Learning}
% 介绍offline RL一些基础的概念
In this work, we consider a Markov Decision Process (MDP) $\mathcal{M} = (\mathcal{S}, \mathcal{A}, r, \rho, p, \gamma)$ ~\citep{sutton,puterman2014markov}, 
where $\mathcal{S}$ denotes the state space, $\mathcal{A} \subseteq \mathbb{R}^d$ is the $d$-dimensional continuous action space, 
$r: \mathcal{S} \times \mathcal{A} \rightarrow \mathbb{R}$ is the reward function, 
$\rho \in \Delta(\mathcal{S})$ is the initial state distribution, 
$p(s'|s, a)$ defines the transition dynamics, and $\gamma \in [0, 1)$ is the discount factor. In offline RL, the agent does not interact with the environment during training. Instead, it learns from a fixed dataset 
$\mathcal{D} = \{\tau^{(n)}\}_{n=1}^{N}$ composed of trajectories $\tau = (s_0, a_0, r_0, \ldots, s_H, a_H, r_H)$, 
collected by some unknown behaviour policy. The typical objective is to learn a policy $\pi_\theta(a|s)$ 
that maximizes the expected discounted return with some regularization:
\begin{equation}
\mathcal{L}_\pi(\theta) = -J(\pi_\theta) + \alpha R(\pi_\theta),\label{equ:j+r}
\end{equation}
where  $J(\pi_\theta) = \mathbb{E}_{\tau \sim p^{\pi_\theta}(\tau)} \left[ \sum_{t=0}^{H} \gamma^t r(s_t, a_t) \right]$, $p^{\pi_\theta}(\tau)$ is the trajectory distribution induced by $\pi_\theta$ under the dynamics $p$ and initial state $\rho$, and $\alpha$ controls the regularisation strength.

In the behaviour cloning (BC) regularised actor-critic learning framework, the objective in Eq~\ref{equ:j+r} becomes
\begin{equation}
\mathcal{L}_\pi(\theta) = \mathbb{E}_{s,a\sim\mathcal{D},\,a^\pi\sim\pi_\theta} \left[ -Q_\phi(s,a^\pi) - \alpha \log \pi_\theta(a|s) \right],
\label{equ:imitation}
\end{equation}
where the second term is a behaviour cloning regularizer, \(\mathcal{D}\) denotes the offline dataset and \(Q_\phi(s,a)\) is the critic parameterised by \(\phi\), which is trained to minimise the Bellman error as follows
\begin{equation}
\mathcal{L}_Q = \mathbb{E}_{(s,a,r,s')\sim\mathcal D}\Bigl[Q_\phi(s,a)-(r+\gamma\,Q_{\bar{\phi}}(s',\pi_\theta(s')))\Bigr]^2,
\end{equation}
where \(\bar{\phi}\) corresponds to the target network.

\subsection{Flow Matching and MeanFlow} 
\label{sec:flow-matching}
Flow Matching learns to transform noise into data via time-dependent velocities, whereas MeanFlow enables one-step sampling by modelling the time-averaged velocity.

\paragraph{Flow matching.} Flow matching~\citep{lipman2023flow,rectifiedflow} is a generative modelling framework that learns a continuous time-dependent velocity field bridging a simple base distribution and a complex data distribution. In our context, let $e \sim p_{\text{prior}}$ be sampled from a simple prior (e.g. a standard Gaussian) and $a \sim p_{\text{data}}$ from the empirical action data distribution. We consider an interpolation $a_t = (1-t)\,a + t\,e$ for $t \in [0,1]$. Flow matching trains a velocity field $v_\theta(a_t,t)$ to predict the ground-truth displacement between $a$ and $e$. Specifically, one trains $v_\theta$ by minimizing:
\begin{equation}
\min_{\theta}\; \underset{\substack{a \sim p_{\text{data}},\\ \,e \sim p_{\text{prior}},\,t\sim \mathcal{U}(0,1)}}{\mathbb{E}}\big[\|v_\theta(a_t,t) - (e - a)\|_2^2\big],
\label{eq:flow_matching}
\end{equation}
where $a_t = (1-t)a + t e$ as above. By learning to predict $e - a$, the model captures the time-dependent instantaneous velocity needed to transform $a$ into $e$ (or vice versa). 

\paragraph{MeanFlow.} MeanFlow~\citep{meanflow} builds on flow matching by introducing an \emph{average velocity} field $u(a_t, b, t)$. For any $0 \leq b < t \leq 1$, the average velocity is defined as the time-average of the instantaneous velocity $v$ over the interval $[b,t]$:
\begin{equation}
u(a_t, b, t) \;\coloneqq\; \frac{1}{\,t - b\,} \int_{b}^{t} v(a_\tau,\tau)\,d\tau.
\label{eq:meanflow_identity}
\end{equation}
Differentiating both sides with respect to $t$ (treating $b$ as constant) yields the \emph{MeanFlow Identity} as introduced in~\citep{meanflow}:
\begin{equation}
u(a_t,b,t) \;=\; v(a_t,t)\;-\;(t-b)\,\frac{\partial}{\partial t}u(a_t,b,t).
\label{eq:meanflow_diff}
\end{equation}

MeanFlow establishes a principled relationship between the average velocity $u$ and the instantaneous velocity $v$. Crucially, it enables \textit{one-step} generation: by choosing $b=0$ and $t=1$ in \eqref{eq:meanflow_identity}, one can generate a sample in a single step:
\begin{equation}
\hat{a} \;=\; e \;-\; u_\theta(e,\;b=0,\;t=1),
\label{eq:meanflow_onestep}
\end{equation}
where $e \sim \mathcal{N}(0,I)$ and $u_\theta$ is the learned average velocity function. This formula generates $\hat{a}$ via one-step velocity estimation using the learned $u_\theta$, without integrating over time.

% \begin{comment}
% Is this section needed?
\paragraph{Flow policies in RL.} Recent works have proposed using flow matching to learn policies in RL. In such \emph{flow policies}, the policy $\pi_\theta$ is implicitly defined by a state-conditional velocity field $v_\theta(s,a_t,t)$. 
% One can train these via a flow-matching loss instead of the vanilla behavioural cloning loss. 
For example, one offline training objective for a flow policy is:
\begin{equation}
\mathcal{L}_{\text{Flow}}(\theta) = \underset{{ \substack{(s,a)\sim D,\; \\ e \sim \mathcal{N}(0,I),\;t \sim \mathcal{U}(0,1) }}}{\mathbb{E}} \;\big[\|v_\theta(s,a_t,t) - (e - a)\|_2^2\big],\footnote{Note that \textit{state} $s$ is an important input of our model. But for simplicity, and to save space in our equations, we will continue to denote the pair $(s,a_t)$ simply by $a_t$}
\label{eq:flow_policy_loss}
\end{equation}
with $a_t = (1-t)a + te$. The flow model $v_\theta$ here plays the role of a generative policy that, given state $s$ and noise $e$, outputs a velocity which can be used to produce an action. 

A straightforward way to integrate flow policies into offline RL is to replace the BC loss in the actor objective \eqref{equ:imitation} with the flow matching loss \eqref{eq:flow_policy_loss}. Compared to Gaussian or diffusion-based policies, flow policies offer a powerful yet efficient framework for modelling multi-modal action distributions.
% \textbf{\color{hiccup}Notational warning:}
% Note that \textit{state} $s$ is an important input of our model. But for simplicity, and to save space in our equations, we will continue to denote the pair $(s,a_t)$ simply by $a_t$.
% \end{comment}

\section{Methodology}\label{sec:method}
% \subsection{Motivation} 
Standard Gaussian policies generate actions directly from noise and offer efficient inference, but they lack the expressivity needed to model complex, multimodal action distributions. Flow-based generative policies improve expressivity through iterative sampling, yet their integration with Q-learning requires backpropagation through time (BPTT), making joint training unstable and inefficient. Distillation-based approaches bypass BPTT by employing a two-stage training process, which sacrifices model expressivity in the final one-step policy. 

This raises a key question: \emph{Can we design a generative policy that enables one-step action sampling while preserving the expressivity of flow-based models and supporting stable joint training with Q-functions?}
To answer this question, we introduce our idea of reformulating MeanFlow into an end-to-end policy function. 
% \subsection{Rethinking One-Step Generative Policies for Q-learning}

\begin{comment}
    Generally, offline reinforcement learning seeks to optimize a policy while respecting certain regularization constraints. This can be formulated as follows:
\begin{equation}
\underset{\pi \in \Pi}{\operatorname{argmax}} \mathbb{E}_{\substack{s, a \sim \mathcal{D}, \\ a^\pi \sim \pi_\theta(\cdot | x)}}[\underbrace{Q_\phi\left(s, a^\pi\right)}_{\text {Q Loss }}+\underbrace{\alpha \log \pi_\theta(a | s)}_{\text {BC Constraint }}]
\end{equation}
where the $Q_\phi$ function is trained by minimizing the Bellman error, while the  behavioural cloning loss can be replaced with a matching loss derived from generative flow models.  Directly combining the matching loss into Q-learning involves extensive backpropagation through time during training. And distillation-based approaches may require more complex, two-stage training procedures (e.g. propagating through teacher/student networks), over which error compounds. Naturally, this motivates us to leverage \textit{MeanFlow}, the powerful one-step generation framework, to enable native one-step action generation within the Q-learning framework.
\end{comment}

\subsection{Naive MeanFlow Policies}
A straightforward approach to enable one-step action generation is to directly adopt the original MeanFlow formulation, which estimates the average velocity and uses it to transform noise into actions. Specifically, MeanFlow generates an action as follows:
\begin{equation}
\begin{aligned}
 v_{\text{ave}} = u(e, b{=}0, t{=}1), {// \small \text{vel. est.}}  \\
\hat{a} = e - v_{\text{ave}}, {// \small \text{vel. integ.}}
\label{equ:meanflow_inference}   
\end{aligned}
\end{equation}
where $e \sim \mathcal{N}(0, I)$ is Gaussian noise, and $u$ denotes the learned average velocity field.

Although this formulation avoids iterative velocity integration over time and thus enables efficient inference, it suffers from a key limitation: the decoupled processes reduce controllability. The sampled noise $e$ often induces the generated actions to fall outside the valid action bounds (e.g., $[-1, 1]$), especially in early training when $u$ is inaccurate. Consequently, the generated actions frequently require post-hoc clipping, which breaks the alignment between the policy output and the actions used in the Bellman target. This mismatch destabilises Q-learning and can degrade overall performance, as shown in the results in Experiments and analysis in the Appendix.

\subsection{\textit{Reformulated MeanFlow}: Direct Noise-to-Action Mapping}  

The core limitation of the original MeanFlow formulation lies in its inference pipeline: noise \( \rightarrow \) velocity \( \rightarrow \) action. This decoupled design, which first predicts a velocity and then transforms noise into an action, often leads to instability—particularly during early training when the estimated velocity field is still inaccurate. 

A natural workaround is to reformulate MeanFlow into a single-step residual form, such as \( a = \epsilon - u(\epsilon, b, t) \), effectively merging velocity estimation and action generation into one forward pass. However, this naive reformulation performs poorly in practice: it fails to capture multimodal distributions, as shown in the toy experiments in the Appendix.

We propose a revised \emph{residual MeanFlow formulation} with a carefully constructed mapping
\begin{equation}
    g(a_t, b, t) = a_t - u(a_t, b, t),
    \label{equ:substitution of meanflow}
\end{equation}
which reduces to the original MeanFlow output when \( b = 0 \), \( t = 1 \), and \( a_1 = e \):
\begin{equation}
    g(e, b=0, t=1) = e - u(e, b=0, t=1).
\end{equation}

This revised residual transformation satisfies two key desiderata:  1) The mapping \( g_\theta \) retains the representational capacity of the original MeanFlow velocity function \( u_\theta \), as guaranteed by the \emph{Universal Approximation Theorem (UAT)}~\citep{leshno1993multilayer, tabuada2020universal}. 2) For inputs of the form given in Equation~\ref{equ:meanflow_inference}, the model enables well-bounded one-step outputs (typically within the [-1,1] range) even in the early stages of training through appropriate initialisation strategies, such as zero initialisation or small-variance Kaiming initialisation. A more detailed derivation of this formulation is provided in Appendix B.

This proposed formulation enables the network to learn a complete noise-to-action transformation directly, without depending on intermediate velocity estimation. Our method alleviates the early-stage instability caused by post-hoc clipping, while preserving the expressivity and efficiency inherent in the MeanFlow framework.

\subsection{Training Objective of Our One-Step Policy}
\label{sec:method_part1}

To train our residual mapping \( g_\theta(a_t, b, t) \), we leverage the \emph{MeanFlow Identity}, which defines the average velocity field \( u(a_t, b, t) \) as the difference between the instantaneous velocity \( v(a_t, t) \) and its time derivative \( \tfrac{d}{dt} u(z_t, b, t) \), as shown in Eq~\ref{eq:meanflow_diff}. Specifically, we follow~\citep{meanflow} to optimise \( g_\theta \) to satisfy this identity by minimising the following  objective:
\begin{equation}
\mathcal{L}_{\text{MFI}}(\theta) = \mathbb{E}\left\|g_\theta(a_t, b, t) - \operatorname{sg}(g_{\text{tgt}})\right\|_2^2,
\end{equation}
where $\mathcal{L}_{\text{MFI}}$ represents loss of MeanFlow Identity,  \( \operatorname{sg}(\cdot) \) denotes the stop-gradient operation. The core challenge lies in constructing an accurate regression target \( g_{\text{tgt}} \), which we derive as follows.

Recall from our reformulation that:
\begin{equation}
g(a_t, b, t) = a_t - u(a_t, b, t).
\end{equation}
We substitute \( u(a_t, b, t) \) using the velocity-based formulation from the MeanFlow Identity in Eq~\ref{eq:meanflow_diff}
which leads to:
\begin{equation}
\begin{aligned}
g(a_t, b, t)
&= a_t - [v(a_t, t) - (t - b) \frac{d}{dt}(a_t - g(a_t, b, t))] \\
&= a_t - [v(a_t, t) - (t - b) (v(a_t, t) - \frac{d}{dt}g(a_t, b, t))].
\end{aligned}
\label{eq:gt}
\end{equation}

To compute the total time derivative \( \frac{d}{dt}g(a_t, b, t) \), we apply the chain rule:
\begin{equation}
\frac{d}{dt}g(a_t, b, t) = \frac{d a_t}{dt} \cdot \partial_{a_t} g + \frac{d b}{dt} \cdot \partial_b g + \frac{dt}{dt} \cdot \partial_t g.
\end{equation}
Since \( \frac{d a_t}{dt} = v(a_t, t) \), \( \frac{d b}{dt} = 0 \), and \( \frac{dt}{dt} = 1 \), we simplify to:
\begin{equation}
\frac{d}{dt}g(a_t, b, t) = v(a_t, t) \cdot \partial_{a_t} g + \partial_t g.
\label{eq:full_derivatives}
\end{equation}
Substituting Equation~\ref{eq:full_derivatives} back into Equation~\ref{eq:gt}, finally get:
\begin{equation}
\begin{aligned}
g_{tgt} =\ & a_t + (t - b - 1) \cdot v(a_t, t) \\
& - (t - b) \cdot \left[v(a_t, t) \cdot \partial_{a_t} g + \partial_t g\right].
\end{aligned}
\label{eq:gn_final}
\end{equation}
It is worth noting that the formulation remains agnostic of any network parameterisation. The target relies solely on the instantaneous velocity \(v\) and the noisy data \(a_t\) as ground-truth signals, without requiring any integral computation. Specifically, recall that \(a_t = (1 - t)a + t e\); by default, the instantaneous velocity \(v(a_t, t)\) simplifies to \(e - a\). The gradient terms \(\partial_{a_t} g\) and \(\partial_t g\) are readily computed using the current policy network \(g_\theta\). Full algorithmic details are provided in Algorithm~\ref{alg:meanflow-ql}.

\begin{algorithm}[t]
\caption{ One-Step Generative Policy with Q-learning}
\label{alg:meanflow-ql}
\begin{algorithmic}
\footnotesize

\While{not converged}

\State Sample batch $\{(s, a, r, s')\} \sim \mathbb{D}$

\BeginBox[fill=gray!8]
\LComment{\color{hiccup} Train Critic $Q_\phi$}
\State $e \sim \mathbb{N}(0, I_d)$
\State $ \{a_1, \ldots, a_K\} = \texttt{vmap}(g_\theta)(s, e_{1:K}, b=0, t=1)$
\State $a' \gets \arg\max_{a_i} Q(s, a_i), \quad a_i \in \{a_1, \ldots, a_K\} $
\State Update {\color{hiccup}$\phi$} to minimize $\mathbb{E}[(Q_{\color{hiccup}\phi}(s, a) - r - \gamma Q_{\bar{\phi}}(s', a'))^2]$
\EndBox
\BeginBox[fill=white]
\BeginBox[fill=mask!60]
\LComment{\color{hiccup} Train One-Step MeanFlow Policy $\pi_\theta$}

\State $t,b \sim \mathbb{U}(0,1)$
\State $a_t = (1 - t)  a + t  e$
\State $v = e - a $             
% \State $g = \mu_{\color{mypurple}\theta}(s,x_t, b, t)  $           
\State $ g , dgdt = \texttt{jvp}(g_{\color{hiccup}\theta},(s, a_t, b, t), (s,v, 0, 1))$
\State $g_{\text{tgt}} = a_t + (t-b-1)v - (t-b)dgdt$
\State ${err} = g - [g_\mathrm{tgt}]_{\mathrm{sg}}$
\State $\mathcal{L}_{\text{MFI}}$ = $\operatorname{metric}({err})$  \Comment{Behaviour Cloning Loss}
\EndBox
% \BeginBox[fill=myblue!8]
% % \LComment{\color{mypurple} Train one-step policy $\pi_\omega$}
% \State $t_m,t_n \sim Uniform(0,1)$
% \State $x_b^{\left(t_m\right)} = x_{t_m}-\left(t_m-b\right) u\left(x_{t_m}, b, t_m\right)$
%     \State $x_b^{\left(t_n\right)} = x_{t_n}-\left(t_n-b\right) u\left(x_{t_n}, b, t_n\right)$
% % z_{t_1}-\left(t_1-r\right) u\left(z_{t_1}, r, t_1\right)
% % \State $z_r^{\left(t_2\right)}=z_{t_2}-\left(t_2-r\right) u\left(z_{t_2}, r, t_2\right)$
% \State $\mathcal{L}_C=metric\left(x_b^{\left(t_m\right)}, x_b^{\left(t_n\right)}\right)$ \Comment{Consistency Loss}
% \EndBox
% \State $\ell_C=metric\left(gn(s,)\right)$ \Comment{Consistency Loss}
\BeginBox[fill=gray!8]
\State $a^\pi \gets g_{\color{hiccup}\theta}(s, e)$
% \State $\ell_{C} = ReLU(-1-a^\pi) + ReLU(a^\pi-1) $ \Comment{Bound Loss}
\State $\mathcal{L}_Q = -Q_\phi(s, a^\pi) $ \Comment{Q-learning Loss}
\EndBox
% \BeginBox[fill=myorange!8]
% % \LComment{\color{mypurple} Train one-step policy $\pi_\omega$}
% \State $z \sim \gN(0, I_d)$
% \State $a^\pi \gets \mu_{\color{mypurple}\theta}(s, z)$
% \State $loss\_q = -Q_\phi(s, a^\pi)$ \Comment{Q-learning loss}
% \EndBox
\State Update {\color{hiccup}$\theta$} to minimize  $\mathcal{L}_Q + \alpha\mathcal{L}_{\text{MFI}} $
% + \beta\mathcal{L}_C 

\EndBox
\EndWhile
\Return One-step policy $g_\theta$

\BeginBox[fill=mask!60]
\vspace{5pt}
\Function{$\text{Policy}(s)$}{} \Comment{ One-Step Policy}
% \State $e = randn(action\_shape)$
\State $ \{a_1, \ldots, a_K\} = \texttt{vmap}(g_\theta)(s, e_{1:K}, b=0, t=1)$
\State $  a' = \arg\max_{a_i} Q(s, a_i), \quad a_i \in \{a_1, \ldots, a_K\} $
\State \Return $a'$
\EndFunction
\EndBox
\end{algorithmic}
\end{algorithm}

\subsection{Value-Guided Action Selection in Bellman Iteration}
While performing behaviour cloning, we also need to update the critic function through the Bellman equation iteratively. However, despite the efficiency gains brought by our one-step generative policy, its stochastic sampling of the input noise induces some action-sampling variance, which destabilises Bellman updates and may lead to divergence during critic training. To address this issue, we adopt a value-guided rejection sampling scheme for action selection. 

Furthermore, by utilising the \texttt{vmap} function in the \texttt{JAX} framework, we can sample these candidates in parallel and evaluate them with negligible computational overhead.
Formally, we define the value-guided behaviour policy sampling as
\[
\pi_\theta^K(a|s) \triangleq \underset{a \in \{a_1, \ldots, a_K \sim g_\theta(s, e, b=0, t=1)\}}{\operatorname{argmax}} Q(s, a),
\]
where $K$ actions are sampled simultaneously from the learned policy. The action with the highest Q-value is selected for the Bellman iteration as follows:
\begin{equation}\label{eq:bellman_loss}
\mathcal{L}_Q
= \mathbb{E}_{(s,a,r,s')\sim\mathcal{D}}\Bigl[
  \bigl(Q_\phi(s,a) - \bigl(r + \gamma\,Q_{\bar{\phi}}(x',a')\bigr)\bigr)^2
\Bigr],
\end{equation}
where $a' \sim \pi_\theta^K(a|s')$, ${\bar{\phi}}$ denotes a delayed target network to promote training stability.
We incorporate value-guided action sampling into the Bellman update without modifying the underlying policy model, effectively filtering out low-quality samples and focusing optimisation on more promising regions of the action space. Consequently, it enhances both training stability and final policy performance.

\subsection{Adaptive Behaviour-Cloning Coefficient}
Overall, the training objective for our one-step generative policy with Q-learning can be formalised as follows:
\begin{equation} 
\theta^\star
  \;=\;
  \underset{\theta}{\arg\max}\;
  \mathbb{E}_{\substack{(s,a)\sim\mathcal D \\ a^\pi\sim\pi_\theta(\cdot\mid s)}}
  \Bigl[
     \underbrace{Q_\phi\!\bigl(s,a^\pi\bigr)}_{\text{Q loss}}
     \;-\;
     \underbrace{\alpha\,\mathcal{L}_{\text{MFI}}}_{\text{BC loss}}
  \Bigr].
\label{eq:offline_rl_obj}
\end{equation}

Empirically, the choice of $\alpha$ is crucial, as it directly governs the trade-off between behaviour cloning and Q-learning. To adaptively balance their contributions, we dynamically adjust the behaviour cloning coefficient~$\alpha$ according to historic training dynamics. Concretely, we maintain a moving average $\overline{\mathcal{L}_Q}$ of the behaviour cloning loss over a sliding window and update $\alpha$ as follows:
\begin{equation}
    \alpha \leftarrow
  \begin{cases}
    1.2 \times \alpha, & \text{if } \mathcal{L}_Q > 5 \cdot \overline{\mathcal{L}_Q},\\[0pt]
    0.8 \times \alpha, & \text{if } \mathcal{L}_Q < 0.2 \cdot \overline{\mathcal{L}_Q},\\[0pt]
    \alpha, & \text{otherwise.}
  \end{cases}
\end{equation}

Despite differences, this strategy is inspired by the adaptive KL penalty introduced in PPO~\citep{schulman2017PPO} and ensures that the behaviour cloning regularisation remains neither dominant nor negligible during training. Thus, this stabilises training and results in more balanced policy updates. Empirical evaluations decided the introduced scaling factors.

\begin{table*}[t]
\caption{
\footnotesize
\textbf{Offline RL results.}
Our method achieves the best or near-best performance on most of the $\mathbf{73}$ diverse, challenging benchmark tasks.
The performances are averaged over $\mathbf{8}$ seeds ($\mathbf{4}$ seeds for pixel-based tasks),
but the cells without the $\pm$ sign indicate that the numbers are taken
from prior works~\citep{tarasov2023corl, idql, srpo}.
See Table \ref{table:offline_full} in Appendix for the full results.}
\label{table:offline}
\centering
\scalebox{0.76}{
\begin{threeparttable}
\begin{tabular}{lccccccccccc}
\toprule
\multicolumn{1}{c}{} & \multicolumn{3}{c}{\texttt{Gaussian Policies}} & \multicolumn{3}{c}{\texttt{Diffusion Policies}} & \multicolumn{5}{c}{\texttt{Flow Policies}} \\
\cmidrule(lr){2-4} \cmidrule(lr){5-7} \cmidrule(lr){8-12}
\texttt{Task Category} & \texttt{BC} & \texttt{IQL} & \texttt{ReBRAC} & \texttt{IDQL} & \texttt{SRPO} & \texttt{CAC} & \texttt{FAWAC} & \texttt{FBRAC} & \texttt{IFQL} & \texttt{FQL} & \texttt{Ours} \\
\midrule
\texttt{OGBench antmaze-large-singletask} ($\mathbf{5}$ tasks) & $11$ {\tiny $\pm 1$} & $53$ {\tiny $\pm 3$} & $\mathbf{81}$ {\tiny $\pm 5$} & $21$ {\tiny $\pm 5$} & $11$ {\tiny $\pm 4$} & $33$ {\tiny $\pm 4$} & $6$ {\tiny $\pm 1$} & $60$ {\tiny $\pm 6$} & $28$ {\tiny $\pm 5$} & $79$ {\tiny $\pm 3$} & $\mathbf{81}$ {\tiny $\pm 3$} \\
\rowcolor{mask!88}
\texttt{OGBench antmaze-giant-singletask} ($\mathbf{5}$ tasks) & $0$ {\tiny $\pm 0$} & $4$ {\tiny $\pm 1$} & $\mathbf{26}$ {\tiny $\pm 8$} & $0$ {\tiny $\pm 0$} & $0$ {\tiny $\pm 0$} & $0$ {\tiny $\pm 0$} & $0$ {\tiny $\pm 0$} & $4$ {\tiny $\pm 4$} & $3$ {\tiny $\pm 2$} & $9$ {\tiny $\pm 6$}  & $0$ {\tiny $\pm 0$} \\
\texttt{OGBench humanoidmaze-medium-singletask} ($\mathbf{5}$ tasks) & $2$ {\tiny $\pm 1$} & $33$ {\tiny $\pm 2$} & $22$ {\tiny $\pm 8$} & $1$ {\tiny $\pm 0$} & $1$ {\tiny $\pm 1$} & $53$ {\tiny $\pm 8$} & $19$ {\tiny $\pm 1$} & $38$ {\tiny $\pm 5$} & $60$ {\tiny $\pm 14$} & $58$ {\tiny $\pm 5$} & $\mathbf{62}$ {\tiny $\pm 1$} \\
\texttt{OGBench humanoidmaze-large-singletask} ($\mathbf{5}$ tasks) & $1$ {\tiny $\pm 0$} & $2$ {\tiny $\pm 1$} & $2$ {\tiny $\pm 1$} & $1$ {\tiny $\pm 0$} & $0$ {\tiny $\pm 0$} & $0$ {\tiny $\pm 0$} & $0$ {\tiny $\pm 0$} & $2$ {\tiny $\pm 0$} & $11$ {\tiny $\pm 2$} & $4$ {\tiny $\pm 2$} & $\mathbf{20}$ {\tiny $\pm 3$} \\
\texttt{OGBench antsoccer-arena-singletask} ($\mathbf{5}$ tasks) & $1$ {\tiny $\pm 0$} & $8$ {\tiny $\pm 2$} & $0$ {\tiny $\pm 0$} & $12$ {\tiny $\pm 4$} & $1$ {\tiny $\pm 0$} & $2$ {\tiny $\pm 4$} & $12$ {\tiny $\pm 0$} & $16$ {\tiny $\pm 1$} & $33$ {\tiny $\pm 6$} & $60$ {\tiny $\pm 2$} & $\mathbf{62}$ {\tiny $\pm 3$} \\
\texttt{OGBench cube-single-singletask} ($\mathbf{5}$ tasks) & $5$ {\tiny $\pm 1$} & $83$ {\tiny $\pm 3$} & $91$ {\tiny $\pm 2$} & $\mathbf{95}$ {\tiny $\pm 2$} & $80$ {\tiny $\pm 5$} & $85$ {\tiny $\pm 9$} & $81$ {\tiny $\pm 4$} & $79$ {\tiny $\pm 7$} & $79$ {\tiny $\pm 2$} & $\mathbf{96}$ {\tiny $\pm 1$} & $\mathbf{95}$ {\tiny $\pm 2$} \\
\rowcolor{mask!88}
\texttt{OGBench cube-double-singletask} ($\mathbf{5}$ tasks) & $2$ {\tiny $\pm 1$} & $7$ {\tiny $\pm 1$} & $12$ {\tiny $\pm 1$} & $15$ {\tiny $\pm 6$} & $2$ {\tiny $\pm 1$} & $6$ {\tiny $\pm 2$} & $5$ {\tiny $\pm 2$} & $15$ {\tiny $\pm 3$} & $14$ {\tiny $\pm 3$} & $\mathbf{29}$ {\tiny $\pm 2$} & $3$ {\tiny $\pm 2$} \\
\texttt{OGBench scene-singletask} ($\mathbf{5}$ tasks) & $5$ {\tiny $\pm 1$} & $28$ {\tiny $\pm 1$} & $41$ {\tiny $\pm 3$} & $46$ {\tiny $\pm 3$} & $20$ {\tiny $\pm 1$} & $40$ {\tiny $\pm 7$} & $30$ {\tiny $\pm 3$} & $45$ {\tiny $\pm 5$} & $30$ {\tiny $\pm 3$} & $56$ {\tiny $\pm 2$} & $\mathbf{60}$ {\tiny $\pm 1$} \\
\texttt{OGBench puzzle-3x3-singletask} ($\mathbf{5}$ tasks) & $2$ {\tiny $\pm 0$} & $9$ {\tiny $\pm 1$} & $21$ {\tiny $\pm 1$} & $10$ {\tiny $\pm 2$} & $18$ {\tiny $\pm 1$} & $19$ {\tiny $\pm 0$} & $6$ {\tiny $\pm 2$} & $14$ {\tiny $\pm 4$} & $19$ {\tiny $\pm 1$} & $30$ {\tiny $\pm 1$} & $\mathbf{66}$ {\tiny $\pm 8$} \\
\texttt{OGBench puzzle-4x4-singletask} ($\mathbf{5}$ tasks) & $0$ {\tiny $\pm 0$} & $7$ {\tiny $\pm 1$} & $14$ {\tiny $\pm 1$} & $29$ {\tiny $\pm 3$} & $10$ {\tiny $\pm 3$} & $15$ {\tiny $\pm 3$} & $1$ {\tiny $\pm 0$} & $13$ {\tiny $\pm 1$} & $25$ {\tiny $\pm 5$} & $17$ {\tiny $\pm 2$} & $\mathbf{40}$ {\tiny $\pm 6$} \\
\texttt{D4RL antmaze} ($\mathbf{6}$ tasks) & $17$ & $57$ & $78$ & $79$ & $74$ & $30$ {\tiny $\pm 3$} & $44$ {\tiny $\pm 3$} & $64$ {\tiny $\pm 7$} & $65$ {\tiny $\pm 7$} & $\mathbf{84}$ {\tiny $\pm 3$} & $\mathbf{83}$ {\tiny $\pm 2$} \\
\texttt{D4RL adroit} ($\mathbf{12}$ tasks) & $48$ & $53$ & $\mathbf{59}$ & $52$ {\tiny $\pm 1$} & $51$ {\tiny $\pm 1$} & $43$ {\tiny $\pm 2$} & $48$ {\tiny $\pm 1$} & $50$ {\tiny $\pm 2$} & $52$ {\tiny $\pm 1$} & $52$ {\tiny $\pm 1$} & $54$ {\tiny $\pm 3$} \\
\texttt{Visual manipulation} ($\mathbf{5}$ tasks) & - & $42$ {\tiny $\pm 4$} & $60$ {\tiny $\pm 2$} & - & - & - & - & $22$ {\tiny $\pm 2$} & $50$ {\tiny $\pm 5$} & $\mathbf{65}$ {\tiny $\pm 2$} & $55$ {\tiny $\pm 2$} \\
\bottomrule
\end{tabular}
\begin{tablenotes}
\item \textsuperscript{1} Due to the high computational cost of pixel-based tasks,
we selectively benchmark $5$ methods that achieve strong performance on state-based \texttt{OGBench} tasks.
\end{tablenotes}
\end{threeparttable}
}
\vspace{-10pt}
\end{table*}

\section{Experiments}
In this section, we present experimental evaluations of our one-step generative policy with Q-learning, comparing its performance against 10 baselines across 73 tasks. Additionally, we conduct analytical studies to help understand our method.
\subsection{Experimental Setup}
\paragraph{Benchmarks.} We evaluate our method on a diverse set of offline RL tasks spanning locomotion and manipulation domains. The primary benchmark is the OGBench suite \citep{fql}, which includes 50 state-based tasks and 5 visual-based tasks across 10 environments: antmaze-large, antmaze-giant, humanoidmaze-medium, humanoidmaze-large, antsoccer-arena, cube-single, cube-double, scene-play, puzzle-3$\times$3, and puzzle-4$\times$4. Additionally, we include 18 challenging tasks from the D4RL benchmark~\citep{d4rl}, covering antmaze and adroit manipulation scenarios. OGBench tasks feature semi-sparse rewards and complex  behavioural distributions, while D4RL tasks are known for their high-dimensional action spaces.
\paragraph{Baselines.} We compare our method against ten representative baselines spanning three policy classes. \emph{Gaussian policies} include  behavioural Cloning (BC)~\citep{BC}, Implicit Q-Learning (IQL)~\citep{iql}, and ReBRAC~\citep{ReBRAC}. \emph{Diffusion-based policies} include IDQL~\citep{idql}, SRPO~\citep{srpo}, and Consistency Actor-Critic (CAC)~\citep{ding2023consistency}. \emph{Flow-based policies} include FAWAC~\citep{fawac}, FBRAC~\citep{fbrac}, IFQL~\citep{ifql}, and Flow Q-Learning (FQL)~\citep{fql}. \emph{Specifically designed for online RL} include Cal-QL \cite{nakamoto2023cal} and PLPD \cite{RLPD}.
\paragraph{Evaluation metrics.} For offline RL, we assess the performance of the methods after a fixed number of gradient steps. Specifically, following \citep{tarasov2023corl} we avoid reporting the best performance across different evaluation epochs to prevent potential bias in the results. In our experiments, we employ 8 random seeds for state-based tasks and 4 seeds for pixel-based tasks. For the OGBench locomotion and manipulation tasks, performance is measured as the percentage of episodes that achieve the goal. For the D4RL Adroit tasks, we measure binary task success rates (in percentage) for antmaze and normalized returns for adroit, following the original evaluation scheme~\citep{d4rl}. We refer to Appendix D for the full training and evaluation details. 
\begin{table*}[t]
\centering
\caption{
\footnotesize
\textbf{Offline-to-online RL results. } Following FQL \cite{fql}, the results are obtained by first training the policy offline for 1M steps and subsequently performing online fine-tuning.}
\label{table:offlinetoonline}
\scalebox{0.72}{
\begin{tabular}{lccccccccc}
\toprule
\texttt{Task} & \texttt{IQL} & \texttt{ReBRAC} & \texttt{Cal-QL} & \texttt{RLPD} & \texttt{IFQL} & \texttt{FQL} & \texttt{Ours} \\
\midrule
\rowcolor{mask!88}
\texttt{antmaze-giant-navigate} &
$2$ {\tiny $\pm 1$} $\to$ $3$ {\tiny $\pm 1$} & $6$ {\tiny $\pm 5$} $\to$ $34$ {\tiny $\pm 4$} & $0$ {\tiny $\pm 0$} $\to$ $0$ {\tiny $\pm 0$} & 
$0$ {\tiny $\pm 0$} $\to$ $48$ {\tiny $\pm 13$} & $0$ {\tiny $\pm 0$} $\to$ $69$ {\tiny $\pm 12$} &
$4$ {\tiny $\pm 2$} $\to$ $9$ {\tiny $\pm 3$} &
$0$ {\tiny $\pm 0$} $\to$ $\mathbf{82}$ {\tiny $\pm 5$} \\
\texttt{humanoidmaze-medium-navigate} &
$21$ {\tiny $\pm 13$} $\to$ $16$ {\tiny $\pm 8$} &
$16$ {\tiny $\pm 20$} $\to$ $1$ {\tiny $\pm 1$} &
$0$ {\tiny $\pm 0$} $\to$ $0$ {\tiny $\pm 0$} &
$0$ {\tiny $\pm 0$} $\to$ $8$ {\tiny $\pm 10$} &
$56$ {\tiny $\pm 35$} $\to$ ${82}$ {\tiny $\pm 20$} &
$12$ {\tiny $\pm 7$} $\to$ $22$ {\tiny $\pm 12$} &
$62$ {\tiny $\pm 1$} $\to$ $\mathbf{100}$ {\tiny $\pm 1$} \\
\texttt{antsoccer-arena-navigate} &
$2$ {\tiny $\pm 1$} $\to$ $0$ {\tiny $\pm 0$} &
$0$ {\tiny $\pm 0$} $\to$ $0$ {\tiny $\pm 0$} &
$0$ {\tiny $\pm 0$} $\to$ $0$ {\tiny $\pm 0$} &
$0$ {\tiny $\pm 0$} $\to$ $0$ {\tiny $\pm 0$} &
$26$ {\tiny $\pm 15$} $\to$ $39$ {\tiny $\pm 10$} &
$28$ {\tiny $\pm 8$} $\to$ $\mathbf{86}$ {\tiny $\pm 5$} &
$62$ {\tiny $\pm 3$} $\to$ $\mathbf{87}$ {\tiny $\pm 5$} \\
\rowcolor{mask!88}
\texttt{cube-double-play} &
$0$ {\tiny $\pm 1$} $\to$ $0$ {\tiny $\pm 0$} &
$6$ {\tiny $\pm 5$} $\to$ $28$ {\tiny $\pm 28$} &
$0$ {\tiny $\pm 0$} $\to$ $0$ {\tiny $\pm 0$} &
$0$ {\tiny $\pm 0$} $\to$ $0$ {\tiny $\pm 0$} &
$12$ {\tiny $\pm 9$} $\to$ $40$ {\tiny $\pm 5$} &
$40$ {\tiny $\pm 11$} $\to$ $\mathbf{92}$ {\tiny $\pm 3$} &
$3$ {\tiny $\pm 2$} $\to$ $\mathbf{95}$ {\tiny $\pm 2$} \\
\texttt{scene-play} &
$14$ {\tiny $\pm 11$} $\to$ $10$ {\tiny $\pm 9$} &
$55$ {\tiny $\pm 10$} $\to$ $\mathbf{100}$ {\tiny $\pm 0$} &
$1$ {\tiny $\pm 2$} $\to$ $50$ {\tiny $\pm 53$} &
$0$ {\tiny $\pm 0$} $\to$ $\mathbf{100}$ {\tiny $\pm 0$} &
$0$ {\tiny $\pm 1$} $\to$ $60$ {\tiny $\pm 39$} &
$82$ {\tiny $\pm 11$} $\to$ $\mathbf{100}$ {\tiny $\pm 1$} &
$60$ {\tiny $\pm 1$} $\to$ $\mathbf{100}$ {\tiny $\pm 1$} \\
\texttt{puzzle-4x4-play} &
$5$ {\tiny $\pm 2$} $\to$ $1$ {\tiny $\pm 1$} &
$8$ {\tiny $\pm 4$} $\to$ $14$ {\tiny $\pm 35$} &
$0$ {\tiny $\pm 0$} $\to$ $0$ {\tiny $\pm 0$} &
$0$ {\tiny $\pm 0$} $\to$ $\mathbf{100}$ {\tiny $\pm 1$} &
$23$ {\tiny $\pm 6$} $\to$ $19$ {\tiny $\pm 33$} &
$8$ {\tiny $\pm 3$} $\to$ $38$ {\tiny $\pm 52$} &
$40$ {\tiny $\pm 6$} $\to$ $\mathbf{100}$ {\tiny $\pm 1$} \\
\midrule
\texttt{antmaze-umaze-v2} &
$77$ $\to$ $\mathbf{96}$ &
$98$ $\to$ $75$ &
$77$ $\to$ $\mathbf{100}$ &
$0$ {\tiny $\pm 0$} $\to$ $\mathbf{98}$ {\tiny $\pm 3$} &
$94$ {\tiny $\pm 5$} $\to$ $\mathbf{96}$ {\tiny $\pm 2$} &
$97$ {\tiny $\pm 2$} $\to$ $\mathbf{99}$ {\tiny $\pm 1$} &
$98$ {\tiny $\pm 1$} $\to$ $\mathbf{99}$ {\tiny $\pm 1$} \\
\texttt{antmaze-umaze-diverse-v2} &
$60$ $\to$ $64$ &
$74$ $\to$ $\mathbf{98}$ &
$32$ $\to$ $\mathbf{98}$ &
$0$ {\tiny $\pm 0$} $\to$ $94$ {\tiny $\pm 5$} &
$69$ {\tiny $\pm 20$} $\to$ $93$ {\tiny $\pm 5$} &
$79$ {\tiny $\pm 16$} $\to$ $\mathbf{100}$ {\tiny $\pm 1$} &
$79$ {\tiny $\pm 2$} $\to$ $\mathbf{100}$ {\tiny $\pm 1$} \\
\texttt{antmaze-medium-play-v2} &
$72$ $\to$ $90$ &
$88$ $\to$ $\mathbf{98}$ &
$72$ $\to$ $\mathbf{99}$ &
$0$ {\tiny $\pm 0$} $\to$ $\mathbf{98}$ {\tiny $\pm 2$} &
$52$ {\tiny $\pm 19$} $\to$ $93$ {\tiny $\pm 2$} &
$77$ {\tiny $\pm 7$} $\to$ $\mathbf{97}$ {\tiny $\pm 2$} &
$86$ {\tiny $\pm 2$} $\to$ $\mathbf{99}$ {\tiny $\pm 1$} \\
\texttt{antmaze-medium-diverse-v2} &
$64$ $\to$ $92$ &
$85$ $\to$ $\mathbf{99}$ &
$62$ $\to$ $\mathbf{98}$ &
$0$ {\tiny $\pm 0$} $\to$ $\mathbf{97}$ {\tiny $\pm 2$} &
$44$ {\tiny $\pm 26$} $\to$ $89$ {\tiny $\pm 4$} &
$55$ {\tiny $\pm 19$} $\to$ $\mathbf{97}$ {\tiny $\pm 3$} &
$78$ {\tiny $\pm 2$} $\to$ $\mathbf{99}$ {\tiny $\pm 2$} \\
\texttt{antmaze-large-play-v2} &
$38$ $\to$ $64$ &
$68$ $\to$ $32$ &
$32$ $\to$ $\mathbf{97}$ &
$0$ {\tiny $\pm 0$} $\to$ $\mathbf{93}$ {\tiny $\pm 5$} &
$64$ {\tiny $\pm 14$} $\to$ $80$ {\tiny $\pm 5$} &
$66$ {\tiny $\pm 40$} $\to$ $84$ {\tiny $\pm 30$} &
$80$ {\tiny $\pm 3$} $\to$ $\mathbf{94}$ {\tiny $\pm 4$} \\
\texttt{antmaze-large-diverse-v2} &
$27$ $\to$ $64$ &
$67$ $\to$ $72$ &
$44$ $\to$ $\mathbf{92}$ &
$0$ {\tiny $\pm 0$} $\to$ $\mathbf{94}$ {\tiny $\pm 3$} &
$69$ {\tiny $\pm 6$} $\to$ $86$ {\tiny $\pm 5$} &
$75$ {\tiny $\pm 24$} $\to$ $\mathbf{94}$ {\tiny $\pm 3$} &
$76$ {\tiny $\pm 1$} $\to$ $\mathbf{96}$ {\tiny $\pm 3$} \\
\midrule
\texttt{pen-cloned-v1} &
$84$ $\to$ $102$ &
$74$ $\to$ $138$ &
$-3$ $\to$ $-3$ &
$3$ {\tiny $\pm 2$} $\to$ $120$ {\tiny $\pm 10$} &
$77$ {\tiny $\pm 7$} $\to$ $107$ {\tiny $\pm 10$} &
$53$ {\tiny $\pm 14$} $\to$ $\mathbf{149}$ {\tiny $\pm 6$} &
$79$ {\tiny $\pm 3$} $\to$ $\mathbf{151}$ {\tiny $\pm 7$} \\
\texttt{door-cloned-v1} &
$1$ $\to$ $20$ &
$0$ $\to$ $\mathbf{102}$ &
$0$ $\to$ $0$ &
$0$ {\tiny $\pm 0$} $\to$ $\mathbf{102}$ {\tiny $\pm 7$} &
$3$ {\tiny $\pm 2$} $\to$ $50$ {\tiny $\pm 15$} &
$0$ {\tiny $\pm 0$} $\to$ $\mathbf{102}$ {\tiny $\pm 5$} &
$3$ {\tiny $\pm 1$} $\to$ ${96}$ {\tiny $\pm 4$} \\
\texttt{hammer-cloned-v1} &
$1$ $\to$ $57$ &
$7$ $\to$ ${125}$ &
$0$ $\to$ $0$ &
$0$ {\tiny $\pm 0$} $\to$ ${128}$ {\tiny $\pm 29$} &
$4$ {\tiny $\pm 2$} $\to$ $60$ {\tiny $\pm 14$} &
$0$ {\tiny $\pm 0$} $\to$ ${127}$ {\tiny $\pm 17$} &
$10$ {\tiny $\pm 5$} $\to$ $\mathbf{132}$ {\tiny $\pm 11$} \\
\texttt{relocate-cloned-v1} &
$0$ $\to$ $0$ &
$1$ $\to$ $7$ &
$0$ $\to$ $0$ &
$0$ {\tiny $\pm 0$} $\to$ $2$ {\tiny $\pm 2$} &
$0$ {\tiny $\pm 0$} $\to$ $5$ {\tiny $\pm 3$} &
$0$ {\tiny $\pm 1$} $\to$ $\mathbf{62}$ {\tiny $\pm 8$} & 
$1$ {\tiny $\pm 1$} $\to$ ${19}$ {\tiny $\pm 8$} \\
\bottomrule
\end{tabular}}
% \begin{tablenotes}
% \scriptsize
% \item \textsuperscript{1} 
% Default \texttt{OG-Bench} tasks (marked with * in Appendix Table~\ref{table:offline_full}) are used, with the common suffix \texttt{singletask-task-v0} omitted for brevity.
% \end{tablenotes}
\vspace{-15pt}
\end{table*}
\subsection{Comparisons with SOTA}
\paragraph{Offline RL performance.}
Following FQL \cite{fql}, we evaluate our method on the offline subset of the OGBench and D4RL benchmarks, which include diverse tasks with varying dynamics and action dimensionalities. Table~\ref{table:offline} shows that our approach achieves strong performance across a wide range of tasks, particularly in environments such as OGBench {antmaze}, {humanoidmaze}, {puzzle}, and {visual manipulation}, which likely demand expressive and multimodal policy behaviours due to complex dynamics.

Compared to flow-based baselines such as FQL, IFQL, and FBRAC, our method either matches or surpasses their performance while maintaining a simpler one-stage training pipeline and a one-step inference mechanism. Although performance on particularly sparse environments remains modest—such as in {giant maze navigation}—the results are consistent with broader trends observed across baselines, indicating inherent task difficulty rather than model-specific limitations. (We additionally add an online-finetuning extension on this task as shown in Table~\ref{table:offlinetoonline} and show that although initial offline performance was limited, our method achieved strong improvements after online finetuning, suggesting the bottleneck may lie in the offline dataset.) Overall, the results indicate that our one-step generative policy serves as a viable and competitive approach for offline reinforcement learning.

% {\color{hiccup} add online result of large-scale maze navigation 0730}

\paragraph{Offline-to-online transfer.}
We further evaluate our method in offline-to-online settings following~\citep{fql} (while adding {antmaze-giant-navigate}), where policies are first pre-trained on static offline datasets and then fine-tuned through online interaction. As shown in Table~\ref{table:offlinetoonline}, our method consistently improves after fine-tuning and achieves strong final performance across a diverse set of tasks. More notably, although the offline performances are not ideal on {antmaze-giant-navigate} and {cube-double-play}, our method gains a significant performance boost in the online adaptation.

These results highlight the robustness and sample efficiency of our approach when transitioning from offline training to online adaptation. Notably, these improvements are obtained without any modifications to the model architecture or training objective, allowing the policy to interact with the environment in a more reward-sensitive and direct manner. This unified design facilitates more effective exploration, which is critical for rapid adaptation from offline to online. Overall, the strong performance of our generative policy in both static and interactive settings highlights its potential for real-world deployment in evolving environments.

\subsection{Further Analysis}
\begin{figure}[t]
% \vspace{-0.1cm}
  \centering{%
    \includegraphics[width=0.48\textwidth]{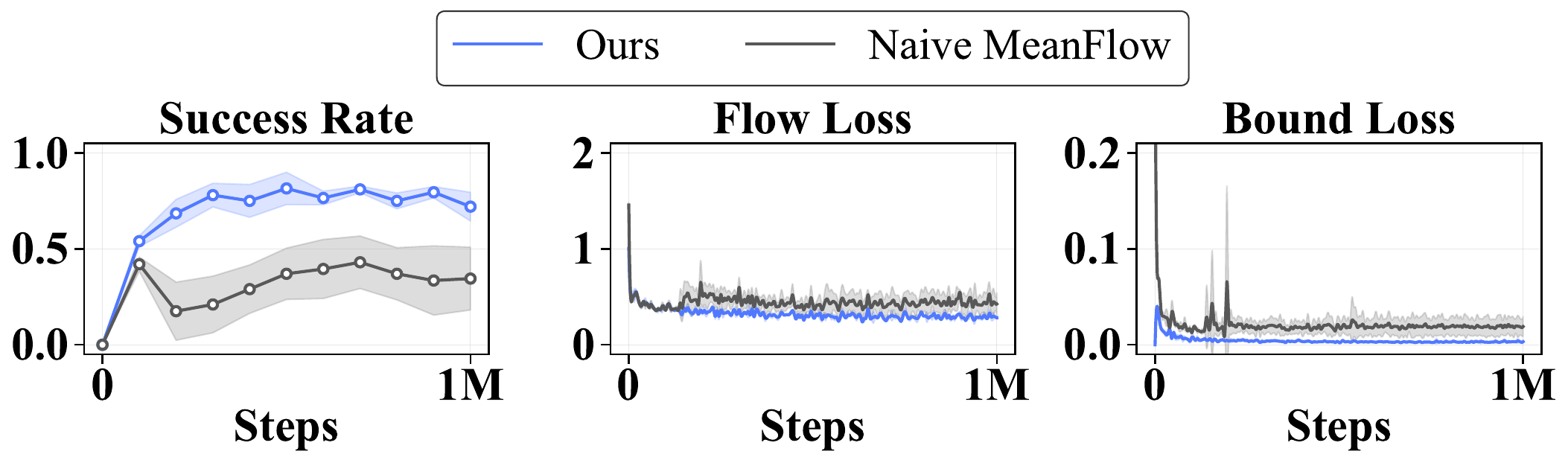}%
  }
  \caption{Comparison of learning curves. Naive MeanFlow exhibits significantly larger bound losses in the early training stages. This instability directly undermines convergence of its flow loss, leading to a substantial performance gap compared to ours on the antsoccer-arena-singletask.}
  % \vspace{-0.3cm}
  \label{fig:meanflow_vs_ours}
\end{figure}

% We show some further analysis to understand our method.
\paragraph{Our method vs naive MeanFlow.} 
To evaluate the effectiveness of our proposed reformulation strategy, we compare it against the original MeanFlow framework under the same environment and with well-tuned hyperparameters. As shown in Fig.~\ref{fig:meanflow_vs_ours}, the naive MeanFlow approach demonstrates performance (left) and flow loss (mid.) metrics that are nearly indistinguishable from ours during the initial stages of training. 
However, the bound loss (right) metric---defined as the mean absolute magnitude of action values outside \([-1, 1]\)---is substantially higher for the naive MeanFlow approach. This indicates it predicts actions not in the valid range. Thus, its performance then degrades significantly due to inaccurate Q-value estimates from the critic, which impair the training of the generative flow model and result in a low success rate.

% \paragraph{Hyperparameter Sensitivity Analysis}
% We conduct a sensitivity analysis to better understand the impact of key hyperparameters in our method. Specifically, we investigate: (i) the weight $\alpha$ used to balance the Q-learning objective and behaviour cloning loss; (ii) the number of action samples $K$ (denoted as \texttt{num\_candidates}) used in value-guided selection; and (iii) the choice of time step $t$ in the flow-based policy learning.

\paragraph{Analysis of coefficient $\alpha$ in policy learning.}

\begin{figure}[t]
% \vspace{-0.1cm}
  \centering
  \includegraphics[width=.48\textwidth]{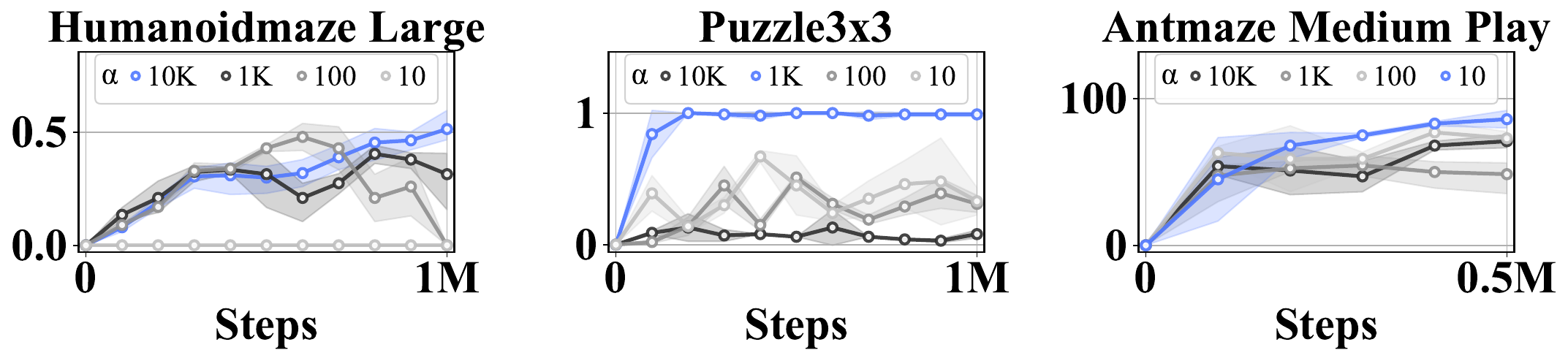} % 左 下 右 上
  \caption{Sensitivity to the behaviour cloning coefficient $\alpha$. Effect of different values of $\alpha$ on offline RL performance.}
  \label{fig:sensitivity_alpha}
\end{figure}

The hyperparameter $\alpha$ controls the relative strength of the behaviour cloning term versus the Q-function objective in the training loss. As shown in Fig.~\ref{fig:sensitivity_alpha}, overly small values of $\alpha$ lead to unstable updates due to over-reliance on potentially noisy Q estimates, while overly large values lead to simply behaviour cloning learning. Specifically, although we employ an adaptive $\alpha$ tuning mechanism in our main experiments, the model's final performance still exhibits sensitivity to initial $\alpha$ values, especially in sparse-reward tasks, e.g. Puzzle3x3. This highlights the importance of a balanced trade-off, and motivates future research on more robust adaptive schemes.

\begin{figure}[t]
  \centering
  \includegraphics[width=.48\textwidth]{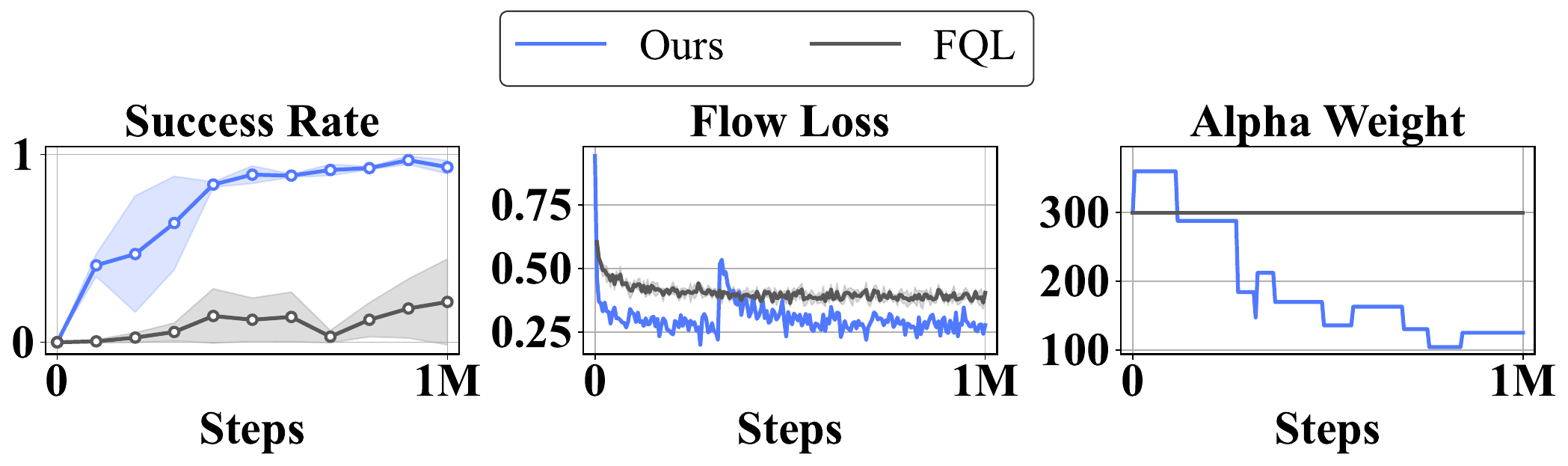} % 左 下 右 上
  \caption{Training dynamics under our framework in humanoidmaze-large-navigate-singletask. Despite potential sensitivity, the adaptive coefficient~$\alpha$ adjusts dynamically and evolves smoothly, contributing to stable and robust policy learning throughout the training process.}
  \label{fig:alpha_weight}
\end{figure}

\begin{figure}[t]
  \centering
  \includegraphics[width=.48\textwidth]{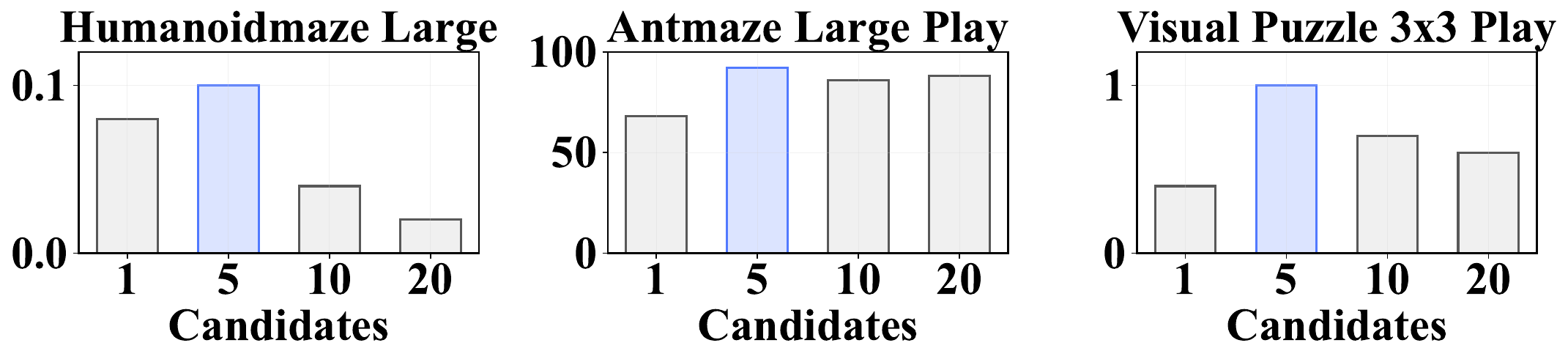} % 左 下 右 上
  \caption{Performance effect of candidates num. $K$ used in value-guided rejection sampling on three representative tasks.}
  \label{fig:sensitivity_candidates}
\end{figure}

To further examine the sensitivity of our method to the coefficient $\alpha$, we analyze its evolution during training under the proposed adaptive adjustment strategy. As shown in Fig.~\ref{fig:alpha_weight}, we can see $\alpha$ evolves smoothly and remains well-regulated throughout learning based on our adaptive mechanism. This dynamic adaptation effectively balances the trade-off between behaviour cloning and value maximisation, thereby enhancing the robustness of our method across different training stages and contributing to our superior performance over FQL~\citep{fql}.
\paragraph{Effect of candidate size $K$ in value-guided selection.}
In our value-guided rejection sampling, hyperparameter $K$ controls the number of candidate actions generated and scored by the Q-function for each state. As shown in Fig.~\ref{fig:sensitivity_candidates}, we observe that increasing $K$ from 1 to 5 leads to a substantial performance gain across various tasks. 
However, setting $K$ too large (e.g., 20) may impair performance by reducing stochasticity and exploration diversity, as overly greedy selection can cause the critic function to focus on a narrow set of high-value actions, thereby limiting  behavioural variability. In practice, we find that $K=5$ offers the best trade-off, preserving policy exploration and maintaining computational efficiency.

\section{Related Works}

\paragraph{Generalised behaviour cloning.}  
Behaviour cloning (BC) casts policy learning as a supervised task over expert state-action pairs~\citep{hussein2017imitation,block2023provable}. Generalised variants, such as advantage-weighted imitation~\citep{peters2007reinforcement}, steer learning toward high-return trajectories by weighting actions by their cumulative rewards. Recent methods extend this idea in different ways: SfBC~\citep{chen2022offline} and IDQL~\citep{idql} pretrain a  behaviour policy and critic to support importance sampling; QGPO~\citep{lu2023contrastive} distills the critic into an energy-based model; GMPO and GMPG~\citep{zhang2024generative_rl} apply weighted policy gradient updates directly without pretraining; and QIPO~\citep{fbrac} incorporates Q-value-derived energy into flow-matching updates. Despite these advances, recent studies highlight their inefficiency and limited performance in complex tasks~\citep{park2024value}.

\paragraph{Generative policy with Q-learning.} 
Incorporating Q-functions into generative policies, enabling direct reward-driven optimisation, offers a more principled alternative to behaviour cloning. These works often optimise policies with Q-learning by differentiating through the generative model's reverse process. Methods such as DQL~\citep{ifql}, Consistency-AC~\citep{ding2023consistency}, and DACER~\citep{wang2024diffusion} perform policy improvement via reverse diffusion, but suffer from high computational cost due to repeated denoising and unrolled optimisation.

To alleviate this, one-step generative policies have recently been proposed to avoid iterative sampling~\citep{chen2024deep_survey,consistency_policy,koirala2025flow}. Existing approaches follow two main directions: distillation-based methods that compress multi-step teachers into one-step student policies~\citep{srpo,fql}, and algorithmic innovations that enable direct action generation, such as consistency~\citep{song2023consistency} and shortcut~\citep{shortcut} models. Among them, FQL~\citep{fql} distils a flow-based policy into a one-step model optimised with Q-learning, while SORL~\citep{fql_shortcut} leverages shortcut frameworks~\citep{shortcut} to flexibly scale training and inference. However, both methods either rely on multi-stage pipelines or require inference-time scaling to achieve strong performance, thus falling short of a fully effective one-step policy learning framework.

\section{Conclusion}
We propose a one-step generative framework for offline RL that enables expressive and stable policy learning without relying on multi-step integration or policy distillation. Through a residual reformulation of MeanFlow, our method supports direct Q-learning with multimodal policies via a single-stage training pipeline. To further improve performance and robustness, we incorporate value-guided rejection sampling and adaptive behaviour cloning regularisation.
Experiment results show our method outperforming or matching prior work on most of the 73 tasks and reaching state-of-the-art performance on nearly all offline-to-online transfers, highlighting robust generalisation and adaptation.
Overall, this work bridges the gap between one-step generative modelling and value-based reinforcement learning, advancing the development of more expressive and scalable offline policy learning.
% \clearpage

\section{Acknowledgments}
This work was supported in part by the National Natural Science Foundation of China under Grant No. 62206305.
Zeyuan Wang completed this work during the author's visiting study at the Shanghai Innovation Institute, whose support are gratefully acknowledged.
Prof. Yanwei Fu was partly supported by the project from the Ministry of Science and Higher Education of the Republic of Kazakhstan, No. BR24992975, 'Development of a digital twin of a food processing enterprise using artificial intelligence and IIoT technologies' (2024–2026).

\bibliography{reference}

\begin{thebibliography}{53}
\providecommand{\natexlab}[1]{#1}

\bibitem[{Adam et~al.(2014)}]{kingma2014adam}
Adam, K. D. B.~J.; et~al. 2014.
\newblock A method for stochastic optimization.
\newblock \emph{arXiv preprint arXiv:1412.6980}, 1412(6).

\bibitem[{Ball et~al.(2023)Ball, Smith, Kostrikov, and Levine}]{RLPD}
Ball, P.~J.; Smith, L.; Kostrikov, I.; and Levine, S. 2023.
\newblock Efficient online reinforcement learning with offline data.
\newblock In \emph{International Conference on Machine Learning (ICML)}.

\bibitem[{Block et~al.(2023)Block, Jadbabaie, Pfrommer, Simchowitz, and Tedrake}]{block2023provable}
Block, A.; Jadbabaie, A.; Pfrommer, D.; Simchowitz, M.; and Tedrake, R. 2023.
\newblock Provable guarantees for generative behavior cloning: Bridging low-level stability and high-level behavior.
\newblock \emph{Advances in Neural Information Processing Systems (NeurIPS)}.

\bibitem[{Bradbury et~al.(2018)Bradbury, Frostig, Hawkins, Johnson, Leary, Maclaurin, Necula, Paszke, VanderPlas, Wanderman-Milne et~al.}]{bradbury2018jax}
Bradbury, J.; Frostig, R.; Hawkins, P.; Johnson, M.~J.; Leary, C.; Maclaurin, D.; Necula, G.; Paszke, A.; VanderPlas, J.; Wanderman-Milne, S.; et~al. 2018.
\newblock JAX: composable transformations of Python+ NumPy programs.

\bibitem[{Chen et~al.(2023{\natexlab{a}})Chen, Lu, Wang, Su, and Zhu}]{srpo}
Chen, H.; Lu, C.; Wang, Z.; Su, H.; and Zhu, J. 2023{\natexlab{a}}.
\newblock Score regularized policy optimization through diffusion behavior.
\newblock \emph{arXiv preprint arXiv:2310.07297}.

\bibitem[{Chen et~al.(2023{\natexlab{b}})Chen, Lu, Ying, Su, and Zhu}]{chen2022offline}
Chen, H.; Lu, C.; Ying, C.; Su, H.; and Zhu, J. 2023{\natexlab{b}}.
\newblock Offline Reinforcement Learning via High-Fidelity Generative Behavior Modeling.
\newblock In \emph{The Eleventh International Conference on Learning Representations (ICLR)}.

\bibitem[{Chen et~al.(2024)Chen, Ganguly, Xu, Mei, Lan, and Aggarwal}]{chen2024deep_survey}
Chen, J.; Ganguly, B.; Xu, Y.; Mei, Y.; Lan, T.; and Aggarwal, V. 2024.
\newblock Deep generative models for offline policy learning: Tutorial, survey, and perspectives on future directions.
\newblock \emph{arXiv preprint arXiv:2402.13777}.

\bibitem[{Chen, Wang, and Zhou(2024)}]{chen2024diffusion}
Chen, T.; Wang, Z.; and Zhou, M. 2024.
\newblock Diffusion policies creating a trust region for offline reinforcement learning.
\newblock \emph{Advances in Neural Information Processing Systems (NeurIPS)}.

\bibitem[{Chi et~al.(2023)Chi, Xu, Feng, Cousineau, Du, Burchfiel, Tedrake, and Song}]{chi2023diffusion}
Chi, C.; Xu, Z.; Feng, S.; Cousineau, E.; Du, Y.; Burchfiel, B.; Tedrake, R.; and Song, S. 2023.
\newblock Diffusion policy: Visuomotor policy learning via action diffusion.
\newblock \emph{The International Journal of Robotics Research}.

\bibitem[{Ding and Jin(2024)}]{ding2023consistency}
Ding, Z.; and Jin, C. 2024.
\newblock Consistency models as a rich and efficient policy class for reinforcement learning.
\newblock In \emph{12th International Conference on Learning Representations (ICLR)}.

\bibitem[{Espinosa-Dice et~al.(2025)Espinosa-Dice, Zhang, Chen, Guo, Oertell, Swamy, Brantley, and Sun}]{fql_shortcut}
Espinosa-Dice, N.; Zhang, Y.; Chen, Y.; Guo, B.; Oertell, O.; Swamy, G.; Brantley, K.; and Sun, W. 2025.
\newblock Scaling Offline RL via Efficient and Expressive Shortcut Models.
\newblock \emph{arXiv preprint arXiv:2505.22866}.

\bibitem[{Esser et~al.(2024)Esser, Kulal, Blattmann, Entezari, M{\"u}ller, Saini, Levi, Lorenz, Sauer, Boesel et~al.}]{esser2024scaling}
Esser, P.; Kulal, S.; Blattmann, A.; Entezari, R.; M{\"u}ller, J.; Saini, H.; Levi, Y.; Lorenz, D.; Sauer, A.; Boesel, F.; et~al. 2024.
\newblock Scaling rectified flow transformers for high-resolution image synthesis.
\newblock In \emph{Forty-first International conference on machine learning (ICML)}.

\bibitem[{Frans et~al.(2025)Frans, Hafner, Levine, and Abbeel}]{shortcut}
Frans, K.; Hafner, D.; Levine, S.; and Abbeel, P. 2025.
\newblock One Step Diffusion via Shortcut Models.
\newblock In \emph{The Thirteenth International Conference on Learning Representations (ICLR)}.

\bibitem[{Fu et~al.(2020)Fu, Kumar, Nachum, Tucker, and Levine}]{d4rl}
Fu, J.; Kumar, A.; Nachum, O.; Tucker, G.; and Levine, S. 2020.
\newblock D4rl: Datasets for deep data-driven reinforcement learning.
\newblock \emph{arXiv preprint arXiv:2004.07219}.

\bibitem[{Geng et~al.(2025)Geng, Deng, Bai, Kolter, and He}]{meanflow}
Geng, Z.; Deng, M.; Bai, X.; Kolter, J.~Z.; and He, K. 2025.
\newblock Mean flows for one-step generative modeling.
\newblock \emph{arXiv preprint arXiv:2505.13447}.

\bibitem[{Hafner and Riedmiller(2011)}]{offlinerlHafner}
Hafner, R.; and Riedmiller, M. 2011.
\newblock Reinforcement learning in feedback control.
\newblock \emph{Machine Learning}.

\bibitem[{Hansen-Estruch et~al.(2023)Hansen-Estruch, Kostrikov, Janner, Kuba, and Levine}]{idql}
Hansen-Estruch, P.; Kostrikov, I.; Janner, M.; Kuba, J.~G.; and Levine, S. 2023.
\newblock Idql: Implicit q-learning as an actor-critic method with diffusion policies.
\newblock \emph{arXiv preprint arXiv:2304.10573}.

\bibitem[{Hussein et~al.(2017)Hussein, Gaber, Elyan, and Jayne}]{hussein2017imitation}
Hussein, A.; Gaber, M.~M.; Elyan, E.; and Jayne, C. 2017.
\newblock Imitation learning: A survey of learning methods.
\newblock \emph{ACM Computing Surveys (CSUR)}.

\bibitem[{Jin et~al.(2025)Jin, Sun, Li, Xu, Xu, Jiang, Zhuang, Huang, Song, MU, and Lin}]{jin2024pyramidal}
Jin, Y.; Sun, Z.; Li, N.; Xu, K.; Xu, K.; Jiang, H.; Zhuang, N.; Huang, Q.; Song, Y.; MU, Y.; and Lin, Z. 2025.
\newblock Pyramidal Flow Matching for Efficient Video Generative Modeling.
\newblock In \emph{The Thirteenth International Conference on Learning Representations (ICLR)}.

\bibitem[{Koirala and Fleming(2025)}]{koirala2025flow}
Koirala, P.; and Fleming, C. 2025.
\newblock Flow-Based Single-Step Completion for Efficient and Expressive Policy Learning.
\newblock \emph{arXiv preprint arXiv:2506.21427}.

\bibitem[{Kostrikov, Nair, and Levine(2021)}]{iql}
Kostrikov, I.; Nair, A.; and Levine, S. 2021.
\newblock Offline Reinforcement Learning with Implicit Q-Learning.
\newblock In \emph{Deep RL Workshop NeurIPS 2021}.

\bibitem[{Leshno et~al.(1993)Leshno, Lin, Pinkus, and Schocken}]{leshno1993multilayer}
Leshno, M.; Lin, V.~Y.; Pinkus, A.; and Schocken, S. 1993.
\newblock Multilayer feedforward networks with a nonpolynomial activation function can approximate any function.
\newblock \emph{Neural networks}.

\bibitem[{Lipman et~al.(2023)Lipman, Chen, Ben-Hamu, Nickel, and Le}]{lipman2023flow}
Lipman, Y.; Chen, R.~T.; Ben-Hamu, H.; Nickel, M.; and Le, M. 2023.
\newblock Flow Matching for Generative Modeling.
\newblock In \emph{11th International Conference on Learning Representations (ICLR)}.

\bibitem[{Liu et~al.(2024)Liu, Le, Vyas, Shi, Tjandra, and Hsu}]{liu2023generative}
Liu, A.~H.; Le, M.; Vyas, A.; Shi, B.; Tjandra, A.; and Hsu, W.-N. 2024.
\newblock Generative Pre-training for Speech with Flow Matching.
\newblock In \emph{The Twelfth International Conference on Learning Representations (ICLR)}.

\bibitem[{Liu, Gong, and Liu(2023)}]{rectifiedflow}
Liu, X.; Gong, C.; and Liu, Q. 2023.
\newblock Flow Straight and Fast: Learning to Generate and Transfer Data with Rectified Flow.
\newblock In \emph{The Eleventh International Conference on Learning Representations (ICLR)}.

\bibitem[{Lu et~al.(2023)Lu, Chen, Chen, Su, Li, and Zhu}]{lu2023contrastive}
Lu, C.; Chen, H.; Chen, J.; Su, H.; Li, C.; and Zhu, J. 2023.
\newblock Contrastive energy prediction for exact energy-guided diffusion sampling in offline reinforcement learning.

\bibitem[{Mees et~al.(2024)Mees, Ghosh, Pertsch, Black, Walke, Dasari, Hejna, Kreiman, Xu, Luo et~al.}]{team2024octo}
Mees, O.; Ghosh, D.; Pertsch, K.; Black, K.; Walke, H.~R.; Dasari, S.; Hejna, J.; Kreiman, T.; Xu, C.; Luo, J.; et~al. 2024.
\newblock Octo: An open-source generalist robot policy.
\newblock In \emph{First Workshop on Vision-Language Models for Navigation and Manipulation at ICRA}.

\bibitem[{Nair et~al.(2020)Nair, Gupta, Dalal, and Levine}]{fawac}
Nair, A.; Gupta, A.; Dalal, M.; and Levine, S. 2020.
\newblock Awac: Accelerating online reinforcement learning with offline datasets.
\newblock \emph{arXiv preprint arXiv:2006.09359}.

\bibitem[{Nakamoto et~al.(2023)Nakamoto, Zhai, Singh, Sobol~Mark, Ma, Finn, Kumar, and Levine}]{nakamoto2023cal}
Nakamoto, M.; Zhai, S.; Singh, A.; Sobol~Mark, M.; Ma, Y.; Finn, C.; Kumar, A.; and Levine, S. 2023.
\newblock Cal-ql: Calibrated offline rl pre-training for efficient online fine-tuning.
\newblock \emph{Advances in Neural Information Processing Systems (NeurIPS)}.

\bibitem[{O’Neill et~al.(2024)O’Neill, Rehman, Maddukuri, Gupta, Padalkar, Lee, Pooley, Gupta, Mandlekar, Jain et~al.}]{o2024open}
O’Neill, A.; Rehman, A.; Maddukuri, A.; Gupta, A.; Padalkar, A.; Lee, A.; Pooley, A.; Gupta, A.; Mandlekar, A.; Jain, A.; et~al. 2024.
\newblock Open x-embodiment: Robotic learning datasets and rt-x models: Open x-embodiment collaboration 0.
\newblock In \emph{2024 IEEE International Conference on Robotics and Automation (ICRA)}.

\bibitem[{Park et~al.(2025)Park, Frans, Eysenbach, and Levine}]{ogbench}
Park, S.; Frans, K.; Eysenbach, B.; and Levine, S. 2025.
\newblock {OGB}ench: Benchmarking Offline Goal-Conditioned {RL}.
\newblock In \emph{The Thirteenth International Conference on Learning Representations (ICLR)}.

\bibitem[{Park et~al.(2024)Park, Frans, Levine, and Kumar}]{park2024value}
Park, S.; Frans, K.; Levine, S.; and Kumar, A. 2024.
\newblock Is value learning really the main bottleneck in offline RL?
\newblock \emph{Advances in Neural Information Processing Systems (NeurIPS)}.

\bibitem[{Park, Li, and Levine(2025)}]{fql}
Park, S.; Li, Q.; and Levine, S. 2025.
\newblock Flow Q-Learning.
\newblock In \emph{Forty-second International conference on machine learning (ICML)}.

\bibitem[{Peebles and Xie(2023)}]{dit}
Peebles, W.; and Xie, S. 2023.
\newblock Scalable diffusion models with transformers.
\newblock In \emph{Proceedings of the IEEE/CVF international conference on computer vision (ICCV)}, 4195--4205.

\bibitem[{Peters and Schaal(2007)}]{peters2007reinforcement}
Peters, J.; and Schaal, S. 2007.
\newblock Reinforcement learning by reward-weighted regression for operational space control.
\newblock In \emph{Proceedings of the 24th International conference on machine learning (ICML)}.

\bibitem[{Pomerleau(1988)}]{BC}
Pomerleau, D.~A. 1988.
\newblock Alvinn: An autonomous land vehicle in a neural network.
\newblock \emph{Advances in Neural Information Processing Systems (NeurIPS)}.

\bibitem[{Prasad et~al.(2024)Prasad, Lin, Wu, Zhou, and Bohg}]{consistency_policy}
Prasad, A.; Lin, K.; Wu, J.; Zhou, L.; and Bohg, J. 2024.
\newblock Consistency Policy: Accelerated Visuomotor Policies via Consistency Distillation.
\newblock \emph{CoRR}.

\bibitem[{Puterman(2014)}]{puterman2014markov}
Puterman, M.~L. 2014.
\newblock \emph{Markov decision processes: discrete stochastic dynamic programming}.
\newblock John Wiley \& Sons.

\bibitem[{Quinton and Rey(2024)}]{quinton2024jacobian}
Quinton, P.; and Rey, V. 2024.
\newblock Jacobian descent for multi-objective optimization.
\newblock \emph{arXiv preprint arXiv:2406.16232}.

\bibitem[{Schulman et~al.(2017)Schulman, Wolski, Dhariwal, Radford, and Klimov}]{schulman2017PPO}
Schulman, J.; Wolski, F.; Dhariwal, P.; Radford, A.; and Klimov, O. 2017.
\newblock Proximal policy optimization algorithms.
\newblock \emph{arXiv preprint arXiv:1707.06347}.

\bibitem[{Sheng et~al.(2025)Sheng, Wang, Li, and Liu}]{sheng2025mp1}
Sheng, J.; Wang, Z.; Li, P.; and Liu, M. 2025.
\newblock MP1: Mean Flow Tames Policy Learning in 1-step for Robotic Manipulation.
\newblock \emph{arXiv preprint arXiv:2507.10543}.

\bibitem[{Song et~al.(2023)Song, Dhariwal, Chen, and Sutskever}]{song2023consistency}
Song, Y.; Dhariwal, P.; Chen, M.; and Sutskever, I. 2023.
\newblock Consistency Models.
\newblock In \emph{International conference on machine learning (ICML)}.

\bibitem[{Song and Prafulla(2024)}]{iTCM}
Song, Y.; and Prafulla, D. 2024.
\newblock Improved Techniques for Training Consistency Models.
\newblock In \emph{12th International Conference on Learning Representations (ICLR)}.

\bibitem[{Sutton, Barto et~al.(1998)}]{sutton}
Sutton, R.~S.; Barto, A.~G.; et~al. 1998.
\newblock \emph{Reinforcement learning: An introduction}, volume~1.
\newblock MIT press Cambridge.

\bibitem[{Tabuada and Gharesifard(2021)}]{tabuada2020universal}
Tabuada, P.; and Gharesifard, B. 2021.
\newblock Universal approximation power of deep residual neural networks via nonlinear control theory.
\newblock In \emph{International Conference on Learning Representations (ICLR)}.

\bibitem[{Tarasov et~al.(2023{\natexlab{a}})Tarasov, Kurenkov, Nikulin, and Kolesnikov}]{ReBRAC}
Tarasov, D.; Kurenkov, V.; Nikulin, A.; and Kolesnikov, S. 2023{\natexlab{a}}.
\newblock Revisiting the minimalist approach to offline reinforcement learning.
\newblock \emph{Advances in Neural Information Processing Systems (NeurIPS)}.

\bibitem[{Tarasov et~al.(2023{\natexlab{b}})Tarasov, Nikulin, Akimov, Kurenkov, and Kolesnikov}]{tarasov2023corl}
Tarasov, D.; Nikulin, A.; Akimov, D.; Kurenkov, V.; and Kolesnikov, S. 2023{\natexlab{b}}.
\newblock CORL: Research-oriented deep offline reinforcement learning library.
\newblock \emph{Advances in Neural Information Processing Systems (NeurIPS)}.

\bibitem[{Vaswani et~al.(2017)Vaswani, Shazeer, Parmar, Uszkoreit, Jones, Gomez, Kaiser, and Polosukhin}]{vaswani2017attention}
Vaswani, A.; Shazeer, N.; Parmar, N.; Uszkoreit, J.; Jones, L.; Gomez, A.~N.; Kaiser, {\L}.; and Polosukhin, I. 2017.
\newblock Attention is all you need.
\newblock \emph{Advances in neural information processing systems (NeurIPS)}.

\bibitem[{Wang et~al.(2024)Wang, Wang, Jiang, Zou, Liu, Song, Wang, Xiao, Wu, Duan et~al.}]{wang2024diffusion}
Wang, Y.; Wang, L.; Jiang, Y.; Zou, W.; Liu, T.; Song, X.; Wang, W.; Xiao, L.; Wu, J.; Duan, J.; et~al. 2024.
\newblock Diffusion actor-critic with entropy regulator.
\newblock \emph{Advances in Neural Information Processing Systems (NeurIPS)}.

\bibitem[{Wang, Hunt, and Zhou(2022)}]{ifql}
Wang, Z.; Hunt, J.~J.; and Zhou, M. 2022.
\newblock Diffusion policies as an expressive policy class for offline reinforcement learning.
\newblock \emph{arXiv preprint arXiv:2208.06193}.

\bibitem[{Zhang et~al.(2024)Zhang, Xue, Niu, Chen, Yang, Li, and Liu}]{zhang2024generative_rl}
Zhang, J.; Xue, R.; Niu, Y.; Chen, Y.; Yang, J.; Li, H.; and Liu, Y. 2024.
\newblock Revisiting Generative Policies: A Simpler Reinforcement Learning Algorithmic Perspective.
\newblock \emph{arXiv preprint arXiv:2412.01245}.

\bibitem[{Zhang et~al.(2025)Zhang, Liu, Fan, Liu, Zeng, and Liu}]{zhang2025flowpolicy}
Zhang, Q.; Liu, Z.; Fan, H.; Liu, G.; Zeng, B.; and Liu, S. 2025.
\newblock Flowpolicy: Enabling fast and robust 3d flow-based policy via consistency flow matching for robot manipulation.
\newblock In \emph{Proceedings of the AAAI Conference on Artificial Intelligence (AAAI)}.

\bibitem[{Zhang, Zhang, and Gu(2025)}]{fbrac}
Zhang, S.; Zhang, W.; and Gu, Q. 2025.
\newblock Energy-weighted flow matching for offline reinforcement learning.
\newblock \emph{arXiv preprint arXiv:2503.04975}.

\end{thebibliography}

\clearpage

\clearpage
\onecolumn
\appendix
\section{Appendix}
\subsection{A\quad Limitations}
\addcontentsline{toc}{section}{Appendix A. Limitations}
Despite promising results, our method has several limitations. First, the final performance remains sensitive to the choice of the hyperparameter $\alpha$, which balances the behaviour cloning loss and the Q-value objective. Although we introduce an adaptive scheme for tuning $\alpha$, it does not fully eliminate the need for hyperparameter search, and the optimal value may still vary across tasks. Future work could explore multi-objective optimisation techniques (such as Jacobian Descent~\citep{quinton2024jacobian}) to better balance these competing objectives.
Second, our method relies on the MeanFlow formulation, which requires computing Jacobian-vector products (JVPs ~\citep{bradbury2018jax}) during training. While this enables efficient one-step flow-based modelling, it can introduce additional computational overhead, potentially affecting training speed in large-scale settings. We consider addressing these issues as promising directions for future work.

\subsection{B\quad Remedies for Action Clipping in One-Step Flow-Based Q-Learning Policies}
In its original form, MeanFlow generates an action by subtracting a learned velocity from a sample of Gaussian noise in one-step like:
\begin{equation}
    \hat{a} = e - u(e, b{=}0, t{=}1)
    \label{equ:meanflow_inference_appendix}
\end{equation}
where $e \sim \mathcal{N}(0, I)$ represents random noise, and $u$ is a neural network trained to predict an \emph{average velocity} vector. Although this formulation allows actions to be sampled without iterative integration over time, it suffers from a key drawback: since the noise $e$ is unbounded, the resulting action $\hat{a}$ often lies outside the valid action range (typically $[-1, 1]$). This issue is especially pronounced early in training, when the learned velocity $u$ is still inaccurate. As a result, actions are frequently clipped back into the valid range, which breaks the alignment between the policy's outputs and the Q-function's targets, leading to unstable and inefficient learning. 

A natural way to mitigate the issue of out-of-bound actions is to ensure that the final output of the generative model remains within the valid action range (e.g., $[-1, 1]$). Broadly, two strategies can be employed to achieve this goal. 

The first approach is to constrain the output through a suitable activation function (such as $\tanh$) applied at the end of the network. This guarantees bounded actions without the need for post-hoc clipping, thus preserving gradient consistency and improving stability during training.

The second, more direct approach is to replace the residual formulation entirely and train a model that maps the noise vector $e$ directly to an action $a$, i.e., learning a function $f_\theta: e \mapsto a$ that generates valid actions by design. This formulation avoids both explicit velocity modelling and action correction, offering a simpler forward process. 
We will discuss these two simple and direct solution in detail as follows: 

\subsubsection{B.1 \quad The Limitations of Gradient Distortion in Activation Mappings}

In the original MeanFlow formulation, actions are generated via a subtraction operation $a = e - u$, where $e$ is sampled from a standard Gaussian and $u$ is a learned average velocity. However, since $e$ is unbounded, the resulting action $a$ may fall outside the valid action range (e.g., $[-1, 1]$). A common workaround is to apply hard clipping post-hoc, but this introduces discontinuities in the gradient flow and leads to gradient distortion during training.

To resolve this issue, we adopt a smooth bounding strategy based on the \emph{softsign} activation function. A natural idea is to directly apply softsign to the output $e - u$:
\begin{equation}
    a = \mathrm{softsign}(e - u) = \frac{e - u}{1 + |e - u|}
\end{equation}
which smoothly squashes the unbounded input into the range $(-1, 1)$, thereby avoiding invalid actions and eliminating the need for hard clipping. However, this also means the model no longer directly learns the true action $a$, but rather a transformed version of it, which may impede expressivity or stability.

To address this, we instead apply the inverse of the softsign function during training. Specifically, given a target action $a \in [-1, 1]$, we compute its approximate inverse:
\begin{equation}
    \mathrm{action\_inv} = \mathrm{inv\_softsign}(a) = \frac{a}{1 - |a| + \varepsilon}
\end{equation}
where $\varepsilon$ is a small constant for numerical stability. The model is then trained to predict $\mathrm{action\_inv}$ in the unconstrained space, and the final action is recovered via a forward softsign transformation:
\begin{equation}
    a = \mathrm{softsign}(\mathrm{action\_inv})
\end{equation}

This design ensures that the model output always lies within valid bounds, while retaining a smooth, differentiable mapping and avoiding post-hoc clipping. Moreover, it preserves the expressivity of the learned actions by decoupling the bounded constraint from the learning target.

\begin{figure}[t]
    \centering
    \includegraphics[width=0.6\linewidth]{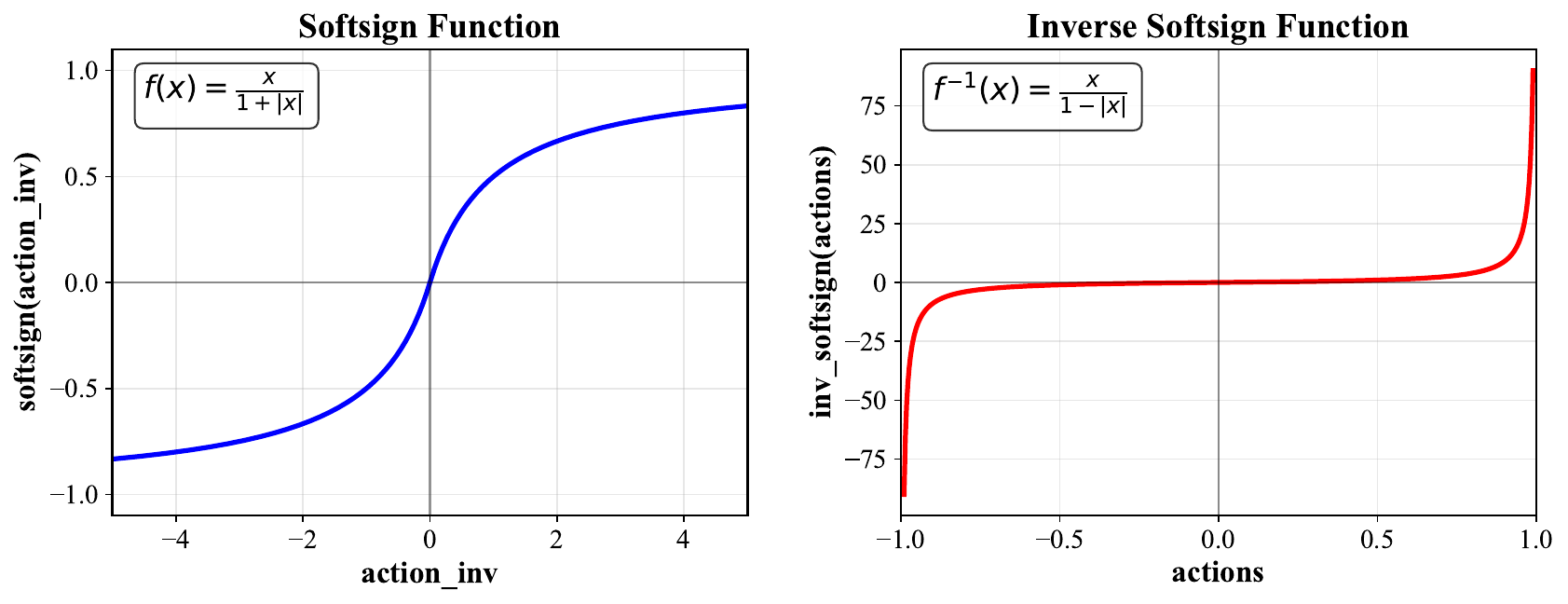}
    \caption{Visualization of the forward and inverse softsign transformations. Left: the \texttt{softsign} function smoothly bounds unbounded inputs to $(-1, 1)$. Right: the approximate inverse \texttt{inv\_softsign} restores an unbounded representation, enabling consistent bidirectional mappings for policy learning.}
    \label{fig:softsign}
\end{figure}

Figure~\ref{fig:softsign} visualizes both the softsign function and its inverse. The softsign curve demonstrates smooth saturation as input magnitude increases, while the inverse function expands the bounded action back into an unbounded domain. Together, they form a stable and differentiable mapping pair for generative policies.

Although the forward-inverse pair formed by the softsign and its approximate inverse exhibits appealing mathematical properties—particularly the ability to perform smooth, lossless value transformations in the numerical sense—the empirical performance of this approach in policy learning remains suboptimal. The key limitation arises not from the value-preserving nature of the transformation, but from its \emph{gradient distortion}.

While the transformation $a = \mathrm{softsign}(x)$ is numerically invertible, its gradient $\frac{d\,\mathrm{softsign}(x)}{dx} = \frac{1}{(1 + |x|)^2}$ is highly non-uniform across the input space. Near the origin (i.e., $x \approx 0$), the gradient remains relatively large and stable. However, as $x$ increases in magnitude, the gradient rapidly diminishes. This imbalance means that inputs near the saturation region (e.g., when $|x| \to 5$) contribute far less to the gradient signal during backpropagation.

More problematically, when performing the inverse mapping $x = \mathrm{inv\_softsign}(a)$, the gradient $\frac{d\,\mathrm{inv\_softsign}(a)}{da} = \frac{1}{(1 - |a| + \varepsilon)^2}$ becomes sharply amplified as $|a| \to 1$. As a result, actions near the boundary (i.e., $a \approx \pm 1$) dominate the gradient update, leading the model to prefer generating actions near the extremes. This skewed gradient flow introduces significant training instability and mode collapse around the action boundaries, ultimately undermining the policy's expressivity and performance.

Therefore, despite its smoothness and reversibility, the softsign-based activation strategy suffers from a mismatch between forward stability and backward learning dynamics—a mismatch that proves detrimental in practice.

\subsubsection{B.2 \quad The Pitfall of Simple Residual Substitution}

Another natural extension of MeanFlow is to explicitly define a residual mapping from noise to action by setting:
\begin{equation}
    g_\theta(a_t, b, t) = e - u_\theta(e, b, t),
\end{equation}
where \( u(e, b, t) \) denotes the average velocity field. This reformulation enables single-stage action generation while preserving the underlying structure of MeanFlow.

To derive the training objective for \( g \), we begin from the \emph{MeanFlow Identity} \citep{meanflow}, which defines the average velocity as:
\begin{equation}
(t - b) \cdot u(a_t, b, t) \triangleq \int_b^t v(a_\tau, \tau)\, d\tau,
\label{eq:av_velocity_appendix}
\end{equation}
where \( a_t = (1 - t) \cdot a + t \cdot e \), and \( v \) denotes the instantaneous velocity field.

Based on this identity, we derive the regression target to supervise the learning of \( g_\theta \). Differentiating both sides of Equation~\ref{eq:av_velocity_appendix} with respect to \( t \), we obtain:
\begin{equation}
u(a_t, b, t) + (t - b) \cdot \frac{d}{dt} u(a_t, b, t) = v(a_t, t).
\label{eq:velocity}
\end{equation}

Substituting \( u(e, b, t) = e - g(a_t, b, t) \) into Equation~\ref{eq:velocity} yields:
\begin{equation}
\begin{aligned}
& (e - g(a_t, b, t)) + (t - b) \cdot \frac{d}{dt}(e - g(a_t, b, t)) = v(a_t, t) \\
\Rightarrow \quad & e - g(a_t, b, t) - (t - b) \cdot \frac{d}{dt} g(a_t, b, t) = v(a_t, t).
\end{aligned}
\end{equation}

Rearranging terms, we arrive at a key identity for the residual function:
\begin{equation}
g(a_t, b, t) = e - v(a_t, t) + (t - b) \cdot \frac{d}{dt} g(a_t, b, t).
\label{eq:g_rewrite}
\end{equation}

To expand the total derivative \( \frac{d}{dt} g(a_t, b, t) \), we apply the chain rule:
\begin{equation}
\frac{d}{dt} g(a_t, b, t) = \frac{d a_t}{dt} \cdot \partial_{a_t} g + \frac{d b}{dt} \cdot \partial_b g + \frac{d t}{dt} \cdot \partial_t g.
\end{equation}
Given \( a_t = (1 - t) a + t e \), we have \( \frac{d a_t}{dt} = e - a = v(a_t, t) \), \( \frac{d b}{dt} = 0 \), and \( \frac{d t}{dt} = 1 \). Thus:
\begin{equation}
\frac{d}{dt} g(a_t, b, t) = v(a_t, t) \cdot \partial_{a_t} g + \partial_t g.
\label{eq:dgdt}
\end{equation}

Substituting Equation~\ref{eq:dgdt} into Equation~\ref{eq:g_rewrite}, we obtain the final regression target:
\begin{equation}
\begin{aligned}
g_{tgt} &= e - v(a_t, t) + (t - b) \left[ v(a_t, t) \cdot \partial_{a_t} g + \partial_t g \right] \\
             &= a + (t - b) \left[ v(a_t, t) \cdot \partial_{a_t} g + \partial_t g \right].
\end{aligned}
\label{eq:g_final}
\end{equation}

We now parameterize \( g_\theta \) as a neural network and treat Equation~\ref{eq:g_final} as a supervision signal for training. Specifically, we define a  behaviour cloning loss to regress the predicted residuals toward their analytical targets:
\begin{equation}
\mathcal{L}_{BC}(\theta) = \mathbb{E}\left\| g_\theta(a_t, b, t) - \operatorname{sg}(g_{\text{tgt}}) \right\|_2^2,
\end{equation}
where the target residual is given by:
\begin{equation}
g_{\text{tgt}} = a + (t - b) \left[ v(a_t, t) \cdot \partial_{a_t} g + \partial_t g \right].
\end{equation}

\begin{figure}[H]
    \centering
    \includegraphics[width=.8\linewidth, trim=0 350 180 0, clip]{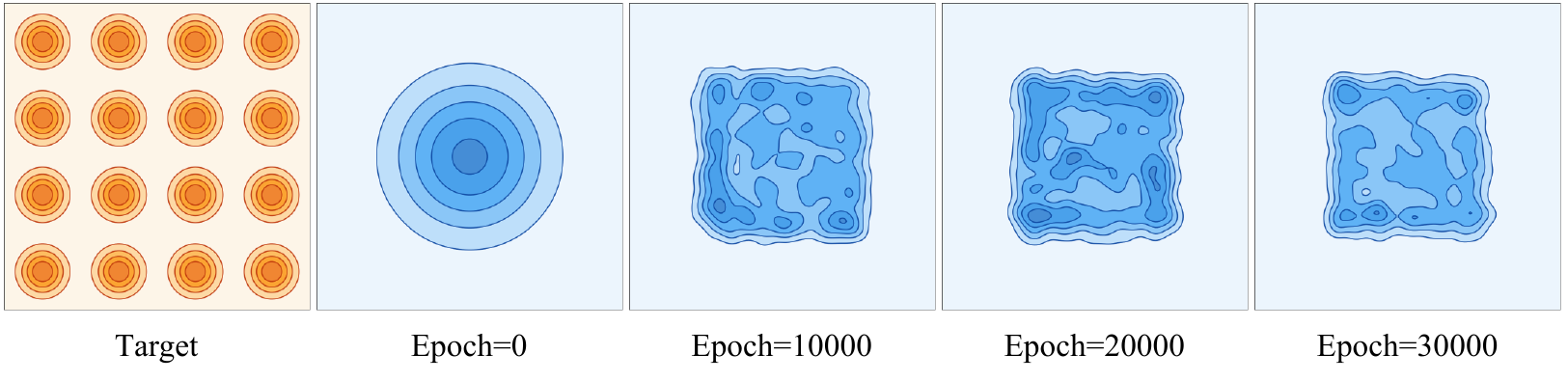}
\caption{Training dynamics of the residual model \( g_\theta(e) = e - u(e, b, t) \) on a 4\(\times\)4 checkerboard distribution. From left to right: the target distribution and the model outputs at different training epochs (0, 10k, 20k, 30k). Although the residual mapping allows for single-stage sampling, the learned distribution fails to recover the desired multi-modal structure, revealing the limitations of directly modelling \( e - u \).}   
    \label{fig:failureofe-u}
\end{figure}

This formulation seemingly preserves the theoretical foundation of MeanFlow while enabling efficient one-stage sampling. However, as we show in our toy dataset experiments in Figure~\ref{fig:failureofe-u}, this residual substitution yields results that deviate significantly from our initial intuition, revealing nontrivial empirical mismatches despite its apparent theoretical appeal.

\subsubsection{B.3 \quad Equivalent Reformulation of MeanFlow}
\label{app:equalref}

Given that both intuitive modifications---bounded activation functions and simple substitutions---fail to match the performance and generality of the original MeanFlow formulation, we are motivated to address the following question:

\begin{center}
    \emph{How can we construct a network that is both theoretically equivalent to MeanFlow}\\
    \emph{and practically enables one-step noise-to-action generation?}
\end{center}

To this end, we revisit the core structure of MeanFlow: its defining characteristic is the integration of a velocity field over time, which effectively captures a continuous flow from the noise vector \( e \) to the action \( a \). Any alternative formulation that aims to preserve this behavior must replicate not only the mapping \( e \mapsto a \), but also the intermediate dynamics and temporal consistency embedded in the original construction. Formally, we propose:

\paragraph{Remark.}
To enable direct one-step inference---where actions are deterministically generated from Gaussian noise---it is beneficial to enforce the following constraint:
\begin{equation}
    \phi(e, b^\ast, t^\ast) = e, \quad \text{for some fixed } b^\ast, t^\ast,
\end{equation}
where \( e \sim \mathcal{N}(0, I) \).

Under this condition, the residual model satisfies:
\begin{equation}
    g_\theta(e, b^\ast, t^\ast) = \phi(e, b^\ast, t^\ast) - u_\theta(e, b^\ast, t^\ast) = a,
\end{equation}
which enables tractable and controlled noise-to-action generation at a designated inference time step.

\paragraph{Design Rationale.}
To construct a residual formulation of the form
\begin{equation}
    g_\theta(a_t, b, t) = \phi(a_t, b, t) - u_\theta(a_t, b, t),
\end{equation}
our goal is to preserve the expressive power and theoretical soundness of the original MeanFlow framework. This requires that the reformulation be both well-posed and learnable in practice. To this end, we identify two key properties that the residual \(\phi\) must satisfy:

\begin{itemize}
    \item \textbf{\(\phi\) should be fixed.}  
    To ensure a unique and interpretable decomposition, \(\phi\) should be fixed and independent of the learnable component \(u_\theta\). If both components are trained jointly, the residual \(g_\theta\) remains invariant under simultaneous perturbations of \(\phi\) and \(u_\theta\), resulting in a degenerate solution space and entangled gradients. Fixing \(\phi\) eliminates this ambiguity and leads to stable, well-defined optimisation.

    \item \textbf{\(\phi\) should be analytic.}  
    In temporally evolving settings—such as flow-based training along interpolated trajectories—differentiability with respect to time is essential. If \(\phi\) lacks regularity, the required derivatives may become ill-defined or unstable. Requiring \(\phi\) to be analytic ensures smooth gradient propagation and coherent temporal behavior.
\end{itemize}

Together, these two desiderata motivate a sufficient condition: \(\phi\) should be a fixed, analytic function with respect to its input variables \( (a_t, b, t) \), defined over a compact domain. The following proposition formalizes this condition and demonstrates that it preserves the expressive capacity of the residual formulation.

\paragraph{Proposition (Sufficiency).}
Let the residual parameterization be defined as
\begin{equation}
    g_\theta(a_t, b, t) = \phi(a_t, b, t) - u_\theta(a_t, b, t),
\end{equation}
where \( \phi \colon \mathcal{X} \to \mathbb{R}^d \) is a fixed, analytic function independent of the learnable component \( u_\theta \), and \( \mathcal{X} \subset \mathbb{R}^m \) is a compact domain. Then, for any target function \( h \in \mathcal{C}(\mathcal{X}; \mathbb{R}^d) \), there exists a neural network \( u_\theta \colon \mathcal{X} \to \mathbb{R}^d \) such that the residual form \( g_\theta = \phi - u_\theta \) approximates \( h \) arbitrarily well in the uniform norm.

\begin{proof}
Let \( h \in \mathcal{C}(\mathcal{X}; \mathbb{R}^d) \) be an arbitrary continuous target function. Define the corresponding residual target:
\begin{equation}
    u^\ast(x) := \phi(x) - h(x), \quad \text{where } x := (a_t, b, t) \in \mathcal{X}.
\end{equation}
Since both \( \phi \) and \( h \) are continuous on the compact domain \( \mathcal{X} \), their difference \( u^\ast \) is also continuous, i.e., \( u^\ast \in \mathcal{C}(\mathcal{X}; \mathbb{R}^d) \).

By the universal approximation theorem~\citep{leshno1993multilayer}, for any \( \varepsilon > 0 \), there exists a neural network \( u_\theta \colon \mathcal{X} \to \mathbb{R}^d \) such that
\begin{equation}
    \sup_{x \in \mathcal{X}} \| u_\theta(x) - u^\ast(x) \| < \varepsilon.
\end{equation}
Substituting into the residual formulation \( g_\theta(x) = \phi(x) - u_\theta(x) \), we obtain:
\begin{align}
    \| g_\theta(x) - h(x) \| 
    &= \| \phi(x) - u_\theta(x) - h(x) \| \nonumber \\
    &= \| u^\ast(x) - u_\theta(x) \| < \varepsilon, \quad \forall x \in \mathcal{X}.
\end{align}
Hence, the residual function \( g_\theta = \phi - u_\theta \) uniformly approximates the target \( h = \phi - u^\ast \) on \( \mathcal{X} \). 
\end{proof}

\paragraph{Conclusion.}
We conclude that:
\begin{align}
&\phi(a_t, b, t) \text{ is a fixed, analytic function of the input variables, independent of } \theta
\nonumber \\
&\Rightarrow\quad 
g_\theta(a_t, b, t) = \phi(a_t, b, t) - u_\theta(a_t, b, t)
\text{ retains the expressivity of the original } u_\theta.
\end{align}

\paragraph{Boundary Condition and Construction of $\phi$.}
To enable a tractable and deterministic noise-to-action mapping at inference time, we seek a formulation where the residual output at \( b = 0 \) and \( t = 1 \) satisfies
\begin{equation}
    g_\theta(e, 0, 1) = \phi(e, 0, 1) - u_\theta(e, 0, 1) \approx a,
\end{equation}
with \( e \sim \mathcal{N}(0, I) \). This requirement can be fulfilled by choosing \( \phi \) as a fixed, continuous (or analytic) function independent of the parameters \( \theta \), such that \( \phi(e, 0, 1) = e \).

We adopt the simplest and most natural choice:
\begin{equation}
    \phi(a_t, b, t) = a_t = (1 - t) \cdot a + t \cdot e,
\end{equation}
which satisfies \( \phi(e, 0, 1) = e \) under the standard linear interpolation schedule. Under this construction, the reformulation takes the form
\begin{equation}
    g_\theta(a_t, b, t) = a_t - u_\theta(a_t, b, t),
\end{equation}
and in particular:
\begin{equation}
    g_\theta(e, 0, 1) = e - u_\theta(e, 0, 1) = a.
\end{equation}

This design, where \( \phi \) is fixed, analytic, and independent of \( u_\theta \), ensures that the residual formulation \( g_\theta = \phi - u_\theta \) retains the full expressivity of the original \( u_\theta \). Moreover, it enables efficient one-step sampling that remains consistent with the MeanFlow decoding identity under the boundary condition \( (b = 0, t = 1) \).

\paragraph{Analysis of Reformulation Variants.}
We revisit the design of reformulation schemes through the lens of our theoretical framework, particularly focusing on the validity of the residual form \( g_\theta = \phi - u_\theta \). 
As summarized in Table~\ref{tab:reparam_variants}, we explore several alternative formulations of the residual function in the form $g(a_t, b, t) = \phi(a_t, b, t) - u_\theta(a_t, b, t)$, each leading to a distinct training target. Below, we derive the corresponding target expressions and inference procedures for four representative variants.%
\footnote{For brevity, we omit variants $g(a_t, b, t) = e - u_\theta(a_t, b, t)$, as it has already been discussed in the preceding section.}

\paragraph{Variant 1: \( \phi = a_t \Rightarrow g = a_t - u_\theta(a_t, b, t) \)}

Substituting the \emph{MeanFlow Identity} into the definition of \( g \), we have:
\begin{equation}
\begin{aligned}
        g(a_t, b, t) 
        &= a_t - u(a_t, b, t) \\
        &= a_t - \left[ v(a_t, t) - (t - b) \cdot \frac{d}{dt}(a_t - g(a_t, b, t)) \right] \\
        &= a_t - v(a_t, t) + (t - b)v(a_t, t) - (t - b) \cdot \frac{d}{dt} g(a_t, b, t).
\end{aligned}
\end{equation}

We compute the total derivative of \( g(a_t, b, t) \) with respect to \( t \):
\begin{equation}
\frac{d}{dt} g(a_t, b, t) = v(a_t, t) \cdot \partial_{a_t} g + \partial_b g + \partial_t g = \texttt{jvp}(g, (a_t, b, t), (v, 0, 1)),
\label{equ:appendix_total_dev}
\end{equation}
where \texttt{jvp} denotes the Jacobian-vector product along the temporal direction.

Substituting Equation~\ref{equ:appendix_total_dev} into the residual formulation and simplifying, we obtain the target residual:
\begin{equation}
    g_{\text{tgt}} =  a_t + (t-b-1)v(a_t, t) - (t - b) \cdot \texttt{jvp}.
\end{equation}

\textbf{Inference:} When \( b = 0, t = 1 \), the action can be directly recovered as:
\begin{equation}
    a = e - u(e, 0, 1) = e - [a_t - g(e, 0, 1)] = g(e, 0, 1).
\end{equation}

\paragraph{Variant 2: \( \phi = 2 \Rightarrow g = 2 - u_\theta(a_t, b, t) \)}

Substituting \emph{MeanFlow Identity} equation into the residual function:
\begin{equation}
\begin{aligned}
        g(a_t, b, t) 
        & = 2 - u(a_t, b, t) \\
        & = 2 - \left[ v(a_t, t) - (t - b) \cdot \frac{d}{dt}(2 - g(a_t, b, t)) \right] \\
        & = 2 - v(a_t, t) + (t - b) \cdot \left( - \frac{d}{dt} g(a_t, b, t) \right) \\
        & = 2 - v(a_t, t) - (t - b) \cdot \frac{d}{dt} g(a_t, b, t)
\end{aligned}
\end{equation}

Substituting the total derivative (Equation \ref{equ:appendix_total_dev}) into the residual equation and regrouping terms, we can obtain:
\begin{equation}
    g_{\text{tgt}} = 2 - v(a_t, t) - (t - b) \cdot \texttt{jvp}
\end{equation}

\textbf{Inference:} When \( b = 0, t = 1 \), the final action is recovered as:
\begin{equation}
a = e - u(e, 0, 1) = e - [2 - g(e, 0, 1)]
\end{equation}

\paragraph{Variant 3: \( \phi = t \Rightarrow g = t - u_\theta(t, b, t) \)}

Substituting \emph{MeanFlow Identity} equation into the residual function:
\begin{equation}
\begin{aligned}
    g(a_t, b, t) 
    &= t - u(a_t, b, t) \\
    &= t - \left[ v(a_t, t) - (t - b) \cdot \frac{d}{dt}(t - g(a_t, b, t)) \right] \\
    &= t - v(a_t, t) + (t - b) \cdot \left( 1 - \frac{d}{dt} g(a_t, b, t) \right) \\
    &= 2t - b - v(a_t, t) - (t - b) \cdot \frac{d}{dt} g(a_t, b, t)
\end{aligned}
\end{equation}

Using the total derivative from Equation~\ref{equ:appendix_total_dev}, substitute into the residual reformulation and group terms:
\begin{equation}
    g_{\text{tgt}} = 2t - b - v(a_t, t) - (t - b)\cdot \texttt{jvp}
\end{equation}

\textbf{Inference:} When \( b = 0, t = 1 \), the final action is recovered as:
\begin{equation}
a = e - u(e, 0, 1) = e - (1- g(e, 0, 1))
\end{equation}

\paragraph{Variant 4: \( \phi = 2a_t \Rightarrow g = 2a_t - u_\theta(a_t, b, t) \)}

Substituting \emph{MeanFlow Identity} equation into the residual function:
\begin{equation}
\begin{aligned}
    g(a_t, b, t) 
    &= 2a_t - u(2a_t, b, t) \\
    &= 2a_t - \left[ v(a_t, t) - (t - b) \cdot \frac{d}{dt}(2a_t - g(a_t, b, t)) \right] \\
    &= 2a_t - v(a_t, t) + (t - b) \cdot \left( 2 \cdot \frac{d a_t}{dt} - \frac{d}{dt} g(a_t, b, t) \right)
\end{aligned}
\end{equation}

Recall that \( \frac{d a_t}{dt} = v(a_t, t) \). Then we have:
\begin{equation}
\begin{aligned}
    g(a_t, b, t)
    &= 2a_t - v(a_t, t) + 2(t - b)v(a_t, t) - (t - b)\cdot \frac{d}{dt} g(a_t, b, t) \\
    &= 2a_t + (2t - 2b - 1)v(a_t, t) - (t - b) \cdot \frac{d}{dt} g(a_t, b, t)
\end{aligned}
\end{equation}

By applying the total derivative from Equation~\ref{equ:appendix_total_dev} and substituting it into the residual formulation, we simplify the expression to obtain:
\begin{equation}
g_{\text{tgt}} = 2a_t + (2t - 2b - 1) v(a_t, t) - (t - b)\cdot \texttt{jvp}
\end{equation}

\textbf{Inference:} When \( b = 0, t = 1 \), we have:
\begin{equation}
a = e - u(e, 0, 1) = e - [2a_t - g(a_t, 0, 1)] = g(e, 0, 1) - e
\end{equation}

Consequently, our theoretical framework offers a principled explanation for the failure of certain residual formulations, such as \( g_\theta = e - u_\theta \) and \( g_\theta = et - u_\theta \), to serve as effective reformulations of \( u_\theta \). Specifically, these variants choose \( \phi(e, b, t) = e \) or \( et \), which are neither continuous nor analytic with respect to the input variables \( (a_t, b, t) \), and fail to depend functionally on the conditioning trajectory. As a result, the decomposition becomes ill-posed: the network \( u_\theta \) is forced to approximate an unstructured and potentially discontinuous residual, thereby violating the assumptions required for universal approximation and stable optimization.

This theoretical insight is directly supported by empirical results on toy datasets, as detailed in Table~\ref{tab:reparam_variants}. Variants that violate our proposed sufficient conditions—such as \( g_\theta = e - u_\theta \) and \( g_\theta = et - u_\theta \)—consistently exhibit degraded performance. Specifically, these settings lead to mode collapse, poor convergence, and substantial distributional shift, as reflected by high 2-Wasserstein distances (\( W_2 = 0.591 \) and \( 0.393 \), respectively) and visibly distorted output distributions.

\begin{figure}[htbp]
    \centering
    \includegraphics[width=0.52\linewidth, trim=0cm 6.5cm 10.8cm 0cm, clip]{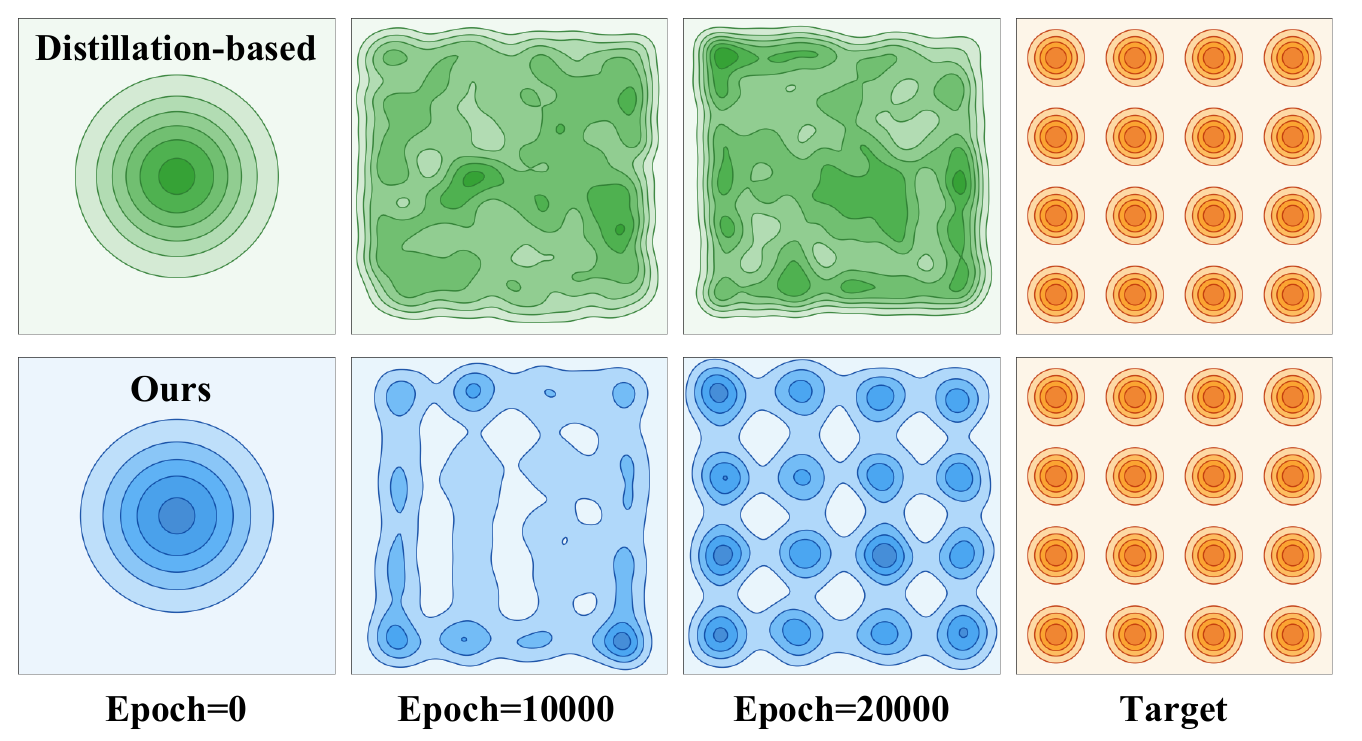}
    \caption{Toy example comparing the multimodal representation learning capabilities of the distillation-based one-step flow matching method and our proposed one-step policy model. The proposed model demonstrates improved accuracy and efficiency in capturing distinct modes of the target distribution, highlighting its superior ability to represent complex multimodal structures.}
    \label{fig:toy example}
    \vspace{-0.2cm}
\end{figure}

In contrast, all residual formulations that satisfy our theoretical conditions—such as \( a_t - u_\theta \), \( 2a_t - u_\theta \), \( t - u_\theta \), and even the constant form \( 2 - u_\theta \)—demonstrate stable training dynamics and successfully recover the target distribution. These formulations achieve low Wasserstein distances (all near \( 0.263 \)) and generate samples that are visually indistinguishable from those produced by the original MeanFlow. The alignment between the “Theory-Compatible” column in Table~\ref{tab:reparam_variants}, the numerical metric, and the distribution visualizations provides strong empirical validation of our theoretical framework. Notably, this holds even for seemingly simple choices of \(\phi\), such as constant or linear functions of time and input, as long as they remain fixed, analytic, and input-dependent.
Subsequently, we examined whether each formulation permits direct noise-to-action mapping via the network, and ultimately selected the residual form \( a_t - u_\theta \) as shown in Table \ref{tab:reparam_variants}. This choice is motivated by the fact that it is the only variant that simultaneously satisfies our theoretical conditions and enables direct generation of actions from noise through the network.
Additionally, we evaluate the practical effectiveness of our reformulation using the same multimodal toy example shown in Fig.~\ref{fig:toy example}. Under the same number of training epochs, our approach consistently outperforms distillation-based generative baselines, demonstrating a superior capacity to capture complex multimodal action distributions.

\paragraph{Summary.}
The residual formulation
\begin{equation}
g_\theta(a_t, b, t) = a_t - u_\theta(x_t, b, t)
\end{equation}
provides a theoretically grounded parameterization for controllable noise-to-action mappings, while retaining the full expressivity of the original MeanFlow. \textit{It is worth noting that this residual form constitutes only one valid instantiation within our general framework; any formulation employing a fixed, continuous, analytic function \( \phi \), independent of \( u_\theta \), equally satisfies the sufficient conditions for expressivity and stability.}

\begin{table*}[t]
\centering
\caption{\textbf{Comparison of Reformulation Variants.} Each row shows a formulation of \( g \), its regression target, inference procedure, theoretical properties, and toy experiment performance. 
The reported metric is the 2-Wasserstein distance, which quantifies distribution-level alignment between generated actions and ground-truth samples in a 2D toy task. \emph{Theory-Compatible?} indicates whether the reformulation variants $g$ conforms to our  theoretical framework. \emph{Noise-to-Action?} indicates whether the method supports direct generation of actions from noise via a  single forward pass. }

\label{tab:reparam_variants}
\renewcommand{\arraystretch}{1.3}
\setlength{\tabcolsep}{6pt}
\begin{tabular}{>{\centering\arraybackslash}m{1.2cm} 
                >{\centering\arraybackslash}m{3.2cm} 
                >{\centering\arraybackslash}m{2.6cm} 
                >{\centering\arraybackslash}m{1.5cm}
                >{\centering\arraybackslash}m{1.5cm}
                >{\centering\arraybackslash}m{1.0cm}
                >{\centering\arraybackslash}m{4.3cm}}
\toprule
\textbf{\( g \)} & {Target} & {Inference} & {Theory-Compatible?} & {Noise-to-Action?} & {Metrics $(\downarrow)$} & {Performance on Toy Experiment} \\
\midrule
\( u \) & \( v - (t - b)\cdot \texttt{jvp} \) & \( g(e, 0, 1) \) & \cmark & \xmark  & $0.263$ & \includegraphics[width=0.95\linewidth]{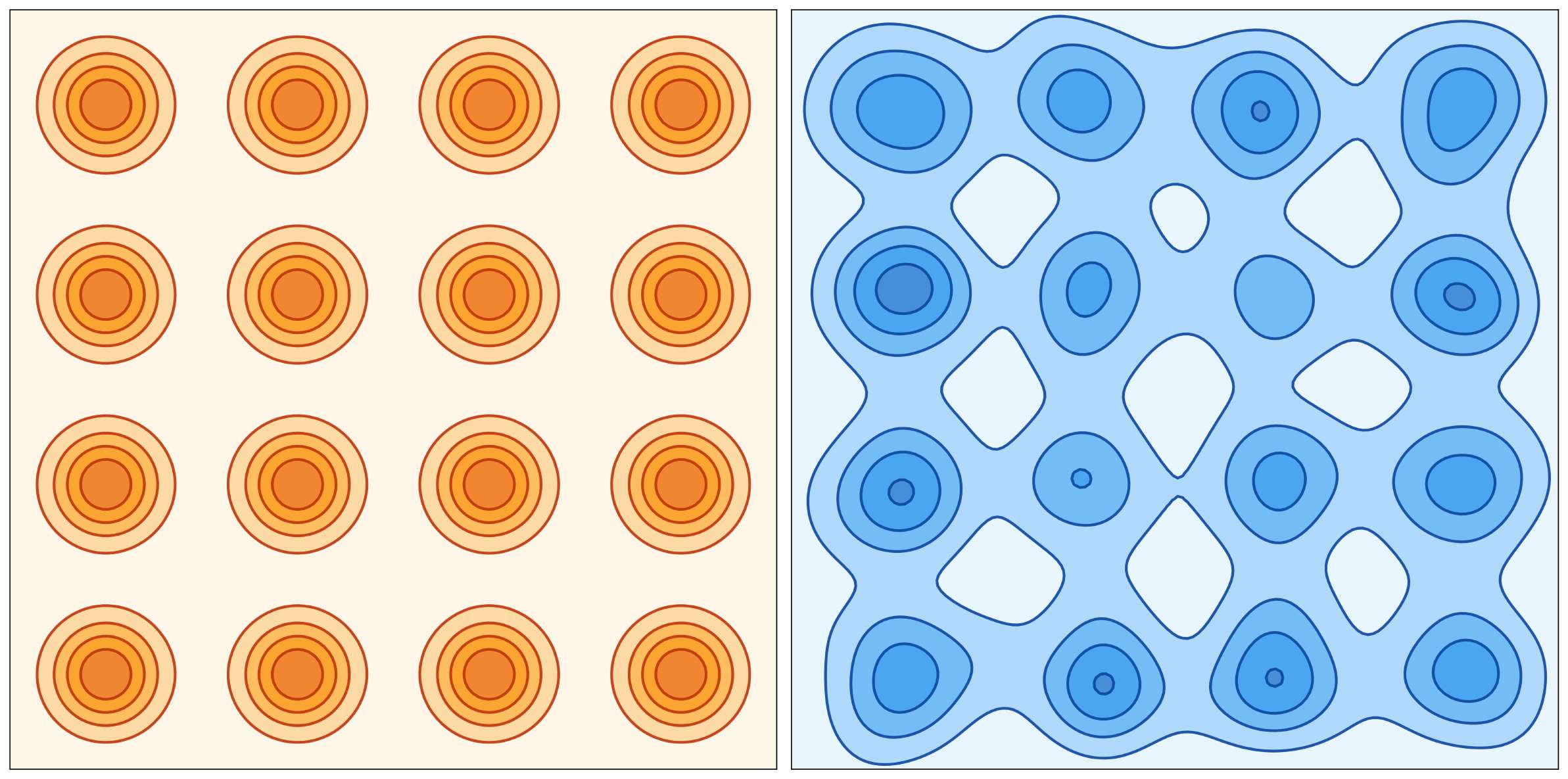} \\
\( a_t - u \) & \( a + (t-b)v - (t-b) \cdot \texttt{jvp} \) & \( g(e, 0, 1) \) & \cmark & \cmark &  $0.262$ & \includegraphics[width=0.95\linewidth]{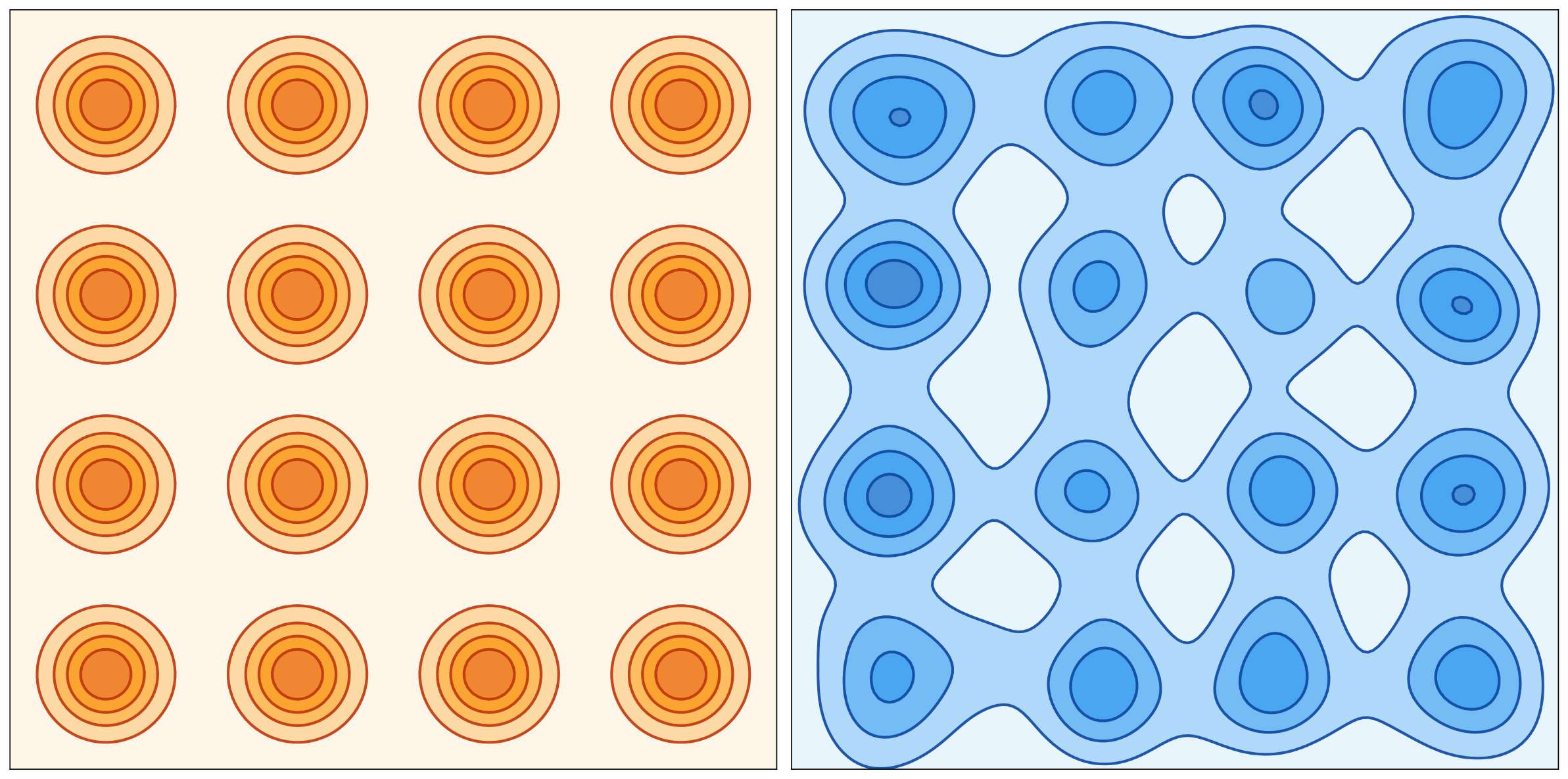} \\
\midrule
\( e - u \) & \( e - v - (t-b) \cdot \texttt{jvp} \) & \( g(e, 0, 1) \) & \xmark & \cmark & $0.591$ & \includegraphics[width=0.95\linewidth]{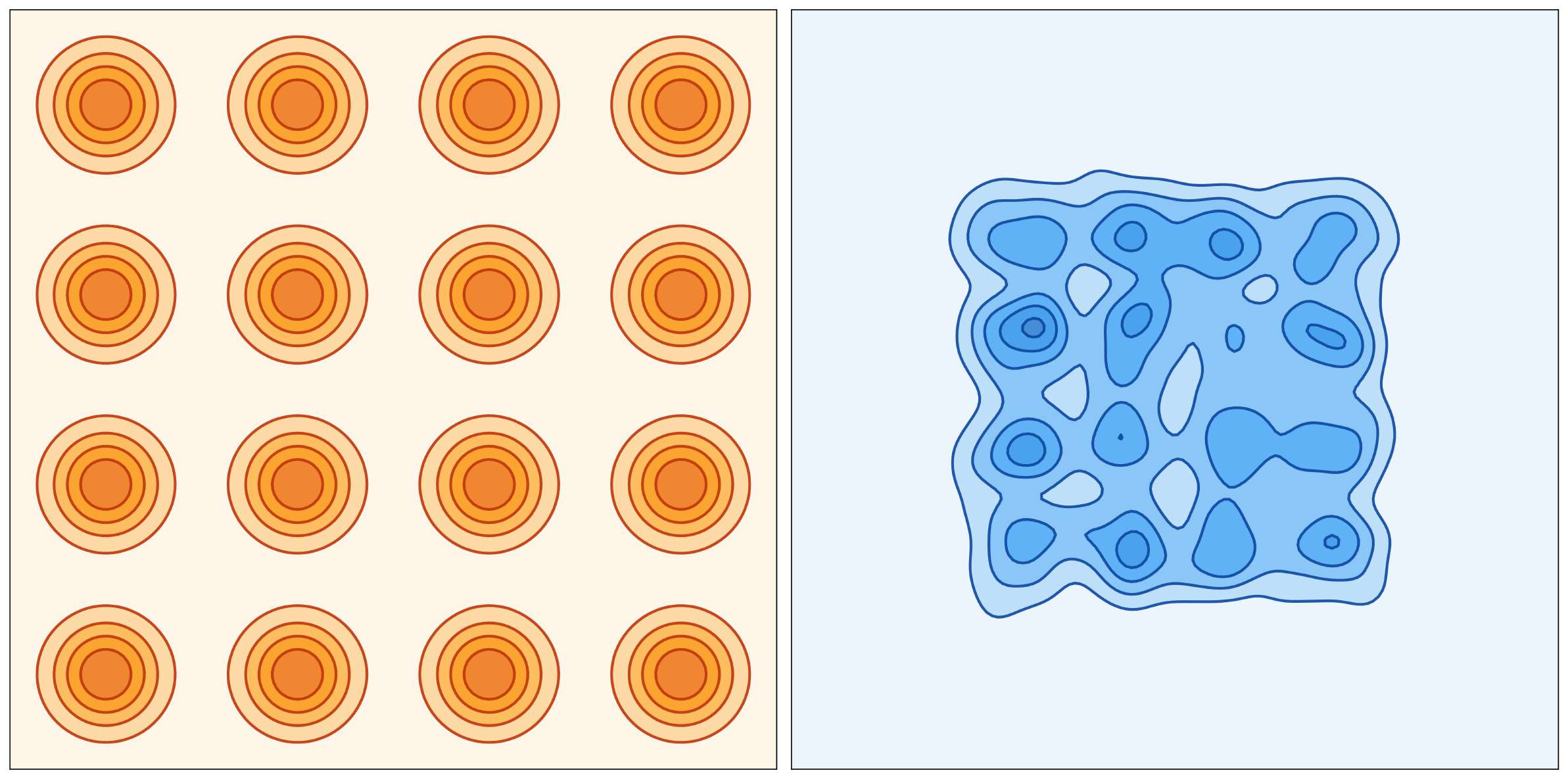} \\
\( et - u \) & \( (2t-b)e - v - (t-b) \cdot \texttt{jvp} \) & \( g(e, 0, 1) \) & \xmark & \cmark & $0.393$ & \includegraphics[width=0.95\linewidth]{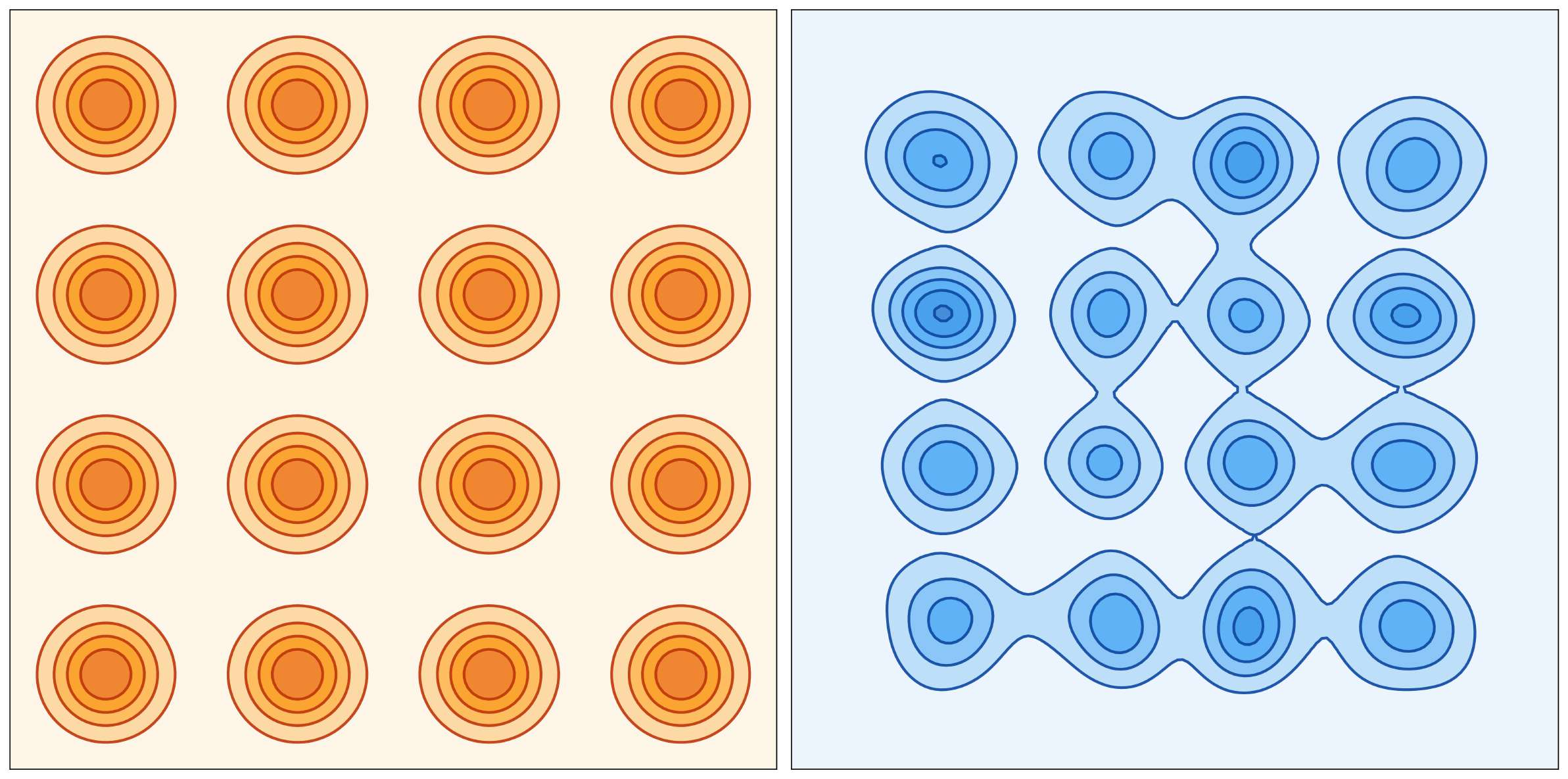} \\
\( 2 - u \) & \( 2 - v - (t-b) \cdot \texttt{jvp} \) & \( e - (2 - g(e, 0, 1)) \) & \cmark & \xmark & $0.267$  & \includegraphics[width=0.95\linewidth]{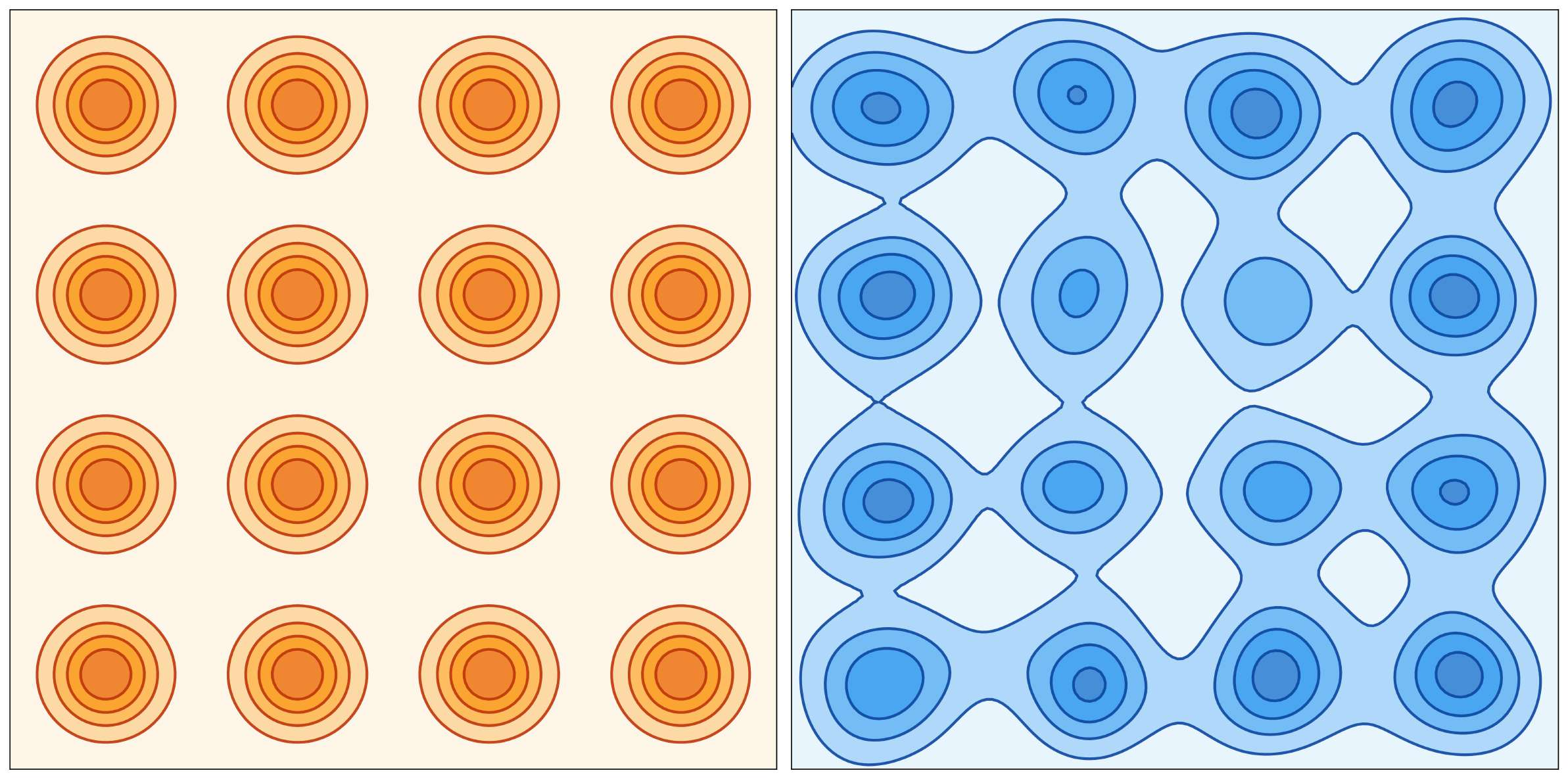} \\
\( t - u \) & \( 2t-b - v - (t-b) \cdot \texttt{jvp} \) & \( e - (1 - g(e, 0, 1)) \) & \cmark & \xmark & $0.264$  & \includegraphics[width=0.95\linewidth]{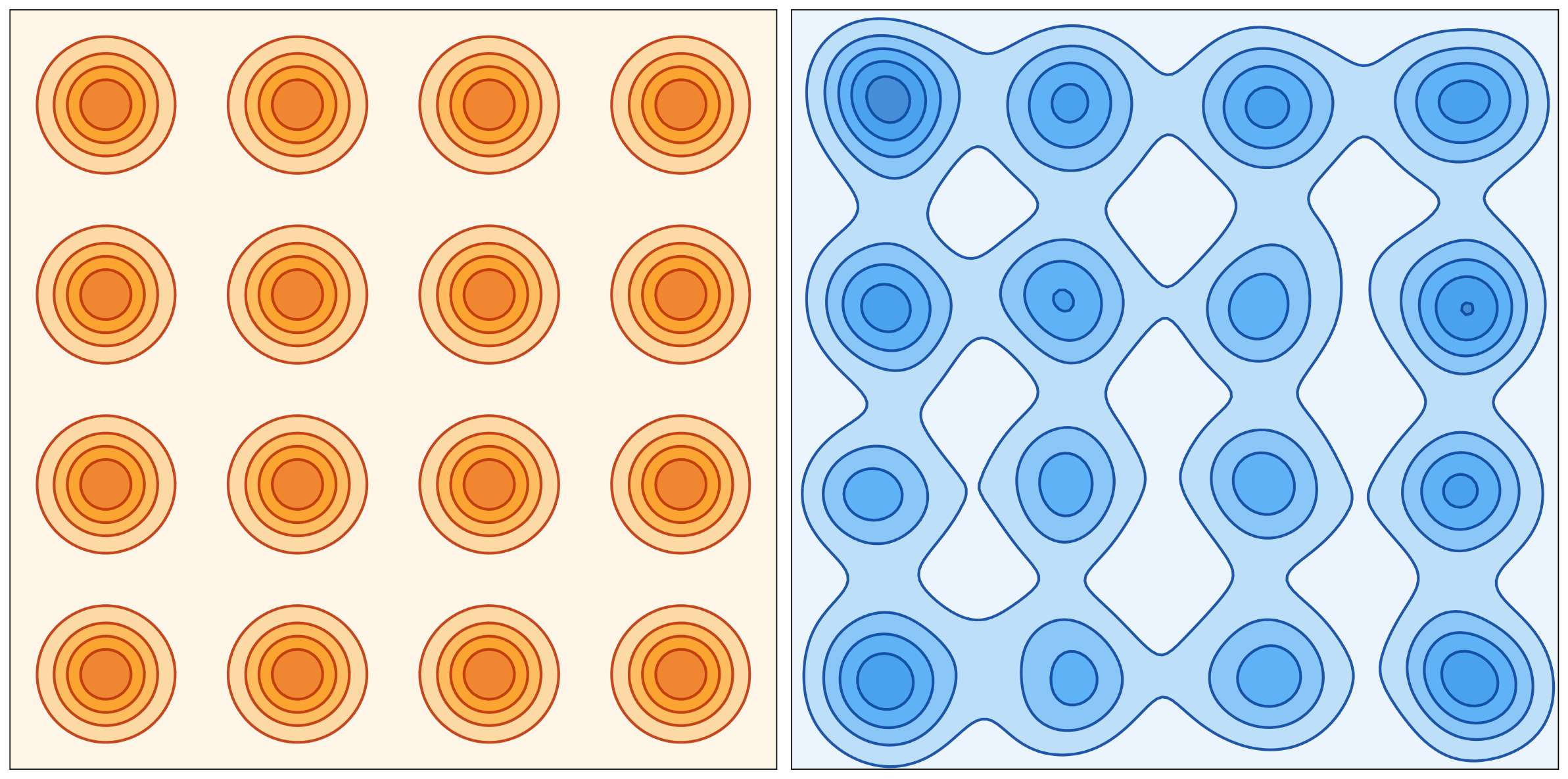} \\
\( 2a_t - u \) & \( 2a_t +   (2t-2b-1)\cdot v - (t-b) \cdot \texttt{jvp} \) & \( g(e, 0, 1) - e \) & \cmark & \xmark & $0.263$ & \includegraphics[width=0.95\linewidth]{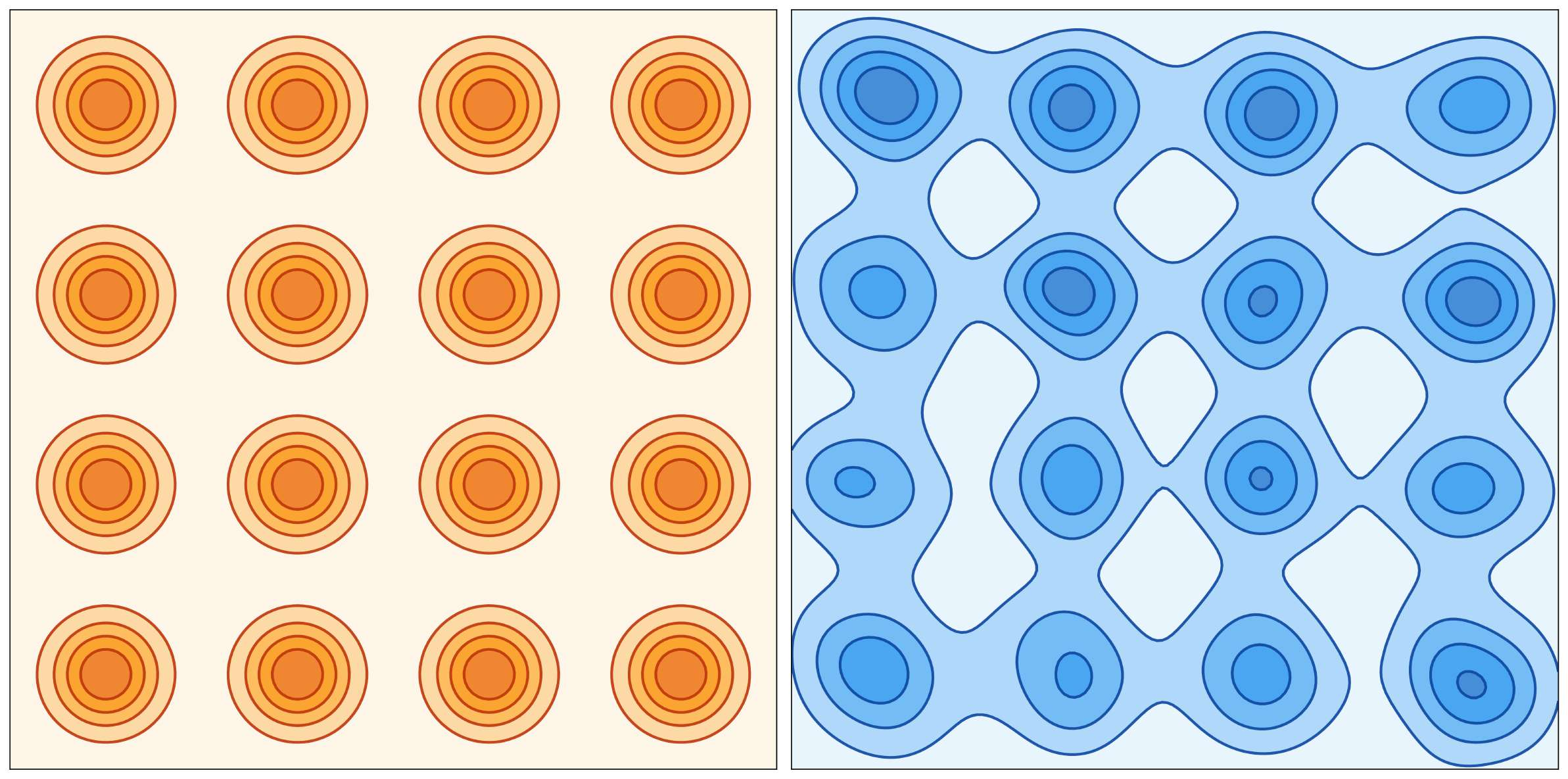} \\
\bottomrule
\end{tabular}
\begin{tablenotes}
\item \textsuperscript{1} \cmark\ indicates “Yes”; \xmark\ indicates “No”. 
\item \textsuperscript{2} Metric denotes the 2-Wasserstein distance (\(W_2\)), which quantifies the transport cost between predicted and ground-truth action  \\ distributions, offering a smooth and geometry-aware measure of distributional discrepancy.
\item \textsuperscript{3} \texttt{jvp} denotes $\texttt{jvp}(g,(a_t,b,t),(v,0,1))$ for simplicity. 
\item \textsuperscript{4} \( e \sim \mathcal{N}(0, I_d) \), \( a_t = (1 - t)a + te \), and \( v = e - a \).
\end{tablenotes}
\end{table*}

\clearpage

\begin{table*}[t]
\vspace{-30pt}
\caption{
\footnotesize
\textbf{Full offline RL results.}
We present the full results on the $73$ OGBench and D4RL tasks.   {(*)} indicates the default task in each environment.
The results are averaged over $8$ seeds ($4$ seeds for pixel-based tasks) unless otherwise mentioned.
}
\label{table:offline_full}
\centering
\vspace{5pt}
\resizebox{\textwidth}{!}
{
\begin{threeparttable}
\begin{tabular}{lccccccccccc}
\toprule
\multicolumn{1}{c}{} & \multicolumn{3}{c}{  {\texttt{Gaussian Policies}}} & \multicolumn{3}{c}{  {\texttt{Diffusion Policies}}} & \multicolumn{5}{c}{  {\texttt{Flow Policies}}} \\
\cmidrule(lr){2-4} \cmidrule(lr){5-7} \cmidrule(lr){8-12}
  {\texttt{Task}} &   {\texttt{BC}} &   {\texttt{IQL}} &   {\texttt{ReBRAC}} &   {\texttt{IDQL}} &   {\texttt{SRPO}} &   {\texttt{CAC}} &   {\texttt{FAWAC}} &   {\texttt{FBRAC}} &   {\texttt{IFQL}} &   {\texttt{FQL}} &   {\color{hiccup}\texttt{Ours}} \\
\midrule
  {\texttt{antmaze-large-navigate-singletask-task1-v0 (*)}} & $0$ {\tiny $\pm 0$} & $48$ {\tiny $\pm 9$} & $\mathbf{91}$ {\tiny $\pm 10$} & $0$ {\tiny $\pm 0$} & $0$ {\tiny $\pm 0$} & $42$ {\tiny $\pm 7$} & $1$ {\tiny $\pm 1$} & $70$ {\tiny $\pm 20$} & $24$ {\tiny $\pm 17$} & $80$ {\tiny $\pm 8$} & $\mathbf{91}$ {\tiny $\pm 2$} \\
  {\texttt{antmaze-large-navigate-singletask-task2-v0}} & $6$ {\tiny $\pm 3$} & $42$ {\tiny $\pm 2$} & $\mathbf{88}$ {\tiny $\pm 4$} & $14$ {\tiny $\pm 8$} & $4$ {\tiny $\pm 4$} & $1$ {\tiny $\pm 1$} & $0$ {\tiny $\pm 1$} & $35$ {\tiny $\pm 12$} & $8$ {\tiny $\pm 3$} & $57$ {\tiny $\pm 10$} & $53$ {\tiny $\pm 2$} \\
  {\texttt{antmaze-large-navigate-singletask-task3-v0}} & $29$ {\tiny $\pm 5$} & $72$ {\tiny $\pm 7$} & $51$ {\tiny $\pm 18$} & $26$ {\tiny $\pm 8$} & $3$ {\tiny $\pm 2$} & $49$ {\tiny $\pm 10$} & $12$ {\tiny $\pm 4$} & $83$ {\tiny $\pm 15$} & $52$ {\tiny $\pm 17$} & $\mathbf{93}$ {\tiny $\pm 3$} & $\mathbf{92}$ {\tiny $\pm 2$} \\
  {\texttt{antmaze-large-navigate-singletask-task4-v0}} & $8$ {\tiny $\pm 3$} & $51$ {\tiny $\pm 9$} & $\mathbf{84}$ {\tiny $\pm 7$} & $62$ {\tiny $\pm 25$} & $45$ {\tiny $\pm 19$} & $17$ {\tiny $\pm 6$} & $10$ {\tiny $\pm 3$} & $37$ {\tiny $\pm 18$} & $18$ {\tiny $\pm 8$} & $80$ {\tiny $\pm 4$} & $80$ {\tiny $\pm 4$} \\
  {\texttt{antmaze-large-navigate-singletask-task5-v0}} & $10$ {\tiny $\pm 3$} & $54$ {\tiny $\pm 22$} & $\mathbf{90}$ {\tiny $\pm 2$} & $2$ {\tiny $\pm 2$} & $1$ {\tiny $\pm 1$} & $55$ {\tiny $\pm 6$} & $9$ {\tiny $\pm 5$} & $76$ {\tiny $\pm 8$} & $38$ {\tiny $\pm 18$} & $83$ {\tiny $\pm 4$} & $\mathbf{87}$ {\tiny $\pm 2$} \\
\midrule
  {\texttt{antmaze-giant-navigate-singletask-task1-v0 (*)}} & $0$ {\tiny $\pm 0$} & $0$ {\tiny $\pm 0$} & $\mathbf{27}$ {\tiny $\pm 22$} & $0$ {\tiny $\pm 0$} & $0$ {\tiny $\pm 0$} & $0$ {\tiny $\pm 0$} & $0$ {\tiny $\pm 0$} & $0$ {\tiny $\pm 1$} & $0$ {\tiny $\pm 0$} & $4$ {\tiny $\pm 5$} & $0$ {\tiny $\pm 0$}\\
  {\texttt{antmaze-giant-navigate-singletask-task2-v0}} & $0$ {\tiny $\pm 0$} & $1$ {\tiny $\pm 1$} & $\mathbf{16}$ {\tiny $\pm 17$} & $0$ {\tiny $\pm 0$} & $0$ {\tiny $\pm 0$} & $0$ {\tiny $\pm 0$} & $0$ {\tiny $\pm 0$} & $4$ {\tiny $\pm 7$} & $0$ {\tiny $\pm 0$} & $9$ {\tiny $\pm 7$} & $0$ {\tiny $\pm 0$}\\
  {\texttt{antmaze-giant-navigate-singletask-task3-v0}} & $0$ {\tiny $\pm 0$} & $0$ {\tiny $\pm 0$} & $\mathbf{34}$ {\tiny $\pm 22$} & $0$ {\tiny $\pm 0$} & $0$ {\tiny $\pm 0$} & $0$ {\tiny $\pm 0$} & $0$ {\tiny $\pm 0$} & $0$ {\tiny $\pm 0$} & $0$ {\tiny $\pm 0$} & $0$ {\tiny $\pm 1$} & $0$ {\tiny $\pm 0$}\\
  {\texttt{antmaze-giant-navigate-singletask-task4-v0}} & $0$ {\tiny $\pm 0$} & $0$ {\tiny $\pm 0$} & $5$ {\tiny $\pm 12$} & $0$ {\tiny $\pm 0$} & $0$ {\tiny $\pm 0$} & $0$ {\tiny $\pm 0$} & $0$ {\tiny $\pm 0$} & $9$ {\tiny $\pm 4$} & $0$ {\tiny $\pm 0$} & $\mathbf{14}$ {\tiny $\pm 23$} & $0$ {\tiny $\pm 0$}\\
  {\texttt{antmaze-giant-navigate-singletask-task5-v0}} & $1$ {\tiny $\pm 1$} & $19$ {\tiny $\pm 7$} & $\mathbf{49}$ {\tiny $\pm 22$} & $0$ {\tiny $\pm 1$} & $0$ {\tiny $\pm 0$} & $0$ {\tiny $\pm 0$} & $0$ {\tiny $\pm 0$} & $6$ {\tiny $\pm 10$} & $13$ {\tiny $\pm 9$} & $16$ {\tiny $\pm 28$} & $0$ {\tiny $\pm 0$}\\
\midrule
  {\texttt{humanoidmaze-medium-navigate-singletask-task1-v0 (*)}} & $1$ {\tiny $\pm 0$} & $32$ {\tiny $\pm 7$} & $16$ {\tiny $\pm 9$} & $1$ {\tiny $\pm 1$} & $0$ {\tiny $\pm 0$} & $38$ {\tiny $\pm 19$} & $6$ {\tiny $\pm 2$} & $25$ {\tiny $\pm 8$} & $69$ {\tiny $\pm 19$} & $19$ {\tiny $\pm 12$} & $\mathbf{94}$ {\tiny $\pm 3$}\\
  {\texttt{humanoidmaze-medium-navigate-singletask-task2-v0}} & $1$ {\tiny $\pm 0$} & $41$ {\tiny $\pm 9$} & $18$ {\tiny $\pm 16$} & $1$ {\tiny $\pm 1$} & $1$ {\tiny $\pm 1$} & $47$ {\tiny $\pm 35$} & $40$ {\tiny $\pm 2$} & $76$ {\tiny $\pm 10$} & $85$ {\tiny $\pm 11$} & $94$ {\tiny $\pm 3$} & $\mathbf{97}$ {\tiny $\pm 1$}\\
  {\texttt{humanoidmaze-medium-navigate-singletask-task3-v0}} & $6$ {\tiny $\pm 2$} & $25$ {\tiny $\pm 5$} & $36$ {\tiny $\pm 13$} & $0$ {\tiny $\pm 1$} & $2$ {\tiny $\pm 1$} & $\mathbf{83}$ {\tiny $\pm 18$} & $19$ {\tiny $\pm 2$} & $27$ {\tiny $\pm 11$} & $49$ {\tiny $\pm 49$} & $74$ {\tiny $\pm 18$} & $14$ {\tiny $\pm 1$}\\
  {\texttt{humanoidmaze-medium-navigate-singletask-task4-v0}} & $0$ {\tiny $\pm 0$} & $0$ {\tiny $\pm 1$} & $\mathbf{15}$ {\tiny $\pm 16$} & $1$ {\tiny $\pm 1$} & $1$ {\tiny $\pm 1$} & $5$ {\tiny $\pm 4$} & $1$ {\tiny $\pm 1$} & $1$ {\tiny $\pm 2$} & $1$ {\tiny $\pm 1$} & $3$ {\tiny $\pm 4$} & $3$ {\tiny $\pm 2$}\\
  {\texttt{humanoidmaze-medium-navigate-singletask-task5-v0}} & $2$ {\tiny $\pm 1$} & $66$ {\tiny $\pm 4$} & $24$ {\tiny $\pm 20$} & $1$ {\tiny $\pm 1$} & $3$ {\tiny $\pm 3$} & $91$ {\tiny $\pm 5$} & $31$ {\tiny $\pm 7$} & $63$ {\tiny $\pm 9$} & $\mathbf{98}$ {\tiny $\pm 2$} & $97$ {\tiny $\pm 2$} & $\mathbf{98}$ {\tiny $\pm 1$}\\
\midrule
  {\texttt{humanoidmaze-large-navigate-singletask-task1-v0 (*)}} & $0$ {\tiny $\pm 0$} & $3$ {\tiny $\pm 1$} & $2$ {\tiny $\pm 1$} & $0$ {\tiny $\pm 0$} & $0$ {\tiny $\pm 0$} & $1$ {\tiny $\pm 1$} & $0$ {\tiny $\pm 0$} & $0$ {\tiny $\pm 1$} & $6$ {\tiny $\pm 2$} & $7$ {\tiny $\pm 6$} & $\mathbf{53}$ {\tiny $\pm 5$}\\
  {\texttt{humanoidmaze-large-navigate-singletask-task2-v0}} & $0$ {\tiny $\pm 0$} & $0$ {\tiny $\pm 0$} & $0$ {\tiny $\pm 0$} & $0$ {\tiny $\pm 0$} & $0$ {\tiny $\pm 0$} & $0$ {\tiny $\pm 0$} & $0$ {\tiny $\pm 0$} & $0$ {\tiny $\pm 0$} & $0$ {\tiny $\pm 0$} & $0$ {\tiny $\pm 0$} & $0$ {\tiny $\pm 0$}\\
  {\texttt{humanoidmaze-large-navigate-singletask-task3-v0}} & $1$ {\tiny $\pm 1$} & $7$ {\tiny $\pm 3$} & $8$ {\tiny $\pm 4$} & $3$ {\tiny $\pm 1$} & $1$ {\tiny $\pm 1$} & $2$ {\tiny $\pm 3$} & $1$ {\tiny $\pm 1$} & $10$ {\tiny $\pm 2$} & $\mathbf{48}$ {\tiny $\pm 10$} & $11$ {\tiny $\pm 7$} & $4$ {\tiny $\pm 1$}\\
  {\texttt{humanoidmaze-large-navigate-singletask-task4-v0}} & $1$ {\tiny $\pm 0$} & $1$ {\tiny $\pm 0$} & $1$ {\tiny $\pm 1$} & $0$ {\tiny $\pm 0$} & $0$ {\tiny $\pm 0$} & $0$ {\tiny $\pm 1$} & $0$ {\tiny $\pm 0$} & $0$ {\tiny $\pm 0$} & $1$ {\tiny $\pm 1$} & $\mathbf{2}$ {\tiny $\pm 3$} & $\mathbf{2}$ {\tiny $\pm 1$}\\
  {\texttt{humanoidmaze-large-navigate-singletask-task5-v0}} & $0$ {\tiny $\pm 1$} & $1$ {\tiny $\pm 1$} & $2$ {\tiny $\pm 2$} & $0$ {\tiny $\pm 0$} & $0$ {\tiny $\pm 0$} & $0$ {\tiny $\pm 0$} & $0$ {\tiny $\pm 0$} & $1$ {\tiny $\pm 1$} & $0$ {\tiny $\pm 0$} & $1$ {\tiny $\pm 3$} & $\mathbf{38}$ {\tiny $\pm 3$}\\
\midrule
  {\texttt{antsoccer-arena-navigate-singletask-task1-v0}} & $2$ {\tiny $\pm 1$} & $14$ {\tiny $\pm 5$} & $0$ {\tiny $\pm 0$} & $44$ {\tiny $\pm 12$} & $2$ {\tiny $\pm 1$} & $1$ {\tiny $\pm 3$} & $22$ {\tiny $\pm 2$} & $17$ {\tiny $\pm 3$} & $61$ {\tiny $\pm 25$} & $77$ {\tiny $\pm 4$} & $\mathbf{83}$ {\tiny $\pm 4$}\\
  {\texttt{antsoccer-arena-navigate-singletask-task2-v0}} & $2$ {\tiny $\pm 2$} & $17$ {\tiny $\pm 7$} & $0$ {\tiny $\pm 1$} & $15$ {\tiny $\pm 12$} & $3$ {\tiny $\pm 1$} & $0$ {\tiny $\pm 0$} & $8$ {\tiny $\pm 1$} & $8$ {\tiny $\pm 2$} & $75$ {\tiny $\pm 3$} & $\mathbf{88}$ {\tiny $\pm 3$} & $73$ {\tiny $\pm 3$}\\
  {\texttt{antsoccer-arena-navigate-singletask-task3-v0}} & $0$ {\tiny $\pm 0$} & $6$ {\tiny $\pm 4$} & $0$ {\tiny $\pm 0$} & $0$ {\tiny $\pm 0$} & $0$ {\tiny $\pm 0$} & $8$ {\tiny $\pm 19$} & $11$ {\tiny $\pm 5$} & $16$ {\tiny $\pm 3$} & $14$ {\tiny $\pm 22$} & $\mathbf{61}$ {\tiny $\pm 6$} & $\mathbf{60}$ {\tiny $\pm 4$}\\
  {\texttt{antsoccer-arena-navigate-singletask-task4-v0 (*)}} & $1$ {\tiny $\pm 0$} & $3$ {\tiny $\pm 2$} & $0$ {\tiny $\pm 0$} & $0$ {\tiny $\pm 1$} & $0$ {\tiny $\pm 0$} & $0$ {\tiny $\pm 0$} & $12$ {\tiny $\pm 3$} & $24$ {\tiny $\pm 4$} & $16$ {\tiny $\pm 9$} & $39$ {\tiny $\pm 6$} & $\mathbf{43}$ {\tiny $\pm 2$}\\
  {\texttt{antsoccer-arena-navigate-singletask-task5-v0}} & $0$ {\tiny $\pm 0$} & $2$ {\tiny $\pm 2$} & $0$ {\tiny $\pm 0$} & $0$ {\tiny $\pm 0$} & $0$ {\tiny $\pm 0$} & $0$ {\tiny $\pm 0$} & $9$ {\tiny $\pm 2$} & $15$ {\tiny $\pm 4$} & $0$ {\tiny $\pm 1$} & $36$ {\tiny $\pm 9$} & $\mathbf{51}$ {\tiny $\pm 4$}\\
\midrule
  {\texttt{cube-single-play-singletask-task1-v0}} & $10$ {\tiny $\pm 5$} & $88$ {\tiny $\pm 3$} & $89$ {\tiny $\pm 5$} & $95$ {\tiny $\pm 2$} & $89$ {\tiny $\pm 7$} & $77$ {\tiny $\pm 28$} & $81$ {\tiny $\pm 9$} & $73$ {\tiny $\pm 33$} & $79$ {\tiny $\pm 4$} & $\mathbf{97}$ {\tiny $\pm 2$} & $\mathbf{96}$ {\tiny $\pm 1$}\\
  {\texttt{cube-single-play-singletask-task2-v0 (*)}} & $3$ {\tiny $\pm 1$} & $85$ {\tiny $\pm 8$} & $92$ {\tiny $\pm 4$} & $96$ {\tiny $\pm 2$} & $82$ {\tiny $\pm 16$} & $80$ {\tiny $\pm 30$} & $81$ {\tiny $\pm 9$} & $83$ {\tiny $\pm 13$} & $73$ {\tiny $\pm 3$} & $\mathbf{97}$ {\tiny $\pm 2$} & $90$ {\tiny $\pm 5$}\\
  {\texttt{cube-single-play-singletask-task3-v0}} & $9$ {\tiny $\pm 3$} & $91$ {\tiny $\pm 5$} & $93$ {\tiny $\pm 3$} & $\mathbf{99}$ {\tiny $\pm 1$} & $96$ {\tiny $\pm 2$} & $98$ {\tiny $\pm 1$} & $87$ {\tiny $\pm 4$} & $82$ {\tiny $\pm 12$} & $88$ {\tiny $\pm 4$} & $98$ {\tiny $\pm 2$} & $\mathbf{99}$ {\tiny $\pm 1$}\\
  {\texttt{cube-single-play-singletask-task4-v0}} & $2$ {\tiny $\pm 1$} & $73$ {\tiny $\pm 6$} & $92$ {\tiny $\pm 3$} & $93$ {\tiny $\pm 4$} & $70$ {\tiny $\pm 18$} & $91$ {\tiny $\pm 2$} & $79$ {\tiny $\pm 6$} & $79$ {\tiny $\pm 20$} & $79$ {\tiny $\pm 6$} & $94$ {\tiny $\pm 3$} & $\mathbf{95}$ {\tiny $\pm 2$}\\
  {\texttt{cube-single-play-singletask-task5-v0}} & $3$ {\tiny $\pm 3$} & $78$ {\tiny $\pm 9$} & $87$ {\tiny $\pm 8$} & $90$ {\tiny $\pm 6$} & $61$ {\tiny $\pm 12$} & $80$ {\tiny $\pm 20$} & $78$ {\tiny $\pm 10$} & $76$ {\tiny $\pm 33$} & $77$ {\tiny $\pm 7$} & $93$ {\tiny $\pm 3$} & $\mathbf{95}$ {\tiny $\pm 2$}\\
\midrule
  {\texttt{cube-double-play-singletask-task1-v0}} & $8$ {\tiny $\pm 3$} & $27$ {\tiny $\pm 5$} & $45$ {\tiny $\pm 6$} & $39$ {\tiny $\pm 19$} & $7$ {\tiny $\pm 6$} & $21$ {\tiny $\pm 8$} & $21$ {\tiny $\pm 7$} & $\mathbf{47}$ {\tiny $\pm 11$} & $35$ {\tiny $\pm 9$} & $61$ {\tiny $\pm 9$} & $11$ {\tiny $\pm 5$}\\
  {\texttt{cube-double-play-singletask-task2-v0 (*)}} & $0$ {\tiny $\pm 0$} & $1$ {\tiny $\pm 1$} & $7$ {\tiny $\pm 3$} & $16$ {\tiny $\pm 10$} & $0$ {\tiny $\pm 0$} & $2$ {\tiny $\pm 2$} & $2$ {\tiny $\pm 1$} & $22$ {\tiny $\pm 12$} & $9$ {\tiny $\pm 5$} & $\mathbf{36}$ {\tiny $\pm 6$} & $2$ {\tiny $\pm 1$}\\
  {\texttt{cube-double-play-singletask-task3-v0}} & $0$ {\tiny $\pm 0$} & $0$ {\tiny $\pm 0$} & $4$ {\tiny $\pm 1$} & $17$ {\tiny $\pm 8$} & $0$ {\tiny $\pm 1$} & $3$ {\tiny $\pm 1$} & $1$ {\tiny $\pm 1$} & $4$ {\tiny $\pm 2$} & $8$ {\tiny $\pm 5$} & $\mathbf{22}$ {\tiny $\pm 5$} & $2$ {\tiny $\pm 1$}\\
  {\texttt{cube-double-play-singletask-task4-v0}} & $0$ {\tiny $\pm 0$} & $0$ {\tiny $\pm 0$} & $1$ {\tiny $\pm 1$} & $0$ {\tiny $\pm 1$} & $0$ {\tiny $\pm 0$} & $0$ {\tiny $\pm 1$} & $0$ {\tiny $\pm 0$} & $0$ {\tiny $\pm 1$} & $1$ {\tiny $\pm 1$} & $\mathbf{5}$ {\tiny $\pm 2$} & $0$ {\tiny $\pm 0$}\\
  {\texttt{cube-double-play-singletask-task5-v0}} & $0$ {\tiny $\pm 0$} & $4$ {\tiny $\pm 3$} & $4$ {\tiny $\pm 2$} & $1$ {\tiny $\pm 1$} & $0$ {\tiny $\pm 0$} & $3$ {\tiny $\pm 2$} & $2$ {\tiny $\pm 1$} & $2$ {\tiny $\pm 2$} & $17$ {\tiny $\pm 6$} & $\mathbf{19}$ {\tiny $\pm 10$} & $1$ {\tiny $\pm 1$}\\
\midrule
  {\texttt{scene-play-singletask-task1-v0}} & $19$ {\tiny $\pm 6$} & $94$ {\tiny $\pm 3$} & $95$ {\tiny $\pm 2$} & $100$ {\tiny $\pm 0$} & $94$ {\tiny $\pm 4$} & $100$ {\tiny $\pm 1$} & $87$ {\tiny $\pm 8$} & $96$ {\tiny $\pm 8$} & $98$ {\tiny $\pm 3$} & $\mathbf{100}$ {\tiny $\pm 0$} & $\mathbf{99}$ {\tiny $\pm 1$}\\
  {\texttt{scene-play-singletask-task2-v0 (*)}} & $1$ {\tiny $\pm 1$} & $12$ {\tiny $\pm 3$} & $50$ {\tiny $\pm 13$} & $33$ {\tiny $\pm 14$} & $2$ {\tiny $\pm 2$} & $50$ {\tiny $\pm 40$} & $18$ {\tiny $\pm 8$} & $46$ {\tiny $\pm 10$} & $0$ {\tiny $\pm 0$} & $76$ {\tiny $\pm 9$} & $\mathbf{91}$ {\tiny $\pm 1$}\\
  {\texttt{scene-play-singletask-task3-v0}} & $1$ {\tiny $\pm 1$} & $32$ {\tiny $\pm 7$} & $55$ {\tiny $\pm 16$} & $94$ {\tiny $\pm 4$} & $4$ {\tiny $\pm 4$} & $49$ {\tiny $\pm 16$} & $38$ {\tiny $\pm 9$} & $78$ {\tiny $\pm 14$} & $54$ {\tiny $\pm 19$} & $\mathbf{98}$ {\tiny $\pm 1$} & $\mathbf{98}$ {\tiny $\pm 1$}\\
  {\texttt{scene-play-singletask-task4-v0}} & $2$ {\tiny $\pm 2$} & $0$ {\tiny $\pm 1$} & $3$ {\tiny $\pm 3$} & $4$ {\tiny $\pm 3$} & $0$ {\tiny $\pm 0$} & $0$ {\tiny $\pm 0$} & $6$ {\tiny $\pm 1$} & $4$ {\tiny $\pm 4$} & $0$ {\tiny $\pm 0$} & $5$ {\tiny $\pm 1$} & $\mathbf{11}$ {\tiny $\pm 3$}\\
  {\texttt{scene-play-singletask-task5-v0}} & $0$ {\tiny $\pm 0$} & $0$ {\tiny $\pm 0$} & $0$ {\tiny $\pm 0$} & $0$ {\tiny $\pm 0$} & $0$ {\tiny $\pm 0$} & $0$ {\tiny $\pm 0$} & $0$ {\tiny $\pm 0$} & $0$ {\tiny $\pm 0$} & $0$ {\tiny $\pm 0$} & $0$ {\tiny $\pm 0$} & $0$ {\tiny $\pm 0$}\\
\midrule
  {\texttt{puzzle-3x3-play-singletask-task1-v0}} & $5$ {\tiny $\pm 2$} & $33$ {\tiny $\pm 6$} & $97$ {\tiny $\pm 4$} & $52$ {\tiny $\pm 12$} & $89$ {\tiny $\pm 5$} & $97$ {\tiny $\pm 2$} & $25$ {\tiny $\pm 9$} & $63$ {\tiny $\pm 19$} & $94$ {\tiny $\pm 3$} & $90$ {\tiny $\pm 4$} & $\mathbf{99}$ {\tiny $\pm 1$}\\
  {\texttt{puzzle-3x3-play-singletask-task2-v0}} & $1$ {\tiny $\pm 1$} & $4$ {\tiny $\pm 3$} & $1$ {\tiny $\pm 1$} & $0$ {\tiny $\pm 1$} & $0$ {\tiny $\pm 1$} & $0$ {\tiny $\pm 0$} & $4$ {\tiny $\pm 2$} & $2$ {\tiny $\pm 2$} & $1$ {\tiny $\pm 2$} & $16$ {\tiny $\pm 5$} & $\mathbf{88}$ {\tiny $\pm 6$}\\
  {\texttt{puzzle-3x3-play-singletask-task3-v0}} & $1$ {\tiny $\pm 1$} & $3$ {\tiny $\pm 2$} & $3$ {\tiny $\pm 1$} & $0$ {\tiny $\pm 0$} & $0$ {\tiny $\pm 0$} & $0$ {\tiny $\pm 0$} & $1$ {\tiny $\pm 0$} & $1$ {\tiny $\pm 1$} & $0$ {\tiny $\pm 0$} & $10$ {\tiny $\pm 3$} & $\mathbf{26}$ {\tiny $\pm 8$}\\
  {\texttt{puzzle-3x3-play-singletask-task4-v0 (*)}} & $1$ {\tiny $\pm 1$} & $2$ {\tiny $\pm 1$} & $2$ {\tiny $\pm 1$} & $0$ {\tiny $\pm 0$} & $0$ {\tiny $\pm 0$} & $0$ {\tiny $\pm 0$} & $1$ {\tiny $\pm 1$} & $2$ {\tiny $\pm 2$} & $0$ {\tiny $\pm 0$} & $16$ {\tiny $\pm 5$} & $\mathbf{67}$ {\tiny $\pm 12$}\\
  {\texttt{puzzle-3x3-play-singletask-task5-v0}} & $1$ {\tiny $\pm 0$} & $3$ {\tiny $\pm 2$} & $5$ {\tiny $\pm 3$} & $0$ {\tiny $\pm 0$} & $0$ {\tiny $\pm 0$} & $0$ {\tiny $\pm 0$} & $1$ {\tiny $\pm 1$} & $2$ {\tiny $\pm 2$} & $0$ {\tiny $\pm 0$} & $16$ {\tiny $\pm 3$} & $\mathbf{49}$ {\tiny $\pm 14$}\\
\midrule
  {\texttt{puzzle-4x4-play-singletask-task1-v0}} & $1$ {\tiny $\pm 1$} & $12$ {\tiny $\pm 2$} & $26$ {\tiny $\pm 4$} & $48$ {\tiny $\pm 5$} & $24$ {\tiny $\pm 9$} & $44$ {\tiny $\pm 10$} & $1$ {\tiny $\pm 2$} & $32$ {\tiny $\pm 9$} & $49$ {\tiny $\pm 9$} & $34$ {\tiny $\pm 8$} & $\mathbf{68}$ {\tiny $\pm 8$}\\
  {\texttt{puzzle-4x4-play-singletask-task2-v0}} & $0$ {\tiny $\pm 0$} & $7$ {\tiny $\pm 4$} & $12$ {\tiny $\pm 4$} & $14$ {\tiny $\pm 5$} & $0$ {\tiny $\pm 1$} & $0$ {\tiny $\pm 0$} & $0$ {\tiny $\pm 1$} & $5$ {\tiny $\pm 3$} & $4$ {\tiny $\pm 4$} & $\mathbf{16}$ {\tiny $\pm 5$} & $\mathbf{16}$ {\tiny $\pm 3$}\\
  {\texttt{puzzle-4x4-play-singletask-task3-v0}} & $0$ {\tiny $\pm 0$} & $9$ {\tiny $\pm 3$} & $15$ {\tiny $\pm 3$} & $34$ {\tiny $\pm 5$} & $21$ {\tiny $\pm 10$} & $29$ {\tiny $\pm 12$} & $1$ {\tiny $\pm 1$} & $20$ {\tiny $\pm 10$} & $50$ {\tiny $\pm 14$} & $18$ {\tiny $\pm 5$} & $\mathbf{75}$ {\tiny $\pm 10$}\\
  {\texttt{puzzle-4x4-play-singletask-task4-v0 (*)}} & $0$ {\tiny $\pm 0$} & $5$ {\tiny $\pm 2$} & $10$ {\tiny $\pm 3$} & $\mathbf{26}$ {\tiny $\pm 6$} & $7$ {\tiny $\pm 4$} & $1$ {\tiny $\pm 1$} & $0$ {\tiny $\pm 0$} & $5$ {\tiny $\pm 1$} & $21$ {\tiny $\pm 11$} & $11$ {\tiny $\pm 3$} & $15$ {\tiny $\pm 3$}\\
  {\texttt{puzzle-4x4-play-singletask-task5-v0}} & $0$ {\tiny $\pm 0$} & $4$ {\tiny $\pm 1$} & $7$ {\tiny $\pm 3$} & $24$ {\tiny $\pm 11$} & $1$ {\tiny $\pm 1$} & $0$ {\tiny $\pm 0$} & $0$ {\tiny $\pm 1$} & $4$ {\tiny $\pm 3$} & $2$ {\tiny $\pm 2$} & $7$ {\tiny $\pm 3$} & $\mathbf{24}$ {\tiny $\pm 10$}\\
\midrule
  {\texttt{antmaze-umaze-v2}} & $55$ & $77$ & $98$ & $94$ & $97$ & $66$ {\tiny $\pm 5$} & $90$ {\tiny $\pm 6$} & $94$ {\tiny $\pm 3$} & $92$ {\tiny $\pm 6$} & $96$ {\tiny $\pm 2$} & $\mathbf{98}$ {\tiny $\pm 1$}\\
  {\texttt{antmaze-umaze-diverse-v2}} & $47$ & $54$ & $84$ & $80$ & $82$ & $66$ {\tiny $\pm 11$} & $55$ {\tiny $\pm 7$} & $82$ {\tiny $\pm 9$} & $62$ {\tiny $\pm 12$} & $\mathbf{89}$ {\tiny $\pm 5$} & $79$ {\tiny $\pm 2$}\\
  {\texttt{antmaze-medium-play-v2}} & $0$ & $66$ & $\mathbf{90}$ & $84$ & $81$ & $49$ {\tiny $\pm 24$} & $52$ {\tiny $\pm 12$} & $77$ {\tiny $\pm 7$} & $56$ {\tiny $\pm 15$} & $78$ {\tiny $\pm 7$} & $\mathbf{86}$ {\tiny $\pm 2$}\\
  {\texttt{antmaze-medium-diverse-v2}} & $1$ & $74$ & $84$ & $\mathbf{85}$ & $75$ & $0$ {\tiny $\pm 1$} & $44$ {\tiny $\pm 15$} & $77$ {\tiny $\pm 6$} & $60$ {\tiny $\pm 25$} & $71$ {\tiny $\pm 13$} & $78$ {\tiny $\pm 2$}\\
  {\texttt{antmaze-large-play-v2}} & $0$ & $42$ & $52$ & $64$ & $54$ & $0$ {\tiny $\pm 0$} & $10$ {\tiny $\pm 6$} & $32$ {\tiny $\pm 21$} & $55$ {\tiny $\pm 9$} & $\mathbf{84}$ {\tiny $\pm 7$} & $80$ {\tiny $\pm 3$}\\
  {\texttt{antmaze-large-diverse-v2}} & $0$ & $30$ & $64$ & $68$ & $54$ & $0$ {\tiny $\pm 0$} & $16$ {\tiny $\pm 10$} & $20$ {\tiny $\pm 17$} & $64$ {\tiny $\pm 8$} & $\mathbf{83}$ {\tiny $\pm 4$} & $76$ {\tiny $\pm 1$}\\
\midrule
  {\texttt{pen-human-v1}} & $71$ & $78$ & $103$ & $76$ {\tiny $\pm 10$} & $69$ {\tiny $\pm 7$} & $64$ {\tiny $\pm 8$} & $67$ {\tiny $\pm 5$} & $77$ {\tiny $\pm 7$} & $71$ {\tiny $\pm 12$} & $53$ {\tiny $\pm 6$} & $\mathbf{84}$ {\tiny $\pm 7$}\\
  {\texttt{pen-cloned-v1}} & $52$ & $83$ & $\mathbf{103}$ & $64$ {\tiny $\pm 7$} & $61$ {\tiny $\pm 7$} & $56$ {\tiny $\pm 10$} & $62$ {\tiny $\pm 10$} & $67$ {\tiny $\pm 9$} & $80$ {\tiny $\pm 11$} & $74$ {\tiny $\pm 11$} & $79$ {\tiny $\pm 3$}\\
  {\texttt{pen-expert-v1}} & $110$ & $128$ & $\mathbf{152}$ & $140$ {\tiny $\pm 6$} & $134$ {\tiny $\pm 4$} & $103$ {\tiny $\pm 9$} & $118$ {\tiny $\pm 6$} & $119$ {\tiny $\pm 7$} & $139$ {\tiny $\pm 5$} & $142$ {\tiny $\pm 6$} & $135$ {\tiny $\pm 3$}\\
  {\texttt{door-human-v1}} & $2$ & $3$ & $0$ & $\mathbf{6}$ {\tiny $\pm 2$} & $3$ {\tiny $\pm 3$} & $5$ {\tiny $\pm 2$} & $2$ {\tiny $\pm 1$} & $4$ {\tiny $\pm 2$} & $7$ {\tiny $\pm 2$} & $0$ {\tiny $\pm 0$} & $1$ {\tiny $\pm 1$}\\
  {\texttt{door-cloned-v1}} & $0$ & $3$ & $0$ & $0$ {\tiny $\pm 0$} & $0$ {\tiny $\pm 0$} & $1$ {\tiny $\pm 0$} & $0$ {\tiny $\pm 1$} & $0$ {\tiny $\pm 0$} & $2$ {\tiny $\pm 2$} & $2$ {\tiny $\pm 1$} & $\mathbf{3}$ {\tiny $\pm 1$}\\
  {\texttt{door-expert-v1}} & $105$ & $\mathbf{107}$ & $106$ & $105$ {\tiny $\pm 1$} & $105$ {\tiny $\pm 0$} & $98$ {\tiny $\pm 3$} & $103$ {\tiny $\pm 1$} & $104$ {\tiny $\pm 1$} & $104$ {\tiny $\pm 2$} & $104$ {\tiny $\pm 1$} & $104$ {\tiny $\pm 1$}\\
  {\texttt{hammer-human-v1}} & $3$ & $2$ & $0$ & $2$ {\tiny $\pm 1$} & $1$ {\tiny $\pm 1$} & $2$ {\tiny $\pm 0$} & $2$ {\tiny $\pm 1$} & $2$ {\tiny $\pm 1$} & $\mathbf{3}$ {\tiny $\pm 1$} & $1$ {\tiny $\pm 1$} & $2$ {\tiny $\pm 1$}\\
  {\texttt{hammer-cloned-v1}} & $1$ & $2$ & $5$ & $2$ {\tiny $\pm 1$} & $2$ {\tiny $\pm 1$} & $1$ {\tiny $\pm 1$} & $1$ {\tiny $\pm 0$} & $2$ {\tiny $\pm 1$} & $2$ {\tiny $\pm 1$} & $11$ {\tiny $\pm 9$} & $\mathbf{10}$ {\tiny $\pm 5$}\\
  {\texttt{hammer-expert-v1}} & $127$ & $129$ & $\mathbf{134}$ & $125$ {\tiny $\pm 4$} & $127$ {\tiny $\pm 0$} & $92$ {\tiny $\pm 11$} & $118$ {\tiny $\pm 3$} & $119$ {\tiny $\pm 9$} & $117$ {\tiny $\pm 9$} & $125$ {\tiny $\pm 3$} & $120$ {\tiny $\pm 3$}\\
  {\texttt{relocate-human-v1}} & $0$ & $0$ & $0$ & $0$ {\tiny $\pm 0$} & $0$ {\tiny $\pm 0$} & $0$ {\tiny $\pm 0$} & $0$ {\tiny $\pm 0$} & $0$ {\tiny $\pm 0$} & $0$ {\tiny $\pm 0$} & $0$ {\tiny $\pm 0$} & $0$ {\tiny $\pm 0$}\\
  {\texttt{relocate-cloned-v1}} & $0$ & $0$ & $\mathbf{2}$ & $0$ {\tiny $\pm 0$} & $0$ {\tiny $\pm 0$} & $0$ {\tiny $\pm 0$} & $0$ {\tiny $\pm 0$} & $1$ {\tiny $\pm 1$} & $0$ {\tiny $\pm 0$} & $0$ {\tiny $\pm 0$} & $1$ {\tiny $\pm 1$}\\
  {\texttt{relocate-expert-v1}} & $\mathbf{108}$ & $106$ & $\mathbf{108}$ & $107$ {\tiny $\pm 1$} & $106$ {\tiny $\pm 2$} & $93$ {\tiny $\pm 6$} & $105$ {\tiny $\pm 3$} & $105$ {\tiny $\pm 2$} & $104$ {\tiny $\pm 3$} & $107$ {\tiny $\pm 1$} & $\mathbf{108}$ {\tiny $\pm 1$}\\
\midrule
  {\texttt{visual-cube-single-play-singletask-task1-v0 (1)}} & $-$ & $70$ {\tiny $\pm 12$} & $\mathbf{83}$ {\tiny $\pm 6$} & $-$ & $-$ & $-$ & $-$ & $55$ {\tiny $\pm 8$} & $49$ {\tiny $\pm 7$} & $81$ {\tiny $\pm 12$} & $49$ {\tiny $\pm 1$}\\
  {\texttt{visual-cube-double-play-singletask-task1-v0 (1)}} & $-$ & $\mathbf{34}$ {\tiny $\pm 23$} & $4$ {\tiny $\pm 4$} & $-$ & $-$ & $-$ & $-$ & $6$ {\tiny $\pm 2$} & $8$ {\tiny $\pm 6$} & $21$ {\tiny $\pm 11$} & $2$ {\tiny $\pm 1$}\\
  {\texttt{visual-scene-play-singletask-task1-v0 (1)}} & $-$ & $97$ {\tiny $\pm 2$} & $98$ {\tiny $\pm 4$} & $-$ & $-$ & $-$ & $-$ & $46$ {\tiny $\pm 4$} & $86$ {\tiny $\pm 10$} & $98$ {\tiny $\pm 3$} & $\mathbf{95}$ {\tiny $\pm 1$}\\
  {\texttt{visual-puzzle-3x3-play-singletask-task1-v0 (1)}} & $-$ & $7$ {\tiny $\pm 15$} & $88$ {\tiny $\pm 4$} & $-$ & $-$ & $-$ & $-$ & $7$ {\tiny $\pm 2$} & $\mathbf{100}$ {\tiny $\pm 0$} & $94$ {\tiny $\pm 1$} & $88$ {\tiny $\pm 4$}\\
  {\texttt{visual-puzzle-4x4-play-singletask-task1-v0 (1)}} & $-$ & $0$ {\tiny $\pm 0$} & $26$ {\tiny $\pm 6$} & $-$ & $-$ & $-$ & $-$ & $0$ {\tiny $\pm 0$} & $8$ {\tiny $\pm 15$} & $\mathbf{33}$ {\tiny $\pm 6$} & $25$ {\tiny $\pm 3$}\\
\bottomrule
\end{tabular}
\end{threeparttable}
}
\end{table*}

\clearpage
\subsection{C.\quad Additional Results}
\begin{figure}[ht]
  \centering
  \includegraphics[width=1\textwidth]{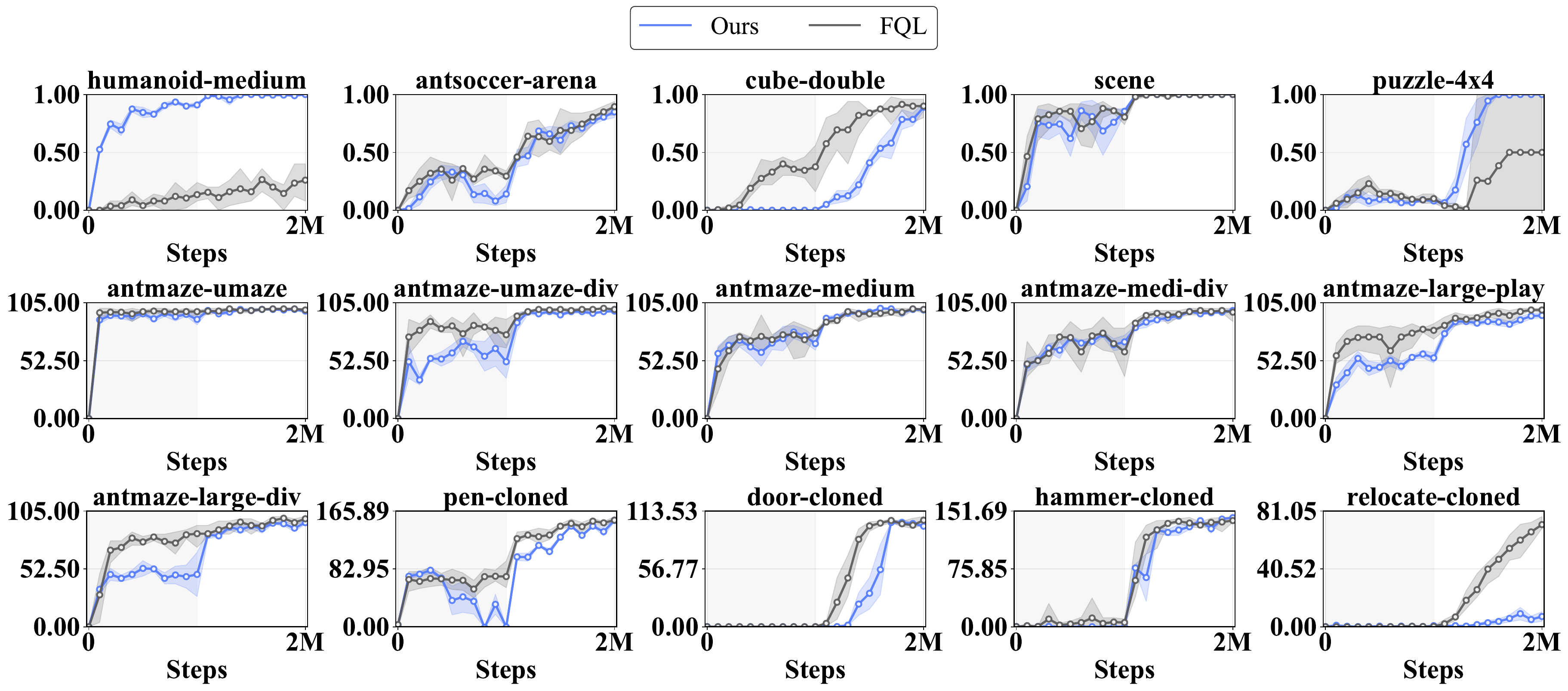} % 左 下 右 上
  \caption{Offline-to-online RL results. Online fine-tuning starts at 1M steps. The results are averaged over 8 seeds unless otherwise mentioned.}
  \label{fig:online_15}
\end{figure}

\begin{figure}[ht]
  \centering
  \includegraphics[width=1\textwidth]{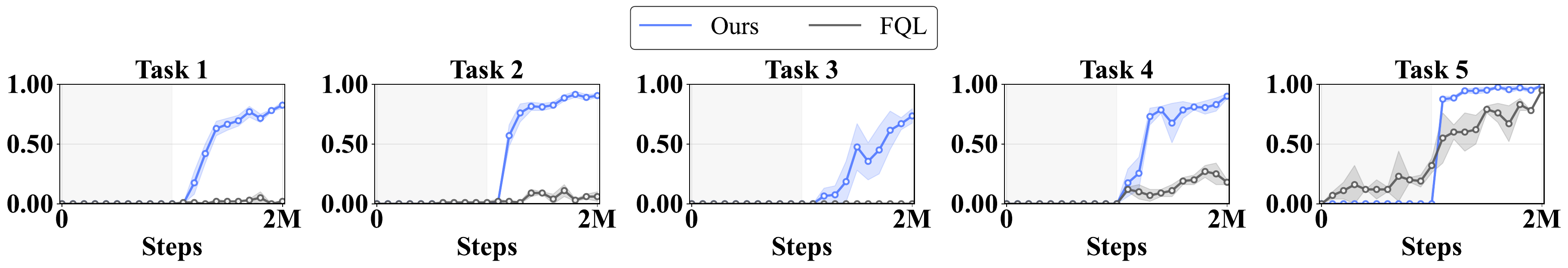} % 左 下 右 上
\caption{Offline-to-online RL results on \texttt{antmaze-giant-navigate-singletask}. Although our method starts from suboptimal offline performance, it exhibits rapid adaptation and consistently surpasses FQL across all challenging tasks during the online fine-tuning phase, demonstrating strong online learning capability.}

  \label{fig:online_antmaze_giant}
\end{figure}

\subsection{D\quad Implementation Details.}

\paragraph{Critic network architectures.}
Following FQL \cite{fql}, we employ $[512, 512, 512, 512]$-sized multi-layer perceptrons (MLPs) for critic function networks.
To improve training stability, we also apply layer normalization to the value networks.

\paragraph{Actor Network architectures.}
Since our method heavily relies on accurate modeling of the time variable, particularly in the context of Jacobian-vector product (JVP) computations, we choose DiT as the underlying network architecture.
We use a shallow Transformer-based policy adapted from DiT \cite{dit}, with 3 layers, 2 attention heads, and a hidden size of $256$. A feature embedding maps 1D observation-action inputs, and the network includes time and positional embeddings, residual mixing (scale 0.1). Specifically, we apply gradient clipping (norm 1.0) and zero-initialised final projection for stability.

\paragraph{Image processing.}
For pixel-based environments, we adopt a lightweight variant of the IMPALA encoder and apply random-shift augmentation with a probability of $0.5$, following the official OGBench implementation~\citep{ogbench}.

\paragraph{Training and evaluation.}
We train all networks for $1$M gradient steps on state-based OGBench tasks, and for $500$K steps on both D4RL and pixel-based OGBench tasks. Evaluation is performed every $100$K steps using $50$ rollout episodes.

For OGBench, we follow the official evaluation protocol~\citep{ogbench}, reporting the average success rate across the final three evaluation epochs: $800$K, $900$K, and $1$M for state-based tasks; $300$K, $400$K, and $500$K for pixel-based tasks.

For D4RL, we report the performance at the final training step. In offline-to-online RL experiments (\ref{table:offlinetoonline}), we report results at both $1$M and $2$M gradient steps.

\paragraph{Loss Metrics}

To define the regression loss for the target function $g_\theta$, we adopt a powered $\ell_2$ loss of the form:
\begin{equation}
    \mathcal{L}_\gamma = \|\Delta\|_2^{2\gamma}, \quad \text{where} \quad \Delta = g_\theta - g_{\mathrm{tgt}}, \ \gamma > 0.
\end{equation}
Following~\cite{iTCM,meanflow}, its gradient can be written as:
\begin{equation}
    \frac{d}{d\theta} \mathcal{L}_\gamma = \gamma \left( \|\Delta\|_2^2 \right)^{\gamma - 1} \cdot \frac{d}{d\theta} \|\Delta\|_2^2,
\end{equation}
which corresponds to an adaptively weighted squared $\ell_2$ loss. We adopt the following stabilised weight formulation:
\begin{equation}
    w = \frac{1}{\left(\|\Delta\|_2^2 + c\right)^p}, \quad \text{where} \quad p = 1 - \gamma,\ c > 0,
\end{equation}
and define the final loss as:
\begin{equation}
    \mathcal{L}_{\text{final}} = \text{sg}(w) \cdot \|\Delta\|_2^2.
\end{equation}
where $\text{sg}(\cdot)$ denotes the stop-gradient operator to prevent gradients from flowing through the adaptive weight. As shown in MeanFlow~\cite{meanflow}, the parameter $p$ plays a significant role in determining the final performance. In our experiments, we conduct a preliminary study on the sensitivity of $p$ and observe that setting $p = 0.2$ consistently yields strong performance across tasks. We therefore recommend tuning this parameter when applying our method, as it can have a notable impact on both stability and accuracy.

\paragraph{Sampling time steps $(b,t)$} Time steps play a critical role in the training of MeanFlow~\cite{meanflow}, as they are directly involved in the Jacobian–vector product (JVP) computation required for learning the flow dynamics. Same to original MeanFlow \cite{vaswani2017attention,meanflow}, we use positional embedding to encode the time variables, which are then combined and provided as the conditioning of the neural network. 

However, we observed in our experiments that directly adopting the time step formulation from MeanFlow leads to training collapse. One possible reason is the significant difference in dimensionality between reinforcement learning and image generation: while the action space in RL typically has a low dimensionality (e.g., 5 to 21), image data often resides in much higher-dimensional spaces (e.g., at least $3 \times 32 \times 32$). This discrepancy causes uniform sampling of $(b, t) \sim \mathcal{U}(0, 1)$ to introduce excessive randomness and instability during training.

To address this issue, we find that appropriately simplifying the original MeanFlow formulation—such as by setting $b = 0$—can substantially reduce training complexity and make it feasible to learn effectively in low-dimensional action spaces. Accordingly, we explore both uniformly sampled discrete and continuous time steps from $\mathcal{U}(0, 1)$s, and conduct the experiments summarized in Table~\ref{tab:timestep_ablation} to evaluate their impact. Our empirical observations suggest that this choice is highly sensitive to the underlying task complexity. As a result, we consider it as a task-dependent hyperparameter subject to tuning. In particular, the choice of the time step should be adapted to the task complexity and the amount of available data. We recommend trying the following settings in order of expressiveness and stability: 
(i) both \( b \) and \( t \) as continuous variables, 
(ii) fixing \( b = 0 \) while keeping \( t \) continuous, and 
(iii) finally, using discrete timestamps for \( t \) with \( b = 0 \). 
This progressive relaxation enables practitioners to balance modeling flexibility and computational efficiency.

\begin{table}[htbp]
\centering
\caption{
\footnotesize
Performance under different time step configurations: discrete 50 steps, discrete 100 steps, and continuous sampling.}
\label{tab:timestep_ablation}
\scalebox{0.95}{
\begin{tabular}{l@{\hspace{6pt}}c@{\hspace{6pt}}c@{\hspace{6pt}}c}
\toprule
\multicolumn{1}{c}{} & \multicolumn{3}{c}{\textbf{Time Step Strategy}} \\
\cmidrule(lr){2-4}
\textbf{Environment} & {\texttt{Disc.(50)}} & {\texttt{Disc.(100)}} & {\texttt{Cont.}} \\
\midrule
{\texttt{antmaze-large-task1}} & $\mathbf{91}$ {\tiny $\pm 2$} & $78$ {\tiny $\pm 6$} & $49$ {\tiny $\pm 10$} \\
{\texttt{humanoidmaze-large-task1}} & $\mathbf{53}$ {\tiny $\pm 5$} & $44$ {\tiny $\pm 5$} & $42$ {\tiny $\pm 2$} \\
{\texttt{cube-single-task4}} & $88$ {\tiny $\pm 2$} & $89$ {\tiny $\pm 3$} & $\mathbf{95}$ {\tiny $\pm 2$} \\
{\texttt{antmaze-umaze-v2}} & $96$ {\tiny $\pm 1$} & $95$ {\tiny $\pm 1$} & $\mathbf{98}$ {\tiny $\pm 1$} \\
{\texttt{antmaze-medium-diverse}} & $\mathbf{78}$ {\tiny $\pm 2$} & $78$ {\tiny $\pm 4$} & $73$ {\tiny $\pm 5$} \\
{\texttt{pen-human-v1}} & $80$ {\tiny $\pm 5$} & $\mathbf{84}$ {\tiny $\pm 7$} & $72$ {\tiny $\pm 1$} \\
\bottomrule
\end{tabular}}
\end{table}

\paragraph{BC coefficient $\alpha$.}
The BC coefficient $\alpha$ is the key hyperparameter in our method, as it directly controls the trade-off between behavior cloning and value-based learning. We begin with a coarse grid search over ${1, 10, 100, 1000, 10000}$ across all tasks. Based on the preliminary results, we identify a promising subrange and perform a finer-grained search within that interval for each environment. The optimal value of $\alpha$ varies significantly across tasks and must be carefully tuned to achieve strong performance. Given its sensitivity and importance, we consider the automatic adaptation of $\alpha$ an important direction for future work. In particular, we plan to explore multi-objective optimisation techniques that jointly balance imitation fidelity and reward maximisation, thereby reducing the reliance on exhaustive manual tuning.

\paragraph{Baselines.}  
We adopt the same baseline configurations as those used in Flow Q-Learning (FQL)~\citep{fql}, ensuring full alignment in implementation and evaluation protocols. All baselines are implemented using the official FQL codebase without modification, and hyperparameters are retained as reported in~\citep{fql}.

\subsection{E\quad Experimental Details}
\subsubsection{E.1\quad Environments and Datasets}

\paragraph{OGBench~\citep{ogbench}.}
OGBench serves as our primary benchmark. We adopt a total of 10 state-based and 5 pixel-based tasks drawn from OGBench, comprising 50 state-based and 5 pixel-based offline RL tasks. Originally designed for offline goal-conditioned reinforcement learning, we utilise the single-task variants (denoted as ``\texttt{-singletask}'') of OGBench to benchmark conventional reward-maximising offline RL algorithms.

Each OGBench environment offers five distinct evaluation goals, corresponding to tasks named \texttt{-singletask-task1} through \texttt{-singletask-task5}, with one designated as the default task (denoted simply as \texttt{-singletask}). For a given evaluation goal, the corresponding single-task dataset provides transitions labelled with a semi-sparse reward function. This reward is computed as the negative number of remaining subtasks in a given state.

In locomotion environments, each task contains a single subgoal (e.g., “reach the target”), yielding rewards of either $-1$ or $0$. In contrast, manipulation environments typically involve multiple subtasks (e.g., “open the drawer”, “turn the button blue”), and the reward ranges from $-n_\mathrm{task}$ to $0$, where $n_\mathrm{task}$ denotes the number of subtasks, up to 16 in our selected environments. Episodes terminate upon successful completion of the specified goal.

We use the following datasets in our experiments, each providing five tasks:
\begin{itemize}[noitemsep, topsep=0pt]
    \item \textbf{State-based datasets:}
    \begin{itemize}[noitemsep, topsep=0pt]
        \item \texttt{antmaze-large-navigate-v0}
        \item \texttt{antmaze-giant-navigate-v0}
        \item \texttt{humanoidmaze-medium-navigate-v0}
        \item \texttt{humanoidmaze-large-navigate-v0}
        \item \texttt{antsoccer-arena-navigate-v0}
        \item \texttt{cube-single-play-v0}
        \item \texttt{cube-double-play-v0}
        \item \texttt{scene-play-v0}
        \item \texttt{puzzle-3x3-play-v0}
        \item \texttt{puzzle-4x4-play-v0}
    \end{itemize}
    \item \textbf{Pixel-based datasets:}
    \begin{itemize}[noitemsep, topsep=0pt]
        \item \texttt{visual-cube-single-play-v0}
        \item \texttt{visual-cube-double-play-v0}
        \item \texttt{visual-scene-play-v0}
        \item \texttt{visual-puzzle-3x3-play-v0}
        \item \texttt{visual-puzzle-4x4-play-v0}
    \end{itemize}
\end{itemize}

These environments present diverse challenges. The \texttt{antmaze} and \texttt{humanoidmaze} tasks require controlling quadrupedal (8-DOF) or humanoid (21-DOF) agents to reach target positions in maze layouts. \texttt{antsoccer} requires navigating a quadrupedal agent while dribbling a ball to a specified location. Manipulation tasks such as \texttt{cube}, \texttt{scene}, and \texttt{puzzle} involve robot arms interacting with various objects. In particular, \texttt{scene} tasks entail long-horizon control of multiple objects (up to 8 subtasks), while \texttt{puzzle} tasks demand combinatorial generalisation.

Pixel-based tasks (prefixed with \texttt{visual-}) involve controlling agents solely from raw pixel observations of size $64 \times 64 \times 3$. We employ standard task types: \texttt{navigate} for locomotion and \texttt{play} for manipulation. These datasets are highly suboptimal, comprising trajectories collected during the execution of \emph{random} tasks (e.g., reaching random goals or manipulating random objects), thus necessitating strong trajectory stitching capabilities. For state-based environments, we evaluate all five tasks, whereas for pixel-based environments, only \texttt{singletask-task1} is used due to computational constraints.

Evaluation follows the original benchmark protocol, using binary task success rates (in percentage).

\paragraph{D4RL~\citep{d4rl}.}
To facilitate direct comparison with prior work, we additionally evaluate on 18 challenging tasks from the D4RL benchmark. These include six \texttt{antmaze} tasks and twelve \texttt{adroit} tasks:
\begin{itemize}[noitemsep, topsep=0pt]
    \item \texttt{antmaze-umaze-v2}, \texttt{antmaze-umaze-diverse-v2}, \texttt{antmaze-medium-play-v2},\\
    \texttt{antmaze-medium-diverse-v2}, \texttt{antmaze-large-play-v2}, \texttt{antmaze-large-diverse-v2}
    \item \texttt{pen-human-v1}, \texttt{pen-cloned-v1}, \texttt{pen-expert-v1}
    \item \texttt{door-human-v1}, \texttt{door-cloned-v1}, \texttt{door-expert-v1}
    \item \texttt{hammer-human-v1}, \texttt{hammer-cloned-v1}, \texttt{hammer-expert-v1}
    \item \texttt{relocate-human-v1}, \texttt{relocate-cloned-v1}, \texttt{relocate-expert-v1}
\end{itemize}

The D4RL \texttt{antmaze} tasks share a similar objective with those in OGBench but use different (and typically less challenging) maze configurations, datasets, and evaluation goals. The \texttt{adroit} suite (\texttt{pen}, \texttt{door}, \texttt{hammer}, and \texttt{relocate}) involves high-dimensional ($24$-DOF) dexterous manipulation tasks. Evaluation metrics follow the original D4RL protocol: binary task success rates for \texttt{antmaze} and normalised returns for \texttt{adroit}.

To prevent excessively large numerical ranges in the input observations, we apply \emph{feature-wise normalisation} to the dataset. This preprocessing step ensures that the input distributions are well-conditioned, which is beneficial for stable and efficient Transformer training.

\subsubsection{E.2\quad Hyper-parameters}
Our hyperparameter configuration is organized into two main categories. The first encompasses general-purpose settings that are held constant across all tasks, including network architecture, optimisation details, and regularization coefficients (see Table~\ref{tab:hyperparameters}). The second category includes task-specific parameters—such as the behavior cloning coefficient~$\alpha$, the number of action candidates, and \texttt{time\_steps}—which are individually tuned for each environment to ensure stable training dynamics and optimal performance (see Table~\ref{tab:task-specific}).

\begin{table}[htbp]
\centering
\caption{Hyperparameters in Experiments}
\label{tab:hyperparameters}
\small
\renewcommand{\arraystretch}{1.15}
\begin{tabular*}{0.8\linewidth}{@{\extracolsep{\fill}}lcp{6.3cm}}  % 左-左-固定宽度描述
\toprule
\textbf{Parameter} & \textbf{Value} & \textbf{Description} \\
\midrule
\rowcolor{mask!88}
\multicolumn{3}{c}{\textbf{General Settings}} \\
sigma & 1.0 & Standard deviation of injected noise \\
batch\_size & 256 & Number of samples per batch \\
q\_agg & mean & Target Q aggregation method \\
normalize\_q\_loss & False & Normalize Q loss term \\
noise\_type & gaussian & Noise distribution type \\
optimizer & Adam~\citep{kingma2014adam} & Optimizer of networks \\
gradient steps & 1000000 (default), 500000 & Total gradient updates \\
discount $\gamma$ & 0.99 (default), 0.995 & Keep same as FQL\citep{fql} \\
image aug. prob. & 0.5 & Probability of image augmentation \\

\rowcolor{mask!88}
\multicolumn{3}{c}{\textbf{Critic Network}} \\
value\_hidden\_dims & (512, 512, 512, 512) & Hidden layer sizes \\
layer\_norm & True & Apply layer normalization \\
lr & $1\mathrm{e}{-4}$ & Learning rate \\
lr\_schedule & constant & Learning rate schedule \\
gradient clipping & 1.0 & Maximum gradient norm \\
network initialization & kaiming\_init & Network initialization techniques \\

\rowcolor{mask!88}
\multicolumn{3}{c}{\textbf{Actor Network}} \\
lr & $1\mathrm{e}{-4}$ & Learning rate \\
lr\_schedule & cosine\_with\_warmup & Learning rate schedule \\
lr\_min\_ratio & 0.1 & Minimum ratio in cosine scheduler \\
actor\_hidden\_dims & 256 & Hidden size of actor MLP \\
actor\_depth & 3 & Number of transformer layers \\
actor\_num\_heads & 2 & Number of attention heads \\
actor\_layer\_norm & False & Use layer norm in actor \\
discount & 0.99 & Discount factor $\gamma$ \\
tau & 0.005 & Target network update rate \\
alpha & Task-dependent & See Table~\ref{tab:task-specific} \\
tanh\_squash & False & Use tanh at actor output \\
use\_output\_layernorm & False & Norm at final actor output \\
gradient clipping & 1.0 & Maximum gradient norm \\
network initialization & zero\_init & Network initialization techniques \\
\rowcolor{mask!88}
\multicolumn{3}{c}{\textbf{Flow-related Settings}} \\
adaptive\_gamma & 0.8 & Velocity loss weight \\
adaptive\_c & $1\mathrm{e}{-4}$ & Position loss weight \\
bound\_loss\_weight & 1.0 & Weight for boundary loss \\
time\_steps & Task-dependent & See Table~\ref{tab:task-specific} \\

\rowcolor{mask!88}
\multicolumn{3}{c}{\textbf{Best-of-N Sampling}} \\
num\_candidates & Task-dependent & See Table~\ref{tab:task-specific} \\
action\_mode & best & Action selection strategy \\

\rowcolor{mask!88}
\multicolumn{3}{c}{\textbf{Alpha Scheduling}} \\
use\_dynamic\_alpha & True & Enable dynamic $\alpha$ tuning \\
alpha\_update\_interval & 2000 & Adjustment frequency \\
loss\_multiplier\_threshold & 5 & Threshold to increase $\alpha$ \\
alpha\_increase\_factor & 1.2 & Multiplier for increase \\
alpha\_decrease\_factor & 0.8 & Multiplier for decrease \\
loss\_history\_window\_size & 20 & History window size \\
\bottomrule
\end{tabular*}
\end{table}

\begin{table*}[htbp]
\vspace{-20pt}
\caption{
\footnotesize
\textbf{Task-Specific Hyperparameter Settings.}
We present the hyperparameters ($\alpha$, number of candidates, and time steps) for all tasks. 
{(*)} indicates the default task in each environment.
}
\label{tab:task-specific}
\centering
\vspace{3pt}
\scalebox{1.2}
{
\begin{threeparttable}
\setlength{\tabcolsep}{12pt}
\tiny
\begin{tabular}{lccc}
\toprule
  {\texttt{Environment}} &   {\texttt{alpha}} &   {\texttt{num\_candidates}} &   {\texttt{time\_steps}} \\
\midrule
  {\texttt{antmaze-large-navigate-singletask-task1-v0 (*)}} & 10 & 5 & 50 \\
  {\texttt{antmaze-large-navigate-singletask-task2-v0}} & 1000 & 5 & 50 \\
  {\texttt{antmaze-large-navigate-singletask-task3-v0}} & 2350 & 5 & 50 \\
  {\texttt{antmaze-large-navigate-singletask-task4-v0}} & 7 & 1 & 50 \\
  {\texttt{antmaze-large-navigate-singletask-task5-v0}} & 15 & 5 & 50 \\
\midrule
  {\texttt{antmaze-giant-navigate-singletask-task1-v0 (*)}} & 1000 & 1 & 50 \\
  {\texttt{antmaze-giant-navigate-singletask-task2-v0}} & 1000 & 1 & 50 \\
  {\texttt{antmaze-giant-navigate-singletask-task3-v0}} & 1000 & 1 & 50 \\
  {\texttt{antmaze-giant-navigate-singletask-task4-v0}} & 1000 & 1 & 50 \\
  {\texttt{antmaze-giant-navigate-singletask-task5-v0}} & 1000 & 1 & 50 \\
\midrule
  {\texttt{humanoidmaze-medium-navigate-singletask-task1-v0 (*)}} & 150 & 5 & 50 \\
  {\texttt{humanoidmaze-medium-navigate-singletask-task2-v0}} & 10000 & 5 & 100 \\
  {\texttt{humanoidmaze-medium-navigate-singletask-task3-v0}} & 9400 & 1 & 100 \\
  {\texttt{humanoidmaze-medium-navigate-singletask-task4-v0}} & 60 & 1 & 50 \\
  {\texttt{humanoidmaze-medium-navigate-singletask-task5-v0}} & 150 & 5 & 50 \\
\midrule
  {\texttt{humanoidmaze-large-navigate-singletask-task1-v0 (*)}} & 10000 & 5 & 50 \\
  {\texttt{humanoidmaze-large-navigate-singletask-task2-v0}} & 1000 & 5 & 50 \\
  {\texttt{humanoidmaze-large-navigate-singletask-task3-v0}} & 1000 & 5 & 100 \\
  {\texttt{humanoidmaze-large-navigate-singletask-task4-v0}} & 1000 & 1 & 50 \\
  {\texttt{humanoidmaze-large-navigate-singletask-task5-v0}} & 10000 & 5 & 10000 \\
\midrule
  {\texttt{antsoccer-arena-navigate-singletask-task1-v0}} & 200 & 10 & 50 \\
  {\texttt{antsoccer-arena-navigate-singletask-task2-v0}} & 100 & 5 & 100 \\
  {\texttt{antsoccer-arena-navigate-singletask-task3-v0}} & 1100 & 5 & 50 \\
  {\texttt{antsoccer-arena-navigate-singletask-task4-v0 (*)}} & 10000 & 5 & 50 \\
  {\texttt{antsoccer-arena-navigate-singletask-task5-v0}} & 200 & 5 & 50 \\
\midrule
  {\texttt{cube-single-play-singletask-task1-v0}} & 200 & 5 & 50 \\
  {\texttt{cube-single-play-singletask-task2-v0 (*)}} & 200 & 1 & 50 \\
  {\texttt{cube-single-play-singletask-task3-v0}} & 50 & 1 & 100 \\
  {\texttt{cube-single-play-singletask-task4-v0}} & 11000 & 5 & 10000 \\
  {\texttt{cube-single-play-singletask-task5-v0}} & 180 & 1 & 50 \\
\midrule
  {\texttt{cube-double-play-singletask-task1-v0}} & 200 & 5 & 50 \\
  {\texttt{cube-double-play-singletask-task2-v0 (*)}} & 120 & 5 & 50 \\
  {\texttt{cube-double-play-singletask-task3-v0}} & 180 & 5 & 50 \\
  {\texttt{cube-double-play-singletask-task4-v0}} & 200 & 5 & 50 \\
  {\texttt{cube-double-play-singletask-task5-v0}} & 250 & 5 & 50 \\
\midrule
  {\texttt{scene-play-singletask-task1-v0}} & 8000 & 5 & 10000 \\
  {\texttt{scene-play-singletask-task2-v0 (*)}} & 11000 & 5 & 50 \\
  {\texttt{scene-play-singletask-task3-v0}} & 12000 & 5 & 50 \\
  {\texttt{scene-play-singletask-task4-v0}} & 9800 & 1 & 10000 \\
  {\texttt{scene-play-singletask-task5-v0}} & 8000 & 5 & 50 \\
\midrule
  {\texttt{puzzle-3x3-play-singletask-task1-v0}} & 1000 & 5 & 100 \\
  {\texttt{puzzle-3x3-play-singletask-task2-v0}} & 30 & 1 & 50 \\
  {\texttt{puzzle-3x3-play-singletask-task3-v0}} & 60 & 1 & 50 \\
  {\texttt{puzzle-3x3-play-singletask-task4-v0 (*)}} & 100 & 5 & 50 \\
  {\texttt{puzzle-3x3-play-singletask-task5-v0}} & 60 & 1 & 50 \\
\midrule
  {\texttt{puzzle-4x4-play-singletask-task1-v0}} & 1000 & 5 & 100 \\
  {\texttt{puzzle-4x4-play-singletask-task2-v0}} & 5550 & 5 & 100 \\
  {\texttt{puzzle-4x4-play-singletask-task3-v0}} & 1200 & 5 & 100 \\
  {\texttt{puzzle-4x4-play-singletask-task4-v0 (*)}} & 8000 & 5 & 10000 \\
  {\texttt{puzzle-4x4-play-singletask-task5-v0}} & 2450 & 5 & 50 \\
\midrule
  {\texttt{antmaze-umaze-v2}} & 100 & 5 & 10000 \\
  {\texttt{antmaze-umaze-diverse-v2}} & 130 & 5 & 100 \\
  {\texttt{antmaze-medium-play-v2}} & 10 & 5 & 50 \\
  {\texttt{antmaze-medium-diverse-v2}} & 50 & 5 & 50 \\
  {\texttt{antmaze-large-play-v2}} & 10 & 5 & 100 \\
  {\texttt{antmaze-large-diverse-v2}} & 6 & 5 & 50 \\
\midrule
  {\texttt{pen-human-v1}} & 10000 & 5 & 100 \\
  {\texttt{pen-cloned-v1}} & 10000 & 5 & 100 \\
  {\texttt{pen-expert-v1}} & 10000 & 5 & 50 \\
  {\texttt{door-human-v1}} & 5000 & 1 & 100 \\
  {\texttt{door-cloned-v1}} & 9000 & 5 & 50 \\
  {\texttt{door-expert-v1}} & 10000 & 5 & 50 \\
\midrule
  {\texttt{hammer-human-v1}} & 7000 & 1 & 100 \\
  {\texttt{hammer-cloned-v1}} & 11000 & 5 & 100 \\
  {\texttt{hammer-expert-v1}} & 12000 & 5 & 50 \\
  {\texttt{relocate-human-v1}} & 10000 & 5 & 100 \\
  {\texttt{relocate-cloned-v1}} & 10000 & 5 & 50 \\
  {\texttt{relocate-expert-v1}} & 15000 & 5 & 100 \\
\midrule
  {\texttt{visual-cube-single-play-singletask-task1-v0}} & 360 & 5 & 100 \\
  {\texttt{visual-cube-double-play-singletask-task1-v0}} & 350 & 5 & 100 \\
  {\texttt{visual-scene-play-singletask-task1-v0}} & 300 & 1 & 100 \\
  {\texttt{visual-puzzle-3x3-play-singletask-task1-v0}} & 300 & 5 & 100 \\
  {\texttt{visual-puzzle-4x4-play-singletask-task1-v0}} & 1500 & 5 & 100 \\
\bottomrule
\end{tabular}
\end{threeparttable}
}
\end{table*}

\end{document}